\documentclass{article} 
\usepackage{iclr2027_conference,times}

\usepackage{amsmath,amsfonts,bm}

\def\eqref#1{equation~\ref{#1}}
\def\Eqref#1{Equation~\ref{#1}}

\def\1{\bm{1}}

\def\eps{{\epsilon}}

\def\vtheta{{\bm{\theta}}}

\def\vu{{\bm{u}}}

\DeclareMathAlphabet{\mathsfit}{\encodingdefault}{\sfdefault}{m}{sl}
\SetMathAlphabet{\mathsfit}{bold}{\encodingdefault}{\sfdefault}{bx}{n}

\def\gA{{\mathcal{A}}}

\def\gL{{\mathcal{L}}}

\def\gO{{\mathcal{O}}}

\newcommand{\E}{\mathbb{E}}

\usepackage{hyperref}
\usepackage{url}

\usepackage{graphicx}
\usepackage{subcaption}
\usepackage{tabularx}
\usepackage{booktabs,multirow}
\usepackage{cleveref}
\usepackage{xcolor}
\usepackage{amsmath, amssymb, amsthm}
\usepackage{graphicx}
\usepackage{enumitem}

\newtheorem{theorem}{Theorem}[section]
\newtheorem{proposition}[theorem]{Proposition}

\newtheorem{corollary}[theorem]{Corollary}

\theoremstyle{definition}

\newtheorem{assumption}{Assumption}
\theoremstyle{remark}

\usepackage{wrapfig}

\newcommand{\tuluthree}{T\"{u}lu~3}

\newcommand{\Cblkfwd}{C_{\mathrm{fwd},\mathrm{blk}}}
\newcommand{\Cfullfwd}{C_{\mathrm{fwd}, \mathrm{full}}}
\newcommand{\Cheadfwd}{C_{\mathrm{fwd},\mathrm{head}}}
\newcommand{\Cfullbwd}{C_{\mathrm{bwd},\mathrm{full}}}
\newcommand{\Cfull}{C_{\mathrm{full}}}
\newcommand{\Cuncached}{C_{\mathrm{uncached}}}
\newcommand{\rhonc}{\rho_{\mathrm{uncached}}}
\newcommand{\Cbuild}{C_{\mathrm{build}}}
\newcommand{\Ccached}{C_{\mathrm{cached}}}
\newcommand{\rhocached}{\rho_{\mathrm{cached}}}
\newcommand{\rhoe}{\rho}
\newcommand{\cblkfwd}{c_{\mathrm{fwd},\mathrm{blk}}}

\newcommand{\arxiv}[1]{{#1}}

\newcommand{\method}{TipSweep}

\title{
Precision-Aware Learning Rate Transfer\\
Across Block Count in LLM Post-Training
}

\author{Nan Chen, Soledad Villar \& Soufiane Hayou  \\
Department of Applied Mathematics and Statistics\\
Johns Hopkins University\\
\texttt{\{nchen38,svillar3,hayou\}@jhu.edu} \\
}

\iclrfinalcopy 
\begin{document}

\maketitle
\lhead{}

\begin{abstract}
The post-training of pretrained large language models, including supervised finetuning (SFT), direct preference optimization (DPO), and reinforcement learning with verifiable reward (RLVR), is very sensitive to the learning rate, and tuning it through grid search is costly in both computation and memory. 
In this work, we propose \textbf{\method{}}, a method to transfer the learning rate across the number of trainable transformer blocks. To do so efficiently, we search for the best learning rate by training only the last $k_0$ blocks (in our experiments, typically $k_0=1$), and provide scaling rules that predict the optimal learning rate when training any set of $k$ blocks, including when $k=L$ (the full transformer). 
Across experiments with models ranging from 1 to 7 billion parameters, we observe that the optimal learning rate follows precision-dependent scaling rules: the optimal rate scales approximately as $k^{-1}$ with float32 (FP32) master weights, but as $k^{-1/2}$ when updates are applied directly to bfloat16 (BF16) weights without a master copy. 
We explain both rules using an analytical model that accounts for update rounding, under assumptions supported by profiling experiments.
\arxiv{Finally, we develop two prototypes of the learning rate search for a 6.7-billion-parameter model.
The first fits the search on a single consumer GPU with 16 GiB of memory.
The second speeds up the search on a high-end GPU by evaluating all candidate learning rates in a single pass over the data, so that a five-candidate search takes about half the GPU hours of one full-depth ($k=L$) run.
}
\end{abstract}

\section{Introduction}

Pretrained large language models (LLMs)~\citep{yang2025qwen3} are publicly available for adaptation through post-training on fixed datasets, including supervised finetuning (SFT)~\citep{ouyang2022training}, direct preference optimization (DPO)~\citep{rafailov2023direct}, and \arxiv{reinforcement learning with verifiable reward (RLVR)~\cite{shao2024deepseekmath}}.
However, post-training performance is sensitive to the learning rate, and selecting it is costly because each candidate value requires a separate run~\citep{lambert2025tulu}. 
For pretraining, the maximal-update parameterization (\(\mu\)P) provides a method for efficient learning rate search, by tuning on a narrower proxy and transferring the selected rate to a wider target model \citep{yang2021tuning}. 
However, this approach cannot be used with pretrained checkpoints since model width is fixed and cannot be changed during post-training. This motivates finding a transfer dimension for LLM post-training that leaves the checkpoint architecture fixed.

In this work, we propose \method{}, a method that transfers learning rate across the number of trainable transformer blocks, which is a suitable transfer axis for post-training. 
Specifically, we search for the optimal learning rate by training only the last \(k_0\) of the model's \(L\) transformer blocks.
The selected rate is then scaled for a subsequent set of $k$ trainable blocks. A particularly useful case is to do a full-depth run in which all transformer blocks are trainable, namely $k=L$.
For \(k_0<L\), restricting backpropagation through transformer blocks to the last \(k_0\) reduces the computation and GPU memory requirements relative to its full-depth counterpart.
In this setting, we cache the frozen prefix's outputs for reuse across multiple trials to reduce computation.
As shown in~\Cref{fig:intro_motivations:flops}, searching with a last-block proxy (\(k_0=1\)) reduces the total FLOPs of a 16-trial learning rate sweep on Llama-2-7B by a factor of approximately \(8\) relative to running all 16 trials purely at full depth. This reduction includes the cache construction and the subsequent full-depth run.

Our approach hinges on one question: Does the optimal learning rate scale predictably with the number of trainable blocks $k$?
As shown in~\Cref{fig:intro_motivations:tf32,fig:intro_motivations:bf16}, we observe that the optimal learning rate roughly follows a power law in \(k\).
Surprisingly, the exponent depends on the precision used to accumulate parameter updates. For the same model and dataset, the optimal learning rate scales as \(k^{-1}\) when updates are accumulated in float32 (FP32) master weights, while it scales as \(k^{-1/2}\) when updates are applied directly to bfloat16 (BF16) weights without a float32 master copy.

To understand these patterns, we provide a theoretical analysis that takes parameter-update rounding into account.
In this analysis, we define a stable loss-reduction criterion that characterizes how the optimal learning rate should behave as the number of trainable blocks increases.
Under assumptions supported by profiling in the training regimes we study, we derive exact scaling rules that match  the observations for both FP32 and BF16 updates.

\arxiv{We verify these scaling rules across SFT and DPO, using 1B--7B models and multiple post-training datasets.
We also show that they extend to on-policy RLVR on an 8B model, and that direct float16 (FP16) updates follow the same $k^{-1/2}$ rule as BF16.}
\arxiv{Furthermore, we implement two prototypes of the proposed learning rate search strategy.
The first caches the frozen prefix's outputs to disk and can tune the learning rate of a 6.7B model on a consumer GPU with approximately 16~GiB of available memory in reasonable time.
This makes it possible to delegate the search trials to widely available consumer GPUs and reserve scarce high-end GPUs only for the final full-depth run.
The second instead uses a high-end GPU to speed up the search itself.
It keeps the frozen prefix's outputs for the current batch in GPU memory rather than on disk and shares them across all trials, so a single pass over the dataset completes the search.
For the same 6.7B model, a five-candidate search then takes about half the GPU hours of a single full-depth run.}
Our contributions are summarized as follows:

\begin{figure}[!t]
\centering
\begin{subfigure}[t]{0.33\textwidth}
  \includegraphics[width=\linewidth]{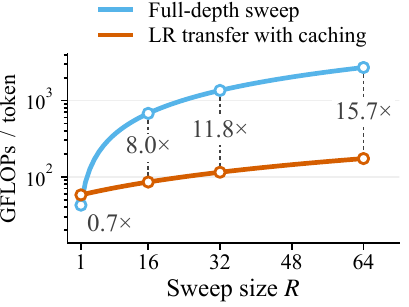}
  \caption{Analytical FLOPs}
  \label{fig:intro_motivations:flops}
\end{subfigure}\hfill
\begin{subfigure}[t]{0.33\textwidth}
  \includegraphics[width=\linewidth]{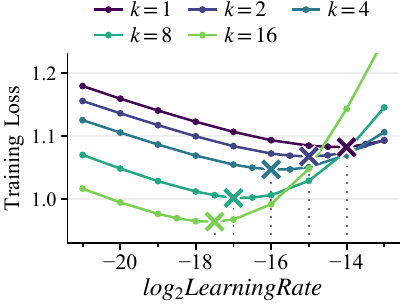}
  \caption{Float32 weights}
  \label{fig:intro_motivations:tf32}
\end{subfigure}\hfill
\begin{subfigure}[t]{0.33\textwidth}
  \includegraphics[width=\linewidth]{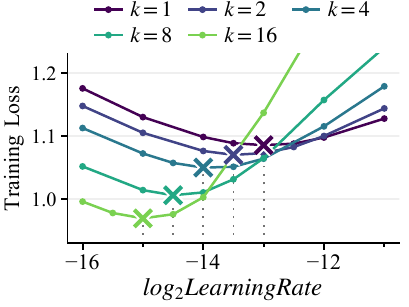}
  \caption{BFloat16 weights}
  \label{fig:intro_motivations:bf16}
\end{subfigure}
\caption{a) FLOPs per target token against sweep size $R$ for a full-depth sweep and for learning rate (LR) transfer with caching. 
b) and c) Learning rate sweeps on Llama-3.2-1B (\tuluthree{}) with the last $k$ blocks trainable under FP32 master-weight and direct BF16 updates, respectively.
}
\label{fig:intro_motivations}
\end{figure}

\begin{itemize}[leftmargin=*]
    \item \textbf{Efficient learning rate search.} We propose learning rate transfer across the number of trainable transformer blocks, combined with frozen-prefix feature caching, to reduce the computation and memory required to select a learning rate for LLM post-training.

    \item \textbf{Precision-dependent scaling.} We identify that the learning rate scaling with trainable block count $k$ follows scaling rules \(k^{-1}\) under FP32 updates and \(k^{-1/2}\) under BF16 updates, supported by SFT, DPO, \arxiv{and RLVR} experiments.

    \item \textbf{Theoretical analysis.} We provide a theoretical analysis explaining these scaling rules, based on a stable loss reduction criterion \arxiv{ and the effect of update rounding in low-precision weights}.

    \item \textbf{Empirical results and case study.} We conduct extensive experiments to examine the observed scaling rules. 
    \arxiv{We implement and evaluate two prototypes of the proposed learning rate search for a 6.7B model.
    One can run on a consumer GPU with approximately 16~GiB of available memory, and the other can run up to 40 trials at the same time on a single high-end GPU.}
\end{itemize}

\section{Related Work}

\paragraph{LLM post-training.}

Post-training teaches a pretrained LLM to follow instructions and improves its reasoning, typically in three stages~\citep{qwen2024qwen25,lambert2025tulu,walsh2025,guo2025deepseek,olmo2025olmo}.
Supervised finetuning (SFT) trains the model to imitate reference responses~\citep{ouyang2022training,zhou2023lima,chung2024scaling,chu2025sft}.
Direct preference optimization (DPO) trains it to favor chosen over rejected responses~\citep{rafailov2023direct,meng2024simpo,liu2024statistical}.
Reinforcement learning with verifiable rewards (RLVR) trains it on its own responses, rewarding those that pass programmatic checks~\citep{shao2024deepseekmath,yu2025dapo,liu2026tricks}.
Each stage requires tuning its own learning rate, typically through costly full-depth runs.
We instead search on a proxy with only the last few blocks trainable and scale the selected rate for a subsequent full-depth run.

Another line of work lowers the cost of post-training: parameter-efficient finetuning (PEFT)~\citep{liu2022fewshot,han2024parameterefficient}, such as low-rank adaptation (LoRA) and its variants~\citep{hu2022lora,kopiczko2024vera,liu2024dora}, reduces the number of trainable parameters, while quantization trades numerical precision for throughput or memory~\citep{peng2023fp8,dettmers2023qlora,li2024loftq,xu2024qalora}.
Our work instead reduces the cost of learning rate search while updating pretrained weights in-place, without adapters.

\paragraph{Mixed-precision training.}
Mixed-precision training uses 16-bit computation but accumulates updates in FP32 master weights~\citep{micikevicius2018mixed,rajbhandari2020zero}.
BF16's FP32-like dynamic range~\citep{burgess2019bfloat16,kalamkar2019study} allows dropping the master copy to save memory~\citep{yu2024collage,ozkara2025stochastic}, but round-to-nearest can then discard small updates at small learning rates~\citep{zamirai2020revisiting,rae2021scaling}.
In this work, we observe that this rounding also changes learning rate scaling with the number of trainable blocks.

\paragraph{Hyperparameter transfer.}
Hyperparameter transfer characterizes how optimal hyperparameters change along a dimension of scale, so they can be tuned on a small proxy and transferred to the target model. 
Early works scale across width in pretraining~\citep{yang2021tensor,yang2021tuning,yang2023spectral}, including with FP8 arithmetic~\citep{blake2025u}.
Depth extensions follow the same recipe~\citep{hayou2019impact,hayou2021stable,he2023deep,hayou2023width,noci2023shaped,yang2024tensor,bordelon2024depthwise}. 
Both lines of work derive their scaling rules from a prescribed random initialization, and are stated for models trained from it. 
Existing post-training works carry that assumption over to LoRA initialization and analyze a single adapted layer~\citep{hayou2024impact,hayou2024lora,libeyond,chen2026learning}, leaving depth out of scope. 
We analyze scaling across the number of trainable blocks when tuning the pretrained weights directly, whose distribution is set by a pretraining run we cannot observe.
We therefore identify the rules empirically and provide a theoretical explanation for their dependence on update precision.

\section{Methodology}
\label{sec:methodology}

\subsection{Precision-aware learning rate transfer}
\label{sec:methodology:transfer}

For a pretrained checkpoint with $L$ transformer blocks, we finetune the model with $R$ distinct learning rates on a grid with only the last $k_0$ blocks trainable, where $1 \le k_0 < L$.
We train all parameters in these blocks, including those of the normalization modules, and freeze the rest of the model.
Let $\widehat{\eta}_{k_0}$ be the optimal learning rate, i.e. the learning rate with smallest training loss. To finetune $k$ blocks, \method{} gives precision-dependent scaling rules for the optimal learning rate $\widehat{\eta}_{k}$:
\[
\widehat{\eta}_{k} \approx
\begin{cases}
({k_0}/{k})\cdot\,\widehat{\eta}_{k_0},
  & \text{FP32 master-weight updates}, \\
\sqrt{{k_0}/{k}}\cdot\,\widehat{\eta}_{k_0},
  & \text{direct BF16 updates without an FP32 master copy}.
\end{cases}
\]
We initialize from the same checkpoint and train the last $k$ transformer blocks at $\widehat{\eta}_{k}$ under the same update precision, freezing all other parameters. \Cref{fig:exp_depth_lr_fit} shows a good match between these scaling rules and empirical observations, and additional results are provided in \Cref{sec:experiments}. Using \method{}, we can tune the learning rate with small $k_0$, and transfer it to the full-depth run ($k=L$). 

\paragraph{Feature caching.}


When searching with only the last $k_0$ blocks trainable, backpropagation passes through the frozen language model (LM) head to the trainable blocks but does not enter the frozen prefix. 
Without feature caching, however, each trial still recomputes the frozen prefix on every forward pass. Feature caching removes this repeated computation.
For a fixed tokenized dataset with deterministic prefix evaluation, we first pass the dataset through blocks $1,\ldots,L-k_0$ in inference mode and cache the outputs of block $L-k_0$ on disk, a stage we call \emph{cache build}.
Each proxy trial (a run on a single learning rate value from the grid) then loads these features and performs forward and backward propagation through the last $k_0$ blocks and LM head, which we call a \emph{cached proxy trial}. 
The full pipeline is illustrated in~\Cref{fig:case_study:prototype}.
\arxiv{Alternatively, all $R$ proxy trials can run simultaneously, each on its own copy of the last $k_0$ blocks, so that the frozen prefix processes each batch once and its output is cached in GPU memory only until every trial has used it (\Cref{fig:case_study:prototype_2}).
When each proxy trial trains for one epoch, this variant is computationally equivalent to one cache build followed by $R$ cached proxy trials, but trades disk storage for GPU memory that grows with $R$.}

\subsection{Analytical FLOPs}
\label{sec:methodology:efficiency}

To estimate computational cost, we calculate FLOPs over a single epoch, counting dense matrix products in forward and backward propagation at two FLOPs per multiply--add~\citep{kaplan2020scaling,narayanan2021efficient,hoffmann2022training}.
We omit non-matrix operations, optimizer updates, and any activation recomputation.
Appendix~\ref{app:flops_analysis} gives the detailed FLOP analysis.

Let $\cblkfwd$ denote the forward FLOPs per token of one transformer block. 
A backward pass through a trainable block costs $2\,\cblkfwd$, and the frozen LM head costs $\gamma\,\cblkfwd$ per pass, forward or backward, for some $\gamma>0$.
Hence, the per-token FLOPs of each procedure are 
\[
\begin{aligned}
\text{Full-depth run:}\quad &(3L+2\gamma)\,\cblkfwd,
&\qquad \text{Uncached proxy trial:}\quad &(L+2k_0+2\gamma)\,\cblkfwd, \\
\text{Cache build:}\quad &(L-k_0)\,\cblkfwd,
&\qquad \text{Cached proxy trial:}\quad &(3k_0+2\gamma)\,\cblkfwd.
\end{aligned}
\]
Without caching, a proxy trial saves less than a factor of three, $(3L+2\gamma)/(L+2k_0+2\gamma)<3$, because the frozen prefix still executes in every trial.
We now compare the complete cost of using \method{} to perform learning rate transfer with caching against a full-depth sweep. We account for the cache construction, $R$ proxy trials, and the final full-depth run.
For the full-depth sweep, since every trial trains the full model, the best of its $R$ trials can be adopted directly without a final retraining.
The resulting per-token FLOP ratio of the full-depth sweep to \method{} is
\begin{equation}
\label{eq:methodology:e2e}
\rhoe(R,k_0) =
\frac{R(3L+2\gamma)}
{\underbrace{(L-k_0)}_{\text{cache build}}
 + \underbrace{R(3k_0+2\gamma)}_{\text{proxy trials}}
 + \underbrace{(3L+2\gamma)}_{\text{full-depth run}}}.
\end{equation}
The ratio in \Eqref{eq:methodology:e2e} monotonically increases with $R$ and decreases with $k_0$, but satisfies
\[
\rhoe(R,k_0)
< \min\!\left\{R,\frac{3L+2\gamma}{3k_0+2\gamma}\right\}
\le \min\!\left\{R,\frac{L}{k_0}\right\}.
\]
The two bounds correspond to two regimes of sweep grid size $R$, distinguished by which term dominates the denominator of \Eqref{eq:methodology:e2e}.
For small sweeps, the one-time costs of cache build and the final full-depth run dominate.
Each additional proxy trial then costs only a small fraction of a full-depth trial, so increasing $R$ amortizes the one-time costs and raises $\rhoe$.
For large sweeps, the proxy trials dominate and $\rhoe$ approaches the per-trial ratio $(3L+2\gamma)/(3k_0+2\gamma)$, the cost of a full-depth trial relative to a cached proxy trial.
Further increasing $R$ yields diminishing gains, whereas reducing $k_0$ raises this limit.
We instantiate this cost comparison on Llama-2~\citep{touvron2023llama} with $k_0=1$.
\Cref{fig:intro_motivations:flops} reports the Llama-2-7B configuration, and \Cref{fig:flops_analysis_llama2} includes more cases.

\section{Learning Rate Scaling under Update Rounding}
\label{sec:theory}

\begin{figure}[!t]
\centering
\begin{subfigure}[t]{0.333333\linewidth}\captionsetup{margin={6.92pt,0pt}}
    \includegraphics[width=\linewidth]{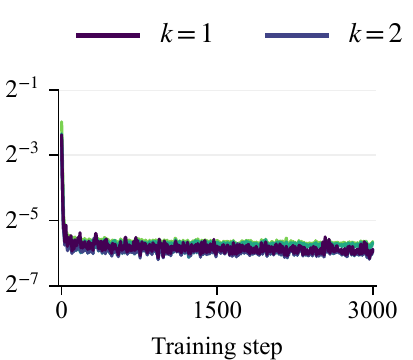}
    \caption{$\eta\,\widetilde{D}_k(\eta)$}
    \label{fig:theory_profile:a}
\end{subfigure}\hfill
\begin{subfigure}[t]{0.333333\linewidth}\captionsetup{margin={9.36pt,0pt}}
    \includegraphics[width=\linewidth]{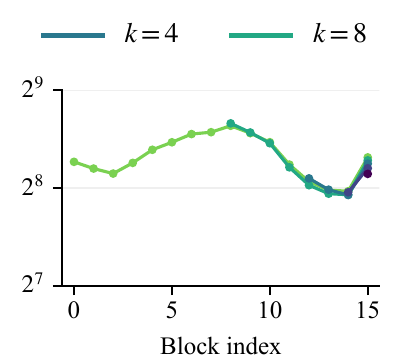}
    \caption{$d_l=\langle\nabla_{\vtheta_l}\gL,\vu_l\rangle$}
    \label{fig:theory_profile:b}
\end{subfigure}\hfill
\begin{subfigure}[t]{0.333333\linewidth}\captionsetup{margin={11.76pt,0pt}}
    \includegraphics[width=\linewidth]{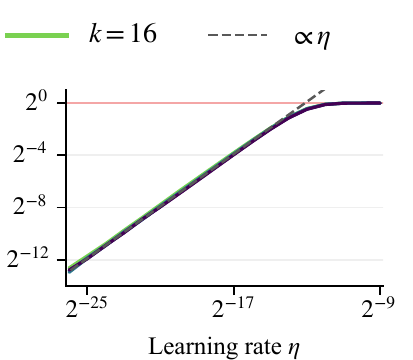}
    \caption{$\phi_k(\eta)=\widetilde D_k(\eta)/D_k$}
    \label{fig:theory_profile:c}
\end{subfigure}
\caption{Profiling results for SFT of Llama-3.2-1B on \tuluthree{} with FP32 weights.
Each curve corresponds to a run with the last $k$ blocks trainable at $\eta=2^{-14}/k$.
a) Realized first-order loss reduction against training step.
b) Geometric mean over training steps of the per-block first-order loss contribution $\widetilde d_l(\eta)$, which equals $d_l$ under the FP32 precision, against block index.
c) BF16 update factor $\phi_k(\eta)$ against the probe learning rate $\eta$ at step 3000.
Complete results in Appendix~\ref{sec:app-theory}.
}
\label{fig:theory_profile}
\end{figure}

In this section, we provide a theoretical justification for the observed scaling rules. The analysis is carried out at a physics level of rigor, making the modeling assumptions explicit and deferring other technical details to the appendix.
The profiling during training supports all the assumptions, and representative results are shown in~\Cref{fig:theory_profile}. 
Due to space constraints, we defer the proofs, profiling setup details, and complete profiling results to Appendix~\ref{sec:app-theory}.

Consider a transformer with $L$ blocks, where $\vtheta_l$ denotes the parameters of block $l$. 
Write $\vtheta=(\vtheta_1,\ldots,\vtheta_L)$ for the collection of block parameters and $\gL(\vtheta)$ for the loss. 
We finetune the last $k$ blocks, $\gA_k=\{L-k+1,\ldots,L\}$, with all remaining parameters frozen. For $l\in\gA_k$, let $\vu_l$ denote the update direction produced by Adam.
With learning rate $\eta$, the stored parameters are updated as
\[
 \vtheta_l^+=Q(\vtheta_l-\eta\vu_l),
 \qquad Q(\vtheta_l)=\vtheta_l,
 \qquad l\in\gA_k,
\]
where $Q$ denotes the storage-precision rounding operator.
Under FP32 master-weight accumulation, $Q$ is the identity function.
Under direct BF16 updates without an FP32 master copy, $Q$ rounds each coordinate to the nearest representable BF16 value.

We define the \emph{nominal} first-order loss contribution of block $l$, and its sum over trainable blocks, as
\[
    d_l=\langle\nabla_{\vtheta_l}\gL,\vu_l\rangle,
    \qquad D_k=\sum\nolimits_{l\in\gA_k}d_l.
\]
With the realized direction $\widetilde{\vu}_l(\eta)=(\vtheta_l-\vtheta_l^+)/\eta$, we define their \emph{realized} counterparts as
\[
 \widetilde d_l(\eta)
 =\langle\nabla_{\vtheta_l}\gL,\widetilde{\vu}_l(\eta)\rangle,\qquad
  \widetilde D_k(\eta)=\sum\nolimits_{l\in\gA_k}\widetilde d_l(\eta).
\]
For $D_k>0$, we define the \emph{update factor} $\phi_k(\eta)=\widetilde D_k(\eta)/D_k$.
This factor is the ratio of the realized first-order loss reduction to its nominal value and equals one under the FP32 precision.

A first order Taylor expansion gives the following approximation to loss reduction after one step
\[
 \Delta\gL_k(\eta)
 :=\gL(\vtheta^+)-\gL(\vtheta)
 =-\eta\widetilde D_k(\eta)+\gO(\eta^2)
  =-\eta D_k\phi_k(\eta)+\gO(\eta^2).
\]
For our analysis, we omit the term $\gO(\eta^2)$ and focus on the first order term. The next assumption is fundamental to our analysis and assumes that optimal learning rates are associated with stable loss reduction.
Throughout, we use the asymptotic notation $\Theta(\cdot)$ to denote two-sided bounds with positive constants independent of $k$ and $L$.\footnote{Precisely, we say for two quantities $A$ and $B$ that $B = \Theta(A)$ if there exists some constants $c_1, c_2 >0$ independent of $k$ and $L$ such that $B/A \in [c_1, c_2]$. In other words, the notation $B = \Theta(A)$ means the two quantities have roughly the same order independently of $k$ and $L$.} 

\begin{assumption}[Stable loss reduction]
\label{asm:theory-stable}
When only the last $k$ blocks are trainable, the optimal learning rate $\eta_k$ satisfies
\begin{equation}
 -\Delta\gL_k(\eta_k)=\Theta(1).
 \label{eq:theory-stable}
\end{equation}
\end{assumption}
Under the first-order approximation, \Eqref{eq:theory-stable} is equivalent to
\begin{equation}
 \eta_k D_k\,\phi_k(\eta_k)=\Theta(1).
 \label{eq:theory-stable-criterion}
\end{equation}



Intuitively, the optimal learning rate will scale with $k$ in a way that makes the loss reduction neither vanish nor explode.
\arxiv{Since $\Theta(1)$ hides constants independent of $k$, the assumption constrains how $\eta_k$ scales with $k$ but does not identify its exact value.}
This assumption is the loss-level counterpart of a well-established criterion for learning rate transfer, which chooses the learning rate so that each update changes the model's features by $\Theta(1)$ as the model size scales up~\citep{yang2021tensor,hayou2024impact,hayou2024lora}.
\Cref{fig:theory_profile:a} shows that the per-step values of $\eta\widetilde D_k(\eta)$ at the empirically tuned learning rates remain on a comparable scale across $k$ during training, supporting this assumption. 

\begin{assumption}[Comparable per-block contributions]
\label{asm:theory-contributions}
The nominal contributions satisfy
\begin{equation}
 d_l=\Theta(1),\qquad l\in\gA_k,
 \label{eq:theory-comparable}
\end{equation}
uniformly over trainable blocks in the regime considered.
\end{assumption}
Assumption \ref{asm:theory-contributions} indicates that for each block, as we scale $k$, the contribution of block $l$ to the loss reduction remains stable (not exploding and not vanishing). \Cref{fig:theory_profile:b} provides empirical support for this assumption, showing that the geometric mean of $d_l$ over training steps is approximately invariant to $k$ within each block and comparable across blocks. 

\begin{assumption}[Dependence of the update factor on the learning rate]
\label{asm:theory-realization}
For a fixed $\alpha\ge0$, the update factor satisfies $\phi_k(\eta)=\Theta(\eta^{\alpha})$,
uniformly over the learning rate regime considered.
\end{assumption}
Under FP32, $\phi_k(\eta)=1$, corresponding to $\alpha=0$. 
\arxiv{For BF16, we conjecture that the coordinates with nonzero realized updates make up a fraction proportional to $\eta$ of the trainable coordinates and contribute about as much per coordinate to $\widetilde{D}_k(\eta)$ as the nominal update does to $D_k$. The intuition is that BF16 rounding cancels many terms, and the number of "active" (non canceled) terms grows linearly with the learning rate $\eta$.}
We justify this claim more rigorously in Appendix~\ref{sec:app-theory-bf16-sparsity} and show that it implies that $\phi_k(\eta) = \Theta(\eta)$, and therefore $\alpha=1$. The empirical results in~\Cref{fig:theory_profile:c} show that the curves for different $k$ nearly coincide and grow approximately linearly in $\eta$ before saturation to a constant. We conjecture that the optimal BF16 learning rates lie in this linear regime (in $\eta$) because post-training generally requires small adaptations to the pretrained weights, whereas larger learning rates can destabilize training easily.

\begin{proposition}[Learning rate scaling for general $\alpha$ (informal)]
\label{prop:theory-scaling}
Under the first-order approximation and Assumptions~\ref{asm:theory-stable}--\ref{asm:theory-realization}, the optimal learning rate $\eta_k$  satisfies
\begin{equation}
 \eta_k=\Theta\!\left(k^{-1/(1+\alpha)}\right).
 \label{eq:theory-scaling}
\end{equation}
\end{proposition}

\begin{corollary}[FP32 and BF16 scaling]
\label{cor:theory-precision}
Under the conditions of Proposition~\ref{prop:theory-scaling}, FP32 master-weight updates have $\phi_k(\eta)=1$, corresponding to $\alpha=0$, whereas direct BF16 updates with $\phi_k(\eta)=\Theta(\eta)$ have $\alpha=1$.
Consequently,
\begin{equation}
 \eta_k =
 \begin{cases}
  \Theta(k^{-1}), & \text{FP32 master-weight updates},\\
  \Theta(k^{-1/2}), & \text{direct BF16 updates}.
 \end{cases}
 \label{eq:theory-precision-scaling}
\end{equation}
\end{corollary}
Intuitively, for BF16 updates in the learning rate regime considered, reducing $\eta$ not only shrinks the parameter updates per step but also causes more of them to be suppressed by rounding, making the realized update
sparser. This suppression contributes a further factor of $\eta$ in $\phi_k$ (see Appendix~\ref{sec:app-theory-bf16-sparsity}), changing the realized first-order loss reduction from $\Theta(k\eta)$ to $\Theta(k\eta^2)$. 
Maintaining a stable loss reduction therefore changes the learning rate scaling from $k^{-1}$ to $k^{-1/2}$.
\arxiv{On the other hand, under stochastic rounding, each coordinate is rounded up or down at random so that the realized update is unbiased, which gives $\mathbb{E}[\phi_k(\eta)]=1$ and removes the extra factor of $\eta$.
Direct BF16 updates with stochastic rounding should approximately follow the $k^{-1}$ rule, which we confirm in Appendix~\ref{app:stochastic_rounding}.}

\section{Experiments}
\label{sec:experiments}

We test \method{} learning rate scaling rules with the number of trainable blocks $k$ and measure the GPU-memory and runtime costs of learning rate transfer with caching.\footnote{Source code is available at~\url{https://github.com/NanChanNN/TipSweep}.} 
Our main experiments span multiple model families and two post-training regimes: supervised finetuning (SFT) and direct preference optimization (DPO).
\arxiv{Complementary experiments extend the evaluation to direct FP16 updates and to on-policy RLVR setting.}
Through these experiments, we aim to address three points:
\begin{enumerate}[itemsep=2.2pt, parsep=0pt, topsep=0pt, leftmargin=2.1em, after=\vspace{-2pt}]
    \item[\textit{1.}] Validate \method{} scaling rules  across post-training regimes and model families.
    \item[\textit{2.}] Study the impact of the placement of the trainable blocks on optimal learning rate scaling.
    \item[\textit{3.}] Quantify the GPU-memory and runtime costs associated with \method{}.
\end{enumerate}

\paragraph{Shared post-training protocol.}
Unless otherwise specified, we train with a maximum sequence length of $1024$ tokens, using the AdamW optimizer with zero weight decay and global-norm gradient clipping at $1.0$.
We vary the learning rate schedule across settings, using either a constant learning rate or cosine annealing with linear warmup, to show that the scaling rules hold across schedules.
For each setting, we independently identify the optimal learning rate under each of two criteria: (i) the final training loss, smoothed with an exponential moving average (EMA), and (ii) a task-specific validation metric.
We first sweep $\eta$ on a coarse $\log_2$-spaced grid with multiplicative steps of $2\times$.
We then refine around the coarse-grid rate with the lowest final EMA-smoothed training loss, evaluating two additional rates at offsets of $\pm0.5$ in $\log_2\eta$.
This local refinement improves resolution around the optimum without the cost of evaluating a uniformly finer grid.
We report training-loss-based sweeps in the main text and defer the other results to Appendix~\ref{app:additional_results}.
Setup details are in Appendix~\ref{app:experimental_setup}.

\paragraph{Interpreting the identified learning rates.}
The reported optimum for each metric is the best value among the evaluated grid points.
Near a flat optimum, neighboring learning rates can perform similarly, so run-to-run noise may determine which one is selected.
Finer grids could make selection more sensitive to these near-ties, and update precision might also affect the flatness of the performance curve.
We therefore assess qualitative agreement with the prescribed scaling rules rather than require an exact numerical match.

\begin{figure}[t]
\centering
\includegraphics[width=\linewidth]{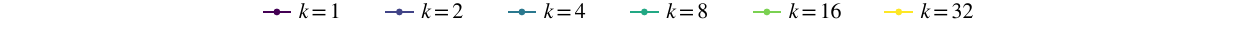}\par\nointerlineskip
\begin{subfigure}[t]{0.333333\linewidth}\captionsetup{margin={18.72pt,0pt}}
  \includegraphics[width=\linewidth]{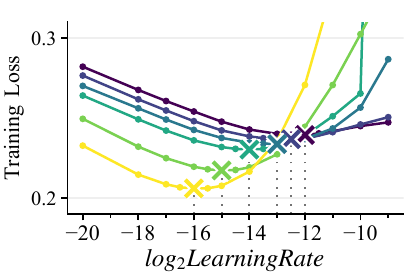}
  \caption{DSCoder-6.7B (SFT, FP32)}
  \label{fig:exp_lr_sweep_fp32_bf16:a}
\end{subfigure}%
\begin{subfigure}[t]{0.333333\linewidth}\captionsetup{margin={18.68pt,0pt}}
  \includegraphics[width=\linewidth]{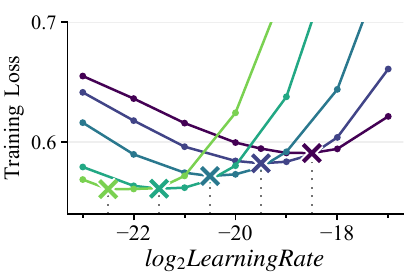}
  \caption{Llama-3.2-1B (DPO, FP32)}
\end{subfigure}%
\begin{subfigure}[t]{0.333333\linewidth}\captionsetup{margin={18.72pt,0pt}}
  \includegraphics[width=\linewidth]{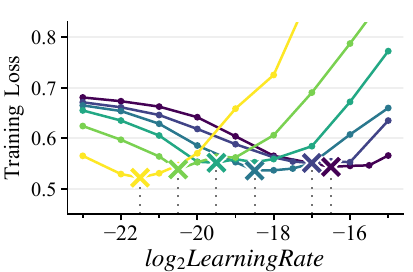}
  \caption{OLMo-2-7B (DPO, FP32)}
  \label{fig:exp_lr_sweep_fp32_bf16:c}
\end{subfigure}%
\\
\begin{subfigure}[t]{0.333333\linewidth}\captionsetup{margin={18.72pt,0pt}}
  \includegraphics[width=\linewidth]{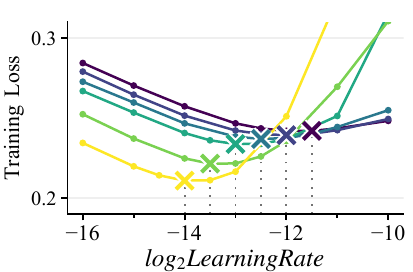}
  \caption{DSCoder-6.7B (SFT, BF16)}
  \label{fig:exp_lr_sweep_fp32_bf16:d}
\end{subfigure}%
\begin{subfigure}[t]{0.333333\linewidth}\captionsetup{margin={18.68pt,0pt}}
  \includegraphics[width=\linewidth]{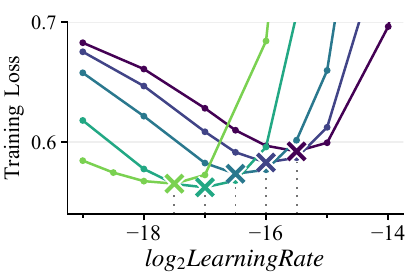}
  \caption{Llama-3.2-1B (DPO, BF16)}
\end{subfigure}%
\begin{subfigure}[t]{0.333333\linewidth}\captionsetup{margin={18.72pt,0pt}}
  \includegraphics[width=\linewidth]{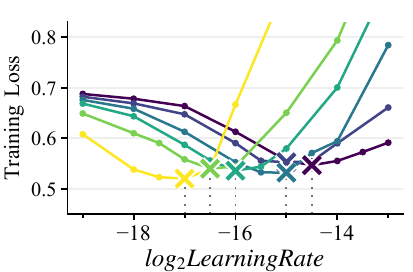}
  \caption{OLMo-2-7B (DPO, BF16)}
  \label{fig:exp_lr_sweep_fp32_bf16:f}
\end{subfigure}%
\caption{
Learning rate sweeps with FP32 master-weight (top) and direct BF16 updates (bottom).
(a, d) SFT of DeepSeek-Coder-6.7B-Base on Magicoder-OSS-Instruct;
(b, e) DPO of Llama-3.2-1B-Instruct on UltraFeedback;
(c, f) DPO of OLMo-2-1124-7B-SFT on its official preference mixture.
Each curve shows final training loss (EMA-smoothed) versus $\log_2$ learning rate with only the last $k$ blocks trainable. 
Crosses and dotted lines mark each curve's optimal learning rate.
}
\label{fig:exp_lr_sweep_fp32_bf16}
\end{figure}

\begin{figure}[t]
\centering
\captionsetup[subfigure]{justification=centering}
\hspace*{-12pt}%
\begin{subfigure}[t]{\dimexpr0.243625\linewidth+24pt}\captionsetup{margin={20.88pt,0pt}}
  \centering\includegraphics[width=\dimexpr\linewidth-24pt]{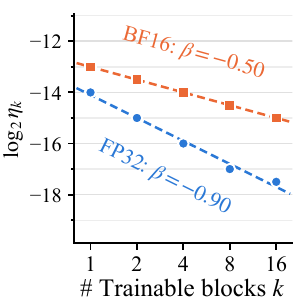}
  \caption{Llama-3.2-1B (SFT)}
\end{subfigure}%
\hspace*{-24pt}%
\begin{subfigure}[t]{\dimexpr0.264632\linewidth+24pt}\captionsetup{margin={14.34pt,0pt}}
  \centering\includegraphics[width=\dimexpr\linewidth-24pt]{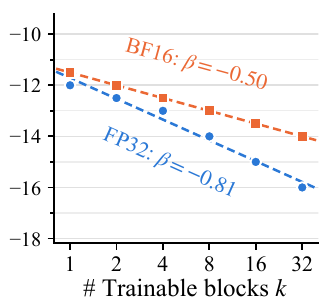}
  \caption{DSCoder-6.7B (SFT)}
\end{subfigure}%
\hspace*{-24pt}%
\begin{subfigure}[t]{\dimexpr0.227110\linewidth+24pt}\captionsetup{margin={14.34pt,0pt}}
  \centering\includegraphics[width=\dimexpr\linewidth-24pt]{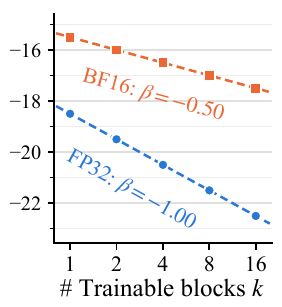}
  \caption{Llama-3.2-1B (DPO)}
\end{subfigure}%
\hspace*{-24pt}%
\begin{subfigure}[t]{\dimexpr0.264632\linewidth+24pt}\captionsetup{margin={14.34pt,0pt}}
  \centering\includegraphics[width=\dimexpr\linewidth-24pt]{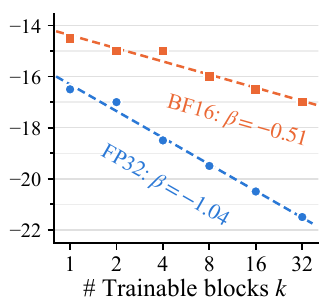}
  \caption{OLMo-2-7B (DPO)}
\end{subfigure}%
\hspace*{-12pt}
\caption{\arxiv{Optimal learning rate against the number of trainable blocks $k$ under FP32 master-weight (blue, circles) and direct BF16 (orange, squares) updates: a) SFT of Llama-3.2-1B on \tuluthree{}; b) SFT of DeepSeek-Coder-6.7B-Base on Magicoder-OSS-Instruct; c) DPO of Llama-3.2-1B-Instruct on UltraFeedback; d) DPO of OLMo-2-1124-7B-SFT on its official preference mixture. Each marker is the learning rate $\eta_k$ that minimizes the final training loss (EMA-smoothed) with only the last $k$ blocks trainable; the Llama models have 16 blocks, so a) and c) end at $k=16$. Dashed lines are least-squares fits of $\log_2 \eta_k$ against $\log_2 k$, each labelled with its fitted slope $\beta$, i.e., $\eta_k \propto k^{\beta}$, to be compared with $\beta=-1$ under FP32 and $\beta=-1/2$ under direct BF16 updates (see \Cref{eq:theory-precision-scaling}).}}
\label{fig:exp_depth_lr_fit}
\end{figure}

\subsection{Precision-Dependent Scaling across the Number of Trainable Blocks}
\label{sec:experiments:scaling}

\paragraph{Setup.}
For SFT tasks, we train Llama-3.2-1B~\citep{grattafiori2024llama} on \tuluthree{}~\citep{lambert2025tulu} and DeepSeek-Coder-6.7B-Base~\citep{guo2024deepseek} on Magicoder-OSS-Instruct~\citep{wei2024magicoder}.
For DPO tasks, we train Llama-3.2-1B-Instruct on UltraFeedback~\citep{cui2023ultrafeedback} using the standard DPO loss with $\beta=0.1$.
We also finetune OLMo-2-1124-7B-SFT on its official preference mixture~\citep{walsh2025} using a length-normalized DPO loss with $\beta=5.0$.
This loss normalizes each response's policy--reference log-ratio by its length before applying the log-sigmoid.
For this setting, we increase the maximum sequence length to $2048$ tokens.
Across the four settings, model depth is $L=16$ for the Llama models and $L=32$ for DeepSeek-Coder and the OLMo model.

\paragraph{Main results.}
The learning rate that minimizes the final training loss scales approximately as $k^{-1}$ under FP32 master-weight updates (Figures~\ref{fig:intro_motivations:tf32} and~\ref{fig:exp_lr_sweep_fp32_bf16:a}--\ref{fig:exp_lr_sweep_fp32_bf16:c}) and as $k^{-1/2}$ under direct BF16 updates (Figures~\ref{fig:intro_motivations:bf16} and~\ref{fig:exp_lr_sweep_fp32_bf16:d}--\ref{fig:exp_lr_sweep_fp32_bf16:f}).
~\Cref{fig:exp_depth_lr_fit} quantifies both trends by plotting the optimal learning rate against $k$ and reporting the scaling exponents from log--log fits.
When finetuning more blocks, the loss also rises more sharply once the learning rate exceeds the optimum, supporting the need to rescale the learning rate rather than reuse the proxy rate unchanged.
\arxiv{An ablation with the gradients and optimizer states kept in FP32 still yields $k^{-1/2}$ scaling under direct BF16 updates (\Cref{fig:lr_sweeps_bf16_fp32optim}), showing that the change from $k^{-1}$ to $k^{-1/2}$ comes from the update precision itself.}
Update precision affects the results in two further ways.
First, for a given setting and $k$, the optimal learning rate is lower under FP32 than under BF16, especially for DPO tasks.
This is consistent with update rounding: BF16 suppresses small parameter increments, so larger learning rates are needed to produce effective weight changes.
Second, the FP32 curves are flatter near their minima, particularly for the 7B-scale models at small $k$, so neighboring learning rates yield similar losses and the selected grid point becomes more sensitive to run-to-run noise.
More results are in Appendix~\ref{app:additional_results:scaling}, including ablations finetuning only self-attention or MLP modules in each trainable block.

\begin{figure}[t]
\centering
\includegraphics[width=\linewidth]{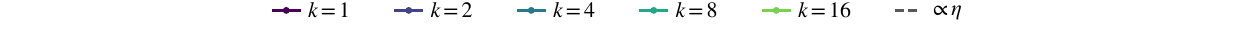}\par\nointerlineskip
\begin{subfigure}[t]{0.333333\linewidth}\captionsetup{margin={18.72pt,0pt}}
  \includegraphics[width=\linewidth]{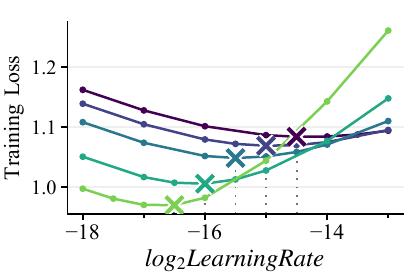}
  \caption{Llama-3.2-1B (SFT, FP16)}
  \label{fig:exp_lr_sweep_fp16:a}
\end{subfigure}%
\begin{subfigure}[t]{0.333333\linewidth}\captionsetup{margin={18.68pt,0pt}}
  \includegraphics[width=\linewidth]{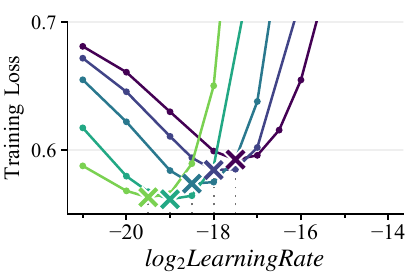}
  \caption{Llama-3.2-1B (DPO, FP16)}
  \label{fig:exp_lr_sweep_fp16:b}
\end{subfigure}%
\begin{subfigure}[t]{0.333333\linewidth}\captionsetup{margin={18.72pt,0pt}}
  \includegraphics[width=\linewidth]{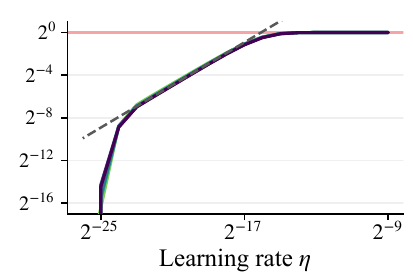}
  \caption{$\phi_k(\eta)=\widetilde D_k(\eta)/D_k$}
  \label{fig:exp_lr_sweep_fp16:c}
\end{subfigure}%
\caption{
\arxiv{Learning rate sweeps and FP16 update factor $\phi_k(\eta)$ with direct FP16 updates. 
a) SFT of Llama-3.2-1B on \tuluthree{}. 
b) DPO of Llama-3.2-1B-Instruct on UltraFeedback.
In a) and b), each curve shows final training loss (EMA-smoothed) versus $\log_2$ learning rate with only the last $k$ blocks trainable.
Crosses and dotted lines mark each curve's optimal learning rate.
c) FP16 update factor $\phi_k(\eta)$ against the probe learning rate $\eta$ at step 3000 for SFT of Llama-3.2-1B on \tuluthree{}. }
}
\label{fig:exp_lr_sweep_fp16}
\end{figure}

\paragraph{Extension to FP16.}
\arxiv{We conduct experiments with direct FP16 updates, in which the weights are stored in float16 without a float32 master copy.
Compared with BF16, FP16 has higher precision but a narrower dynamic range.
Despite these differences, the optimal learning rate scales approximately as $k^{-1/2}$ for both SFT and DPO (\Cref{fig:exp_lr_sweep_fp16:a,fig:exp_lr_sweep_fp16:b}), as it does under direct BF16 updates.
The update factor $\phi_k(\eta)$ in \Cref{fig:exp_lr_sweep_fp16:c} explains this pattern.
It again grows approximately linearly with $\eta$ before saturating, but saturation now occurs at learning rates roughly $2^3$ times smaller than under BF16 (\Cref{fig:theory_profile:c}), consistent with the three additional mantissa bits of FP16.
The optimal learning rates also fall within the linear regime, and the largest one, at $k=1$ under SFT, lies near the transition to saturation.
At extremely small learning rates, the narrower dynamic range causes updates to underflow and $\phi_k(\eta)$ to drop sharply to zero.
The optimizer states also underflow in FP16 storage, so we store them in FP32 to ensure stable training in this experiment.}


\paragraph{Extension to Reinforcement Learning with Verifiable Rewards (RLVR)}
\arxiv{
To examine whether the scaling rules extend to on-policy RLVR setting, we train Llama-3.1-8B ($L=32$)~\citep{grattafiori2024llama} on GSM8K~\citep{cobbe2021gsm8k} using GRPO~\citep{shao2024deepseekmath} with the token-level DAPO loss~\citep{yu2025dapo}.
Correct answers receive a reward, with an additional bonus for the requested format.
Training runs for $100$ optimizer steps, with $128$ problems and $16$ sampled completions per problem at each step.
Further setup details are provided in Appendix~\ref{app:exp_setup_rlvr}.
}

\arxiv{
\Cref{fig:exp_lr_sweep_gsm8k} shows that the optimal learning rate scales approximately as $k^{-1}$ under FP32 master-weight updates and as $k^{-1/2}$ under direct BF16 updates, with fitted exponents of $-0.80$ and $-0.57$, respectively.
For $k \in \{2,4,8,16\}$, the FP32 reward curves exhibit broader near-optimal plateaus than their BF16 counterparts, which makes the selected grid point more sensitive to noise (e.g., at $k=16$) and slightly flattens the fitted FP32 slope.
Furthermore, both precisions reach comparable peak training rewards at their corresponding optimal learning rates across all $k$, indicating that direct BF16 updates remain a viable memory-saving option for on-policy RLVR.
}

\begin{figure}[t]
\centering
\includegraphics[width=\linewidth]{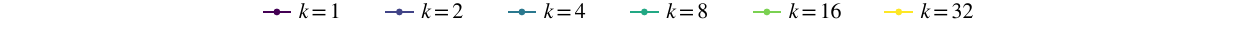}\par\nointerlineskip
\begin{subfigure}[t]{0.333333\linewidth}\captionsetup{margin={18.72pt,0pt}}
  \includegraphics[width=\linewidth]{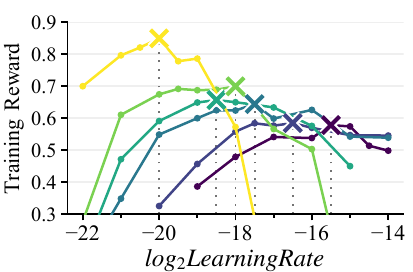}
  \caption{Float32 weights}
\end{subfigure}%
\begin{subfigure}[t]{0.333333\linewidth}\captionsetup{margin={18.68pt,0pt}}
  \includegraphics[width=\linewidth]{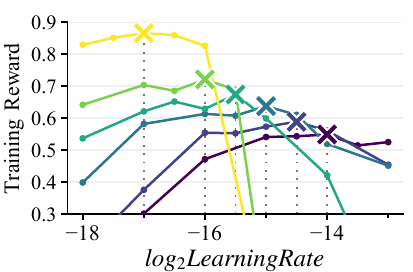}
  \caption{BFloat16 weights}
\end{subfigure}%
\begin{subfigure}[t]{0.333333\linewidth}\captionsetup{margin={20.84pt,0pt}}
  \includegraphics[width=\linewidth]{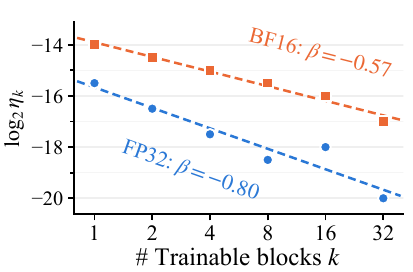}
  \caption{Optimal learning rate vs.\ $k$}
\end{subfigure}%
\caption{
\arxiv{Learning rate sweeps for RLVR of Llama-3.1-8B (32 layers) on GSM8K.
(a, b) FP32 master weight and direct BF16 updates, respectively.
Each curve shows final training reward (EMA-smoothed) versus $\log_2$ learning rate with only the last $k$ blocks trainable.
Crosses and dotted lines mark each curve's optimal learning rate, corresponding to the highest reward.
c) Optimal learning rate against the number of trainable blocks $k$ under FP32 master-weight (blue, circles) and direct BF16 (orange, squares) updates.
Dashed lines are least-squares fits of $\log_2\eta_k$ against $\log_2 k$.}}
\label{fig:exp_lr_sweep_gsm8k}
\end{figure}

\subsection{Effect of Low-Rank Adaptation (LoRA)}
\label{sec:experiments:lora}

\paragraph{Setup.}
\arxiv{We next study how the optimal learning rate scales with $k$ under LoRA~\citep{hu2022lora}, which freezes the pretrained weights and trains a low-rank update $BA$ in each of the last $k$ blocks.
We finetune Llama-3.2-1B on \tuluthree{}.
By default, LoRA stores and updates the adapter weights in FP32, and we also consider direct BF16 updates of the adapter weights, without an FP32 master copy.
Both settings adopt Init[A] (random $A$, $B=0$)~\citep{hayou2024impact} with rank $128$.
We additionally consider Init[B] (random $B$, $A=0$), for which the learning rate tuned with LoRA is expected to transfer to full finetuning~\citep{chen2026learning}.}

\begin{figure}[t]
\centering
\includegraphics[width=\linewidth]{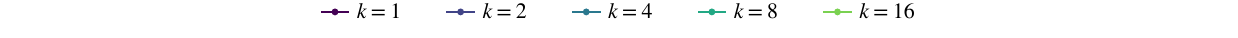}\par\nointerlineskip
\begin{subfigure}[t]{0.333333\linewidth}\captionsetup{margin={18.72pt,0pt}}
  \includegraphics[width=\linewidth]{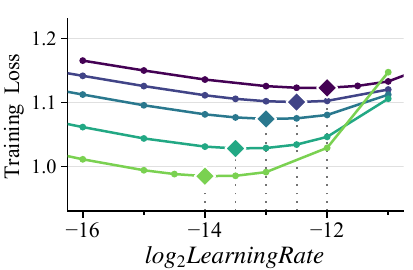}
  \caption{LoRA Init[A] (FP32)}
  \label{fig:exp_lr_sweep_lora:a}
\end{subfigure}%
\begin{subfigure}[t]{0.333333\linewidth}\captionsetup{margin={18.68pt,0pt}}
  \includegraphics[width=\linewidth]{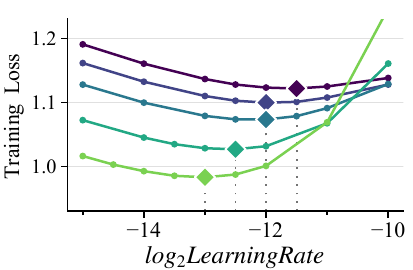}
  \caption{LoRA Init[A] (BF16)}
  \label{fig:exp_lr_sweep_lora:b}
\end{subfigure}%
\begin{subfigure}[t]{0.333333\linewidth}\captionsetup{margin={18.72pt,0pt}}
  \includegraphics[width=\linewidth]{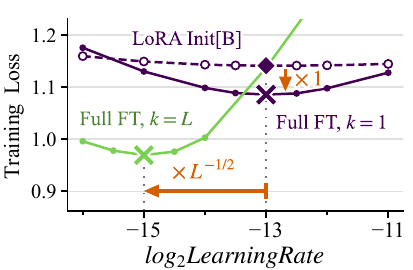}
  \caption{LoRA to full finetuning}
  \label{fig:exp_lr_sweep_lora:c}
\end{subfigure}%
\caption{
\arxiv{Learning rate sweeps with LoRA for SFT of Llama-3.2-1B on \tuluthree{}.
(a, b) Init[A] on the last $k$ blocks with FP32 and BF16 adapter weights, respectively.
c) Init[B] on the last block (dashed) and the $k=1$ and $k=L$ sweeps of \Cref{fig:intro_motivations:bf16} (solid); arrows show the LoRA learning rate transferred unchanged to $k=1$ and scaled by $L^{-1/2}$ to $k=L$.
Each curve shows final training loss (EMA-smoothed) versus $\log_2$ learning rate.
Diamonds and dotted lines mark each curve's optima.}
}
\label{fig:exp_lr_sweep_lora}
\end{figure}

\paragraph{Results.}
\arxiv{
\Cref{fig:exp_lr_sweep_lora:a,fig:exp_lr_sweep_lora:b} show that LoRA with Init[A] scales differently with $k$ than full finetuning.
The optimal learning rate still decreases as $k$ grows, but with FP32 adapter weights it scales approximately as $k^{-1/2}$ instead of $k^{-1}$.
With BF16 adapter weights, it decreases at a similar or slightly slower pace.
We hypothesize that this difference from full finetuning arises because LoRA changes each weight matrix through the product of two trainable factors rather than by a direct update, and we leave the analysis to future work.
Nevertheless, LoRA can serve as a proxy for learning rate search.
In \Cref{fig:exp_lr_sweep_lora:c}, LoRA with Init[B] on the last block has the same optimal learning rate as full finetuning of that block, in line with \citet{chen2026learning}.
Scaling this rate by $L^{-1/2}$, as \method{} prescribes for direct BF16 updates, gives the optimal learning rate of the full-depth run.
A learning rate tuned with LoRA on the last block can therefore be reused for full finetuning of that block and then rescaled for the full-depth run, which can further reduce the memory needed for the search.
}

\begin{figure}[t]
\centering
\begin{subfigure}[t]{0.333333\linewidth}
  \includegraphics[width=\linewidth]{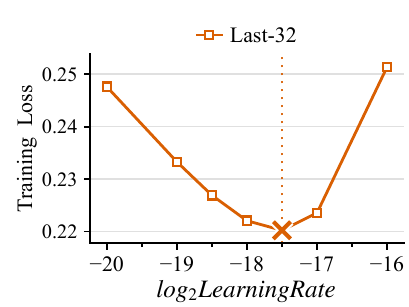}
\end{subfigure}\hfill
\begin{subfigure}[t]{0.333333\linewidth}
  \includegraphics[width=\linewidth]{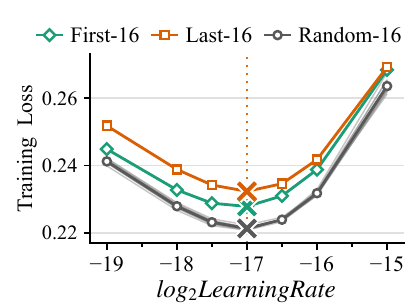}
\end{subfigure}\hfill
\begin{subfigure}[t]{0.333333\linewidth}
  \includegraphics[width=\linewidth]{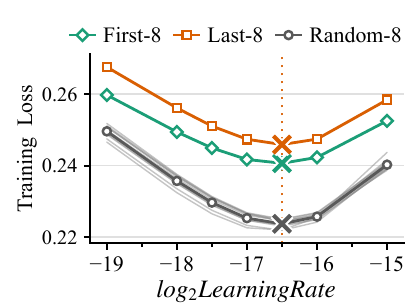}
\end{subfigure}
\par\medskip
\begin{subfigure}[t]{0.333333\linewidth}
  \includegraphics[width=\linewidth]{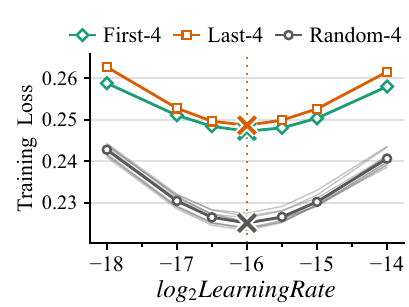}
\end{subfigure}\hfill
\begin{subfigure}[t]{0.333333\linewidth}
  \includegraphics[width=\linewidth]{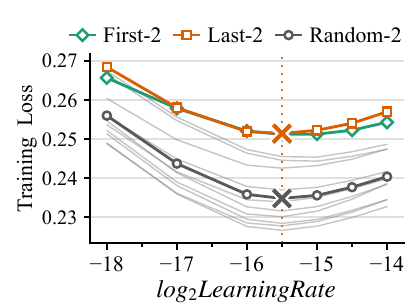}
\end{subfigure}\hfill
\begin{subfigure}[t]{0.333333\linewidth}
  \includegraphics[width=\linewidth]{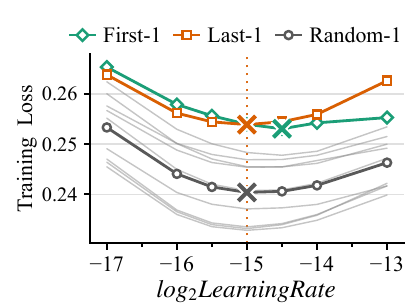}
\end{subfigure}
\caption{
Learning rate sweeps across trainable-block placements. 
Each panel shows final training loss (EMA-smoothed) versus $\log_2$ learning rate for SFT of Mistral-7B-v0.1 under direct BF16 updates, with $k = 32, 16, \dots, 1$ trainable blocks.
For random-$k$, thin gray curves show the sampled subsets for each $k$, and the bold curve denotes their average.
Crosses mark the optimal learning rate.
}
\label{fig:layer_placement}
\end{figure}

\subsection{Effect of Trainable-Block Placement}
\label{sec:experiments:placement}

\paragraph{Setup.}
Our analysis extends to any subset of $k$ trainable blocks under the same assumptions, since the scaling argument sums comparable per-block contributions and does not depend on block positions.
To test this prediction empirically, we compare three ways of choosing which $k$ blocks to train: the last $k$ blocks (\emph{last-$k$}), the first $k$ blocks (\emph{first-$k$}), and a random subset of $k$ blocks (\emph{random-$k$}).
For \emph{random-$k$}, we sample 10 distinct subsets for each $k<L$.
In this part, we finetune Mistral-7B-v0.1 ($L=32$)~\citep{jiang2023mistral7b} on Magicoder-OSS-Instruct, under direct BF16 weight updates.

\paragraph{Results.}
\Cref{fig:layer_placement} shows that, for each $k$, the three placement strategies reach their minimum training loss at roughly the same learning rate.
As $k$ increases from $1$ to $32$, the optimal rate scales approximately as $k^{-1/2}$, consistent with the scaling rules for direct BF16 updates.
Together, these observations suggest that the optimal learning rate depends on the number of trainable blocks but not on their placement.
Although the optimal rate is nearly the same across placements, the final training loss varies substantially.
Specifically, random-$k$ generally yields the lowest loss while last-$k$ gives the highest.
The spread of final training loss across random subsets also tends to widen as $k$ decreases.
\Cref{fig:layer_distribution} shows the positions of the trainable blocks within each sampled subset, with subsets ordered by final training loss.
For $1<k$, we find that subsets whose trainable blocks are spread more evenly across the model tend to obtain lower loss.
For $k=1$, training a block near the middle of the model tends to yield better performance than training one near the input or output.

\subsection{Case study}
\label{sec:experiments:case_study}


\begin{figure}[!t]
\centering
\begin{subfigure}[t]{0.333333\linewidth}\captionsetup{margin={18.72pt,0pt}}
  \includegraphics[width=\linewidth]{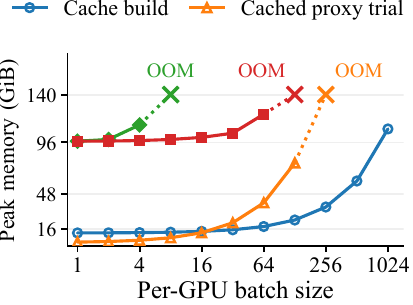}
  \caption{Peak GPU Memory, FP32}\label{fig:main:case_study:memory_fp32}
\end{subfigure}%
\begin{subfigure}[t]{0.333333\linewidth}\captionsetup{margin={18.72pt,0pt}}
  \includegraphics[width=\linewidth]{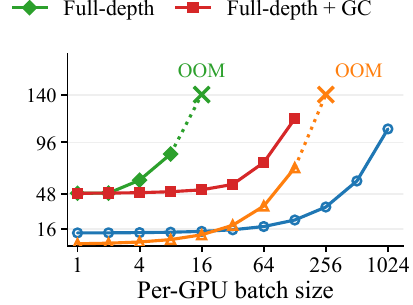}
  \caption{Peak GPU Memory, BF16}\label{fig:main:case_study:memory_bf16}
\end{subfigure}%
\begin{subfigure}[t]{0.333333\linewidth}\captionsetup{margin={18.72pt,0pt}}
  \includegraphics[width=\linewidth]{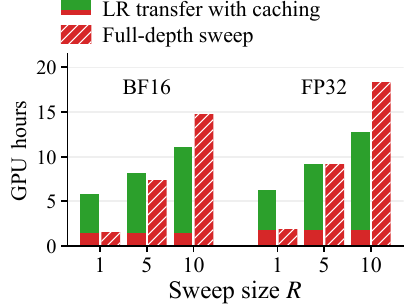}
  \caption{GPU hours vs.\ sweep size}\label{fig:main:case_study:hours}
\end{subfigure}
\caption{a) and b) depict peak allocated GPU memory of the two-stage prototype against per-GPU batch size under FP32 and BF16 weights, respectively. 
c) GPU hours for a search over $R$ candidate learning rates under each precision: cache build and $R$ proxy trials on an RTX 5060 Ti (green) plus one full-depth run on H200 (red), versus $R$ full-depth runs on H200 (red, hatched).}
\label{fig:main:case_study}
\end{figure}

\paragraph{Setup.}
\arxiv{
To measure the peak GPU memory and runtime costs, we implement two prototypes of TipSweep and evaluate them by finetuning DeepSeek-Coder-6.7B-Base on Magicoder-OSS-Instruct.
We set $k_0=1$, so each trial trains only the last block, and compare against a full-depth sweep that trains all blocks at every candidate rate.
The two prototypes differ in how they avoid recomputing the frozen prefix for every candidate rate.
The \emph{two-stage prototype} follows~\Cref{sec:methodology:transfer}: a cache build stores the prefix outputs on disk, and the cached proxy trials then run one candidate at a time.
The \emph{cache-free prototype} instead keeps $R$ copies of the trainable blocks in GPU memory, one per candidate rate, and feeds the prefix output of each batch to all of them, so a single pass over the dataset completes every proxy trial.
The first keeps the GPU memory footprint small enough for a consumer GPU at the cost of disk storage.
The second needs no disk cache and finishes the search in one pass, at the cost of GPU memory that grows with $R$.
Details are in Appendices~\ref{app:case_study} and~\ref{app:cache_free}.
}


\paragraph{Two-stage prototype.}
\arxiv{
\Cref{fig:main:case_study:memory_fp32,fig:main:case_study:memory_bf16} report the peak allocated GPU memory of the two-stage prototype as the per-GPU batch size grows from $1$ to $1024$.
To evaluate worst-case memory use, we construct training batches in which every sequence contains exactly $1024$ tokens.
At small batch sizes, both stages use substantially less memory than full-depth training, particularly with FP32 master weights.
This is because full-depth training stores gradients and optimizer states for all $L$ blocks even at a batch size of $1$, whereas the cache build requires neither and the proxy trials store them for one block.
As the batch size grows, gradient checkpointing slows the growth of memory in full-depth training at the cost of extra recomputation.
The proxy trials do not use checkpointing, so their memory curve is steeper.
Nonetheless, they use less memory than a checkpointed full-depth run up to a batch size of $128$ under both precisions.
}

\arxiv{
We deploy the prototype on a single NVIDIA RTX 5060 Ti with $15.45$\,GiB of available GPU memory and measure its runtime.
Since full-depth training needs more memory than this GPU provides even at a batch size of $1$, we run it on NVIDIA H200 GPUs, for the final full-depth run of our method and for every trial of the full-depth sweep.
\Cref{fig:main:case_study:hours} reports the resulting GPU hours by hardware tier against the sweep size $R$.
With $R=5$, LR transfer with caching uses roughly the same number of GPU hours as the full-depth sweep under both precisions.
However, it spends about $80\%$ of these hours on a consumer GPU with lower capacity, searching for the learning rate on a small proxy, and employs the H200 only for the final full-depth run, whereas the full-depth sweep occupies the high-capacity GPU throughout.
As $R$ grows, this advantage widens, because the H200 hours of our method stay fixed at one full-depth run while those of the full-depth sweep grow linearly with $R$.}

\begin{figure}[!t]
\centering
\begin{subfigure}[t]{0.333333\linewidth}\captionsetup{margin={18.72pt,0pt}}
  \includegraphics[width=\linewidth]{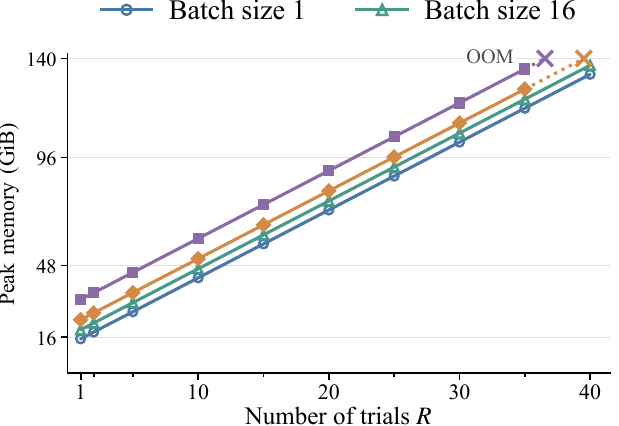}
  \caption{Peak GPU Memory, FP32}\label{fig:main:case_study_cache_free:memory_fp32}
\end{subfigure}%
\begin{subfigure}[t]{0.333333\linewidth}\captionsetup{margin={18.72pt,0pt}}
  \includegraphics[width=\linewidth]{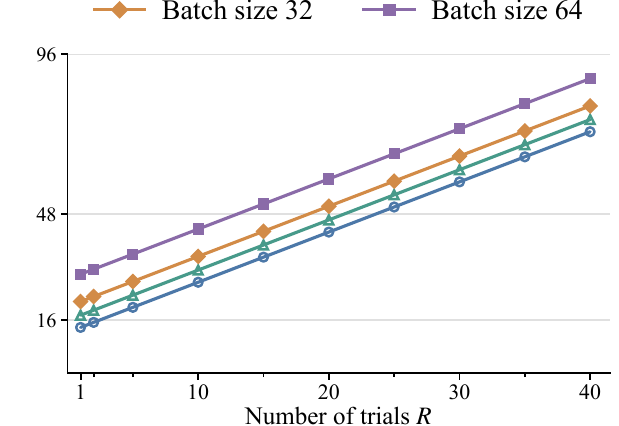}
  \caption{Peak GPU Memory, BF16}\label{fig:main:case_study_cache_free:memory_bf16}
\end{subfigure}%
\begin{subfigure}[t]{0.333333\linewidth}\captionsetup{margin={18.72pt,0pt}}
  \includegraphics[width=\linewidth]{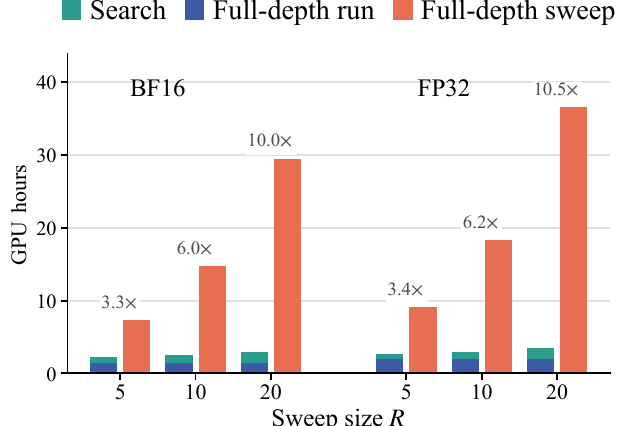}
  \caption{GPU hours vs. sweep size}\label{fig:main:case_study_cache_free:running_time}
\end{subfigure}
\caption{\arxiv{a) and b) depict peak allocated GPU memory of the cache-free prototype against the number of trials $R$ at four per-GPU batch sizes, under FP32 and BF16 weights, respectively.
c) H200 GPU hours for a search over $R$ candidate learning rates under each precision: one search pass that tunes $R$ copies of the last block (green) plus one full-depth run (blue), versus the full-depth sweep (orange). The number above each pair of bars is the speed-up over the full-depth sweep.}}
\label{fig:main:case_study_cache_free}
\end{figure}

\paragraph{Cache-free prototype.}
\arxiv{
The two-stage prototype needs disk storage for its feature cache, which grows with the size of the training set ($336.8$\,GiB here), and it leaves open how TipSweep can take advantage of high-end GPUs.
The cache-free prototype addresses both, at the cost of GPU memory that grows with the sweep size $R$.
\Cref{fig:main:case_study_cache_free:memory_fp32,fig:main:case_study_cache_free:memory_bf16} report its peak allocated GPU memory on a single H200 against the number of trials $R$, for per-GPU batch sizes of $1$, $16$, $32$, and $64$ and the same worst-case batches as above.
Peak memory grows by $1.51$\,GiB per additional trial under BF16 and by $3.02$\,GiB under FP32, at every batch size.
In both cases, the increment matches the memory of four block-sized tensors, which hold each copy's weights, gradients, and AdamW first- and second-moment estimates.
Activations do not add to the increment because the copies run sequentially within each update step, so the GPU holds the activations of only one copy at a time.
Under BF16, all tested configurations fit in the H200 GPU, including $40$ trials at a per-GPU batch size of $64$, which peaks at $88.6$\,GiB.
Under FP32, the H200 holds up to $35$ trials at that batch size.}

\arxiv{
Unlike the two-stage prototype, the cache-free prototype runs both the search and the final full-depth run on a single H200.
\Cref{fig:main:case_study_cache_free:running_time} compares its GPU hours, including the final full-depth run, with those of the full-depth sweep.
The search itself is cheap.
Even with $20$ candidates, it takes no longer than a single full-depth run under both precisions.
Its time grows linearly with $R$, with a fixed cost of about half an hour for the forward pass through the frozen prefix and about $3$ minutes per additional candidate.
In the full-depth sweep, each additional candidate instead requires one more full-depth run, at roughly $30$ times the cost.
Consequently, the speed-up over the full-depth sweep grows with $R$, from about $3\times$ at $R=5$ to about $10\times$ at $R=20$ under both precisions.}

\begin{table}[!t]
\centering
\caption{Pass@1 (\%, higher is better) on HumanEval and HumanEval+ for DeepSeek-Coder-6.7B-Base and its finetuned variants. Full-depth trains all 32 transformer blocks and Full-parameter also trains the embeddings and LM head. Both are trained at the learning rate transferred from a $k_0=1$ proxy search. The Base model and Magicoder-DS rows are reported by \citet{wei2024magicoder}.}
\label{tab:case_study:main_quality}
{\normalsize\setlength{\tabcolsep}{4pt}
\begin{tabular*}{0.99\linewidth}{@{\extracolsep{\fill}}llcccc@{}}
\toprule
Model / scope & Precision & Base LR & Transferred LR & HumanEval & HumanEval+ \\
\midrule
DeepSeek-Coder-6.7B-Base & --- & --- & --- & 47.6 & 39.6 \\
Magicoder-DS & FP32 & --- & --- & \textbf{66.5} & \textbf{60.4} \\
\midrule
Full-depth (ours) & BF16 & $2^{-11.5}$ & $2^{-14}$ & 64.0 & 57.3 \\
Full-parameter (ours) & BF16 & $2^{-11.5}$ & $2^{-14}$ & 64.6 & 59.1 \\
Full-depth (ours) & FP32 & $2^{-12}$ & $2^{-17}$ & 64.0 & 57.9 \\
Full-parameter (ours) & FP32 & $2^{-12}$ & $2^{-17}$ & 65.2 & 59.8 \\
\bottomrule
\end{tabular*}}
\end{table}

\paragraph{Quality of the transferred learning rate.}
We assess the quality of the learning rate transferred from the proxy search in full-depth training.
We also include a \emph{full-parameter} variant that additionally trains the embeddings and the LM head at the same rate.
We train both variants under each precision and evaluate their final checkpoints on HumanEval and HumanEval+.
\Cref{tab:case_study:main_quality} reports pass@1 alongside two reference values from \citet{wei2024magicoder}: the base model without finetuning, and Magicoder-DS, the released model finetuned on the same dataset.
At the transferred rates, both variants perform competitively with Magicoder-DS under both precisions.
Our training recipe differs from that of Magicoder-DS in epochs, optimizer, and batch size, so this comparison does not isolate the learning rate choice, but shows that the transferred rates yield competitive checkpoints under our training protocol.
BF16 and FP32 also perform comparably for each variant and benchmark, which indicates that direct BF16 updates are a safe choice for saving memory with little loss in performance.
In addition, full-parameter training outperforms full-depth training at the same rate.
Although our analysis covers only the transformer blocks, this result shows that the transferred rates remain effective when all parameters are trained, without separate tuning.
\arxiv{In Appendix~\ref{app:multi_seed}, we repeat the learning rate transfer on additional seeds and find the optimal learning rate in the proxy trials is consistent across seeds and the transferred rate continues to deliver a competitive performance.}

\section{Conclusion and Limitations}
We propose learning rate transfer across the number of trainable blocks for LLM post-training and find that the optimal rate scales roughly as $k^{-1}$ under FP32 master-weight updates and as $k^{-1/2}$ under direct BF16 updates.
Extensive experiments across model families, datasets, and post-training settings support these scaling rules.
\arxiv{Moreover, we implement two instantiations of our proposed learning rate search for a 6.7B model and demonstrate that it is feasible on a single consumer GPU with approximately 16~GiB of memory, and that on a single high-end GPU it can reduce GPU hours by about $10\times$ relative to a full-depth sweep over 20 candidate learning rates.}

Our work also has limitations that warrant further investigation.
First, we study only learning rate transfer in this work.
Whether other hyperparameters, such as weight decay and batch size, transfer similarly requires further investigation.
Our results nevertheless identify the number of trainable blocks as a promising axis for accelerating hyperparameter search.
\arxiv{Second, both prototypes require a fixed offline dataset.}
\arxiv{In on-policy training, such as RLVR and on-policy distillation (OPD), the evolving policy of each trial generates its own completions during training, so their features can neither be cached once and reused across trials nor shared among trials within a pass.}
\arxiv{Each proxy trial must then recompute the frozen prefix, limiting the analytical FLOP reduction to less than $3\times$ relative to a full-depth trial.}
\arxiv{Our RLVR experiments support the scaling rules, but achieving comparable search savings in on-policy training remains an open direction.}
Third, our theoretical analysis relies on assumptions supported by profiling experiments.
A more rigorous treatment is needed to establish when these assumptions hold and when the resulting scaling rules apply.
Nonetheless, our findings highlight update rounding as a key factor shaping learning rate scaling under real-world mixed-precision training, pointing out a possible direction for further theoretical study.


\clearpage

\subsection*{AI use statement}

In this work, we used general-purpose LLM assistants to discuss and refine theoretical arguments, assumptions, hypotheses, and proof drafts, and to polish mathematical notation.
We also sought feedback on research methodology, experiments, and the interpretation of results.
We did not use LLM assistants to implement our method.
Generating synthetic datasets, translation, cleaning or reformatting datasets, and qualitative or thematic data analysis are not applicable to this work.
Additionally, we used LLM assistants to write plotting code for scientific figures, edit the paper for readability, and identify relevant literature.
We reviewed all AI-assisted work and checked the mathematical claims and proof arguments.
We verified that the plotted values match our experimental results and that the plotting code does not modify the underlying data.
Every AI-suggested citation was checked against the original paper.
We take responsibility for the final content of this work, including text, claims, and artifacts produced with the aid of generative AI.

\subsection*{Ethics statement}

All authors have read and adhere to the ICLR Code of Ethics. 
This work studies learning rate scaling for LLM post-training and involves no human subjects. 
We use only publicly available LLMs and public post-training datasets, in accordance with their respective licenses, and we do not collect or release any new data. 
We will release code to reproduce all of our experiments. 
By reducing the cost of hyperparameter tuning, our findings can make LLM post-training cheaper and more accessible, particularly to researchers with limited compute. 
We do not anticipate negative societal impacts beyond those generally associated with improved capabilities of large language models, and therefore do not discuss such impacts further. 
We have no conflicts of interest to disclose.

\subsection*{Reproducibility statement}

We have made several efforts to ensure that the results reported in this paper can be reproduced. 
Anonymized source code for our method, together with the scripts and configuration files needed to rerun every experiment, is included in the supplementary materials. 
The complete experimental setup---datasets and preprocessing steps, model architectures, hyperparameters, training schedules, evaluation protocol, and the hardware used---is documented in Appendix~\ref{app:experimental_setup}, which expands on the summary in Section~\ref{sec:experiments}. 
Assumptions for all theoretical claims are stated in Section~\ref{sec:theory}, and proofs are given in Appendix~\ref{sec:app-theory}.



\bibliography{iclr2027_conference}
\bibliographystyle{iclr2027_conference}

\newpage

\appendix

\section{Proofs and Profiling for Learning Rate Scaling}
\label{sec:app-theory}

This appendix provides the profiling setup, complete profiling results, and proofs of the scaling results in~\Cref{sec:theory}. 
We use the same notation and assumptions introduced in the main text.

\subsection{Profiling Setup}
\label{sec:app-theory-profiling}

This subsection describes how we obtain the profiling results in~\Cref{fig:theory_profile} and Appendix~\ref{sec:app-theory-results}, which provide empirical support for the analysis in~\Cref{sec:theory}.
In each profiled run, we record at every optimizer step the gradient and the update direction of every trainable block, compute from them the first-order contributions $\widetilde d_l(\eta)$ and $\widetilde D_k(\eta)$, and measure the realized loss change $\Delta\gL_k(\eta)$ of the step.
Furthermore, at selected steps of the FP32 runs, we compute the BF16 update factor $\phi_k(\eta)$ and the fraction of nonzero realized updates $p_k(\eta)$ over a wide grid of learning rates.
Below, we describe the profiled runs and their configurations, define each measured quantity, and state how the measurements are aggregated for visualization.

\paragraph{Profiled runs.}
We profile the two Llama-3.2-1B settings used in~\Cref{sec:experiments:scaling}: Llama-3.2-1B on \tuluthree{} for SFT and Llama-3.2-1B-Instruct on UltraFeedback for DPO.
For each setting, we profile last-$k$ runs with $k\in\{1,2,4,8,16\}$ under both update precisions, FP32 master-weight updates and direct BF16 updates, which gives a total of $20$ profiled runs.
Both models have $L=16$ blocks, so the profiled runs range from the smallest proxy to the full-depth run.
The runs follow the training protocol specified in Appendix~\ref{app:experimental_setup}.
We restate the settings that matter for interpreting the measurements: a constant learning rate, so that $\eta$ is a single number for the whole run, AdamW with zero weight decay, so that the update direction contains no decay term, global-norm gradient clipping at $1.0$, BF16 autocast for computation, and effective batch sizes of $32$ examples for SFT and $8$ preference pairs for DPO.
To limit the cost of profiling, we profile the first $3000$ optimizer steps of each run, which is about $30\%$ of the $10{,}000$-step SFT epoch and $40\%$ of the roughly $7{,}600$-step DPO epoch.

\begin{table}[!htbp]
\centering
\caption{Settings and learning rates used for profiling.
For each setting, we profile runs with $k\in\{1,2,4,8,16\}$ trainable blocks using the listed learning rate scaling rule.}
\label{tab:app-theory-profiling-settings}
\begin{tabular}{lllll}
\toprule
Task & Model & Dataset & Update precision & Learning rate $\eta$ \\
\midrule
SFT & Llama-3.2-1B & \tuluthree{} & FP32 master weights & $2^{-14}/k$ \\
SFT & Llama-3.2-1B & \tuluthree{} & Direct BF16 & $2^{-13}/\sqrt{k}$ \\
DPO & Llama-3.2-1B-Instruct & UltraFeedback & FP32 master weights & $2^{-18.5}/k$ \\
DPO & Llama-3.2-1B-Instruct & UltraFeedback & Direct BF16 & $2^{-15.5}/\sqrt{k}$ \\
\bottomrule
\end{tabular}
\end{table}

\paragraph{Learning rates used for profiling.}
We profile runs at the learning rates specified in~\Cref{tab:app-theory-profiling-settings}.
These rates match those selected by the sweeps in~\Cref{sec:experiments:scaling} using the training-loss metric in $19$ of $20$ cases.
The only exception is Llama-3.2-1B with FP32 master weights at $k=16$: the sweep identifies $2^{-17.5}$, whereas we profile at the rule value $2^{-14}/16=2^{-18}$, a factor of $\sqrt{2}$ lower.
We use the rule value to maintain exact learning rate scaling across $k$.

Prescribing the learning rates does not impose the scaling properties examined by the profiling.
For the same parameters, minibatch, and Adam moment estimates, changing the learning rate of the current step leaves both the raw gradient $\nabla_{\vtheta_l}\gL$ and the clipped, Adam-normalized update direction $\vu_l$ unchanged.
Thus, $d_l=\langle\nabla_{\vtheta_l}\gL,\vu_l\rangle$ has no direct dependence on the current step's learning rate, although the learning rate could still affect it indirectly through earlier training trajectories.
We measure $\eta\widetilde D_k(\eta)$ to check whether the prescribed rates yield first-order loss reductions on a comparable scale as the number of trainable blocks $k$ varies.
With $\eta\propto 1/k$ under FP32 master-weight updates, this requires the measured sum $\widetilde D_k(\eta)$ to grow as $k$.
With $\eta\propto 1/\sqrt{k}$ under direct BF16 updates, it requires this sum to grow as $\sqrt{k}$.
These growth rates are checked from the recorded raw gradients and realized update directions, rather than assumed from the learning rate rules.
We also examine the individual $d_l$ in the FP32 runs to check whether their values are comparable across blocks and last-$k$ runs, rather than checking only their sum.
The dependence of the BF16 update factor $\phi_k(\eta)$ on $\eta$ is tested separately.
At selected steps of the FP32 runs, we hold the model parameters, raw gradients, and update directions fixed and vary the probe learning rate over a wide grid.
This directly tests whether $\phi_k(\eta)$ grows approximately linearly with $\eta$ in the relevant regime, rather than inferring this dependence from comparisons across runs trained at different learning rates.

\paragraph{First-order contributions.}
At each profiled step, we record the raw gradient $\nabla_{\vtheta_l}\gL$ and the realized update direction $\widetilde{\vu}_l(\eta)$ for every trainable block $l\in\gA_k$.
The gradient is computed on the complete minibatch before gradient clipping and recorded in BF16 under both update precisions.
We compute the realized direction from the parameters before and after the optimizer step:
\[
    \widetilde{\vu}_l(\eta)
=\frac{\vtheta_l-\vtheta_l^+}{\eta},
\]
using parameter values in the corresponding update accumulation precision.

Let $\vu_l$ denote the nominal direction produced by Adam after gradient clipping, before multiplication by $\eta$.
Under direct BF16 updates,
\[
    \vtheta_l^+=Q(\vtheta_l-\eta\vu_l),
\]
where $Q$ rounds each coordinate to the nearest representable BF16 value.
Thus, the recorded direction $\widetilde{\vu}_l(\eta)$ includes the effect of update rounding.
Under FP32 master-weight updates, we neglect FP32 rounding and treat the recorded direction as the nominal direction, $\widetilde{\vu}_l(\eta)=\vu_l$.
From the recorded gradients and realized directions, we compute the first-order contribution of each block and its sum over all trainable blocks:
\[
\widetilde d_l(\eta)
=\bigl\langle\nabla_{\vtheta_l}\gL,\,\widetilde{\vu}_l(\eta)\bigr\rangle,
\qquad
\widetilde D_k(\eta)
=\sum_{l\in\gA_k}\widetilde d_l(\eta).
\]
Each inner product includes all parameters of the block and is computed in float64 precision.
Under the FP32 idealization, these quantities equal the nominal $d_l$ and $D_k$.
Hence, $\eta\widetilde D_k(\eta)$ is the first-order approximation of the step's loss reduction.
\Cref{fig:app:first_order_loss_decrease_row} presents this approximation.

\paragraph{BF16 update factor.}
For fixed parameters, gradients, and nominal update directions, the update factor $\phi_k(\eta)=\widetilde D_k(\eta)/D_k$ depends on the learning rate.
We therefore measure it over a grid of learning rates rather than only at an individual run's learning rate.
At selected steps of each FP32 run, we save the parameters $\vtheta_l$, raw gradients $\nabla_{\vtheta_l}\gL$, and realized update directions for every trainable block $l\in\gA_k$.
We use FP32 runs so that, under the FP32 idealization, the recorded update directions equal the nominal directions $\vu_l$ needed for these probes.
Using these saved quantities, we compute, for each probe learning rate $\eta\in\{2^{-8},2^{-9},\ldots,2^{-26}\}$,
\[
\begin{aligned}
\widetilde{\vu}_l(\eta)
&=\frac{Q(\vtheta_l)-Q\bigl(Q(\vtheta_l)-\eta\vu_l\bigr)}{\eta},
\\
\widetilde D_k(\eta)
&=\sum_{l\in\gA_k}
\bigl\langle\nabla_{\vtheta_l}\gL,\widetilde{\vu}_l(\eta)\bigr\rangle,
\\
\phi_k(\eta)
&=\frac{\widetilde D_k(\eta)}{D_k}.
\end{aligned}
\]
Here, $Q$ rounds each coordinate to the nearest representable BF16 value and is implemented by casting to \texttt{torch.bfloat16} in PyTorch.
The subtraction $Q(\vtheta_l)-\eta\vu_l$ is performed in FP32, and the denominator $D_k=\sum_{l\in\gA_k}\langle\nabla_{\vtheta_l}\gL,\vu_l\rangle$ uses the same saved gradients and directions and remains fixed across different probe learning rates.
The inner $Q$ rounds the starting parameters to BF16, and the outer $Q$ rounds the updated parameters.
The probe grid extends below and above the training rates in \Cref{tab:app-theory-profiling-settings}, without changing the run's training rate.
All probes at a given step use the same saved parameters, gradients, and update directions, isolating the effect of $\eta$ on update rounding from its effect on the training trajectory.
From the same probes, we also record the fraction of the $N_k$ trainable parameter elements whose realized update $\widetilde u_{l,i}(\eta)$ is nonzero.
This procedure also applies to other update precisions.
To compute the update factor for FP16, we replace the \texttt{torch.bfloat16} casts with \texttt{torch.float16} casts and keep the rest of the computation unchanged.

\paragraph{Profiled steps and aggregation.}
We record $\widetilde d_l(\eta)$, $\widetilde D_k(\eta)$, and $\Delta\gL_k(\eta)$ at every step of each run, whereas $\phi_k(\eta)$ and $p_k(\eta)$ are measured at certain steps $t\in\{500,1000,1500,2000,2500,3000\}$.

\Cref{fig:app:first_order_relative_error_row} plots the per-step relative error $|\Delta\gL_k(\eta)+\eta\widetilde D_k(\eta)|/|\eta\widetilde D_k(\eta)|$.
\Cref{fig:app:first_order_loss_decrease_row} plots $\eta\widetilde D_k(\eta)$ at each training step, smoothed by averaging over that step and the nine steps before it.
\Cref{fig:app:per_block_loss_decrease} plots the geometric mean of $\widetilde d_l(\eta)$ over all $3000$ steps against the block index $l$.
We use the geometric rather than the arithmetic mean to reduce the influence of large, noisy values in the initial steps.
\Cref{fig:app:rounding_llama_row,fig:app:rounding_llama_instruct_row} plot $\phi_k(\eta)$, and \Cref{fig:app:nonzero_updates,fig:app:nonzero_updates_instruct} plot $p_k(\eta)$, against the probe learning rate $\eta$ at each of the six selected training steps.

\subsection{Complete Profiling Results}
\label{sec:app-theory-results}

The following profiling results complement the representative results in~\Cref{fig:theory_profile}.

\begin{figure}[!htbp]
    \centering
    \includegraphics[width=\linewidth]{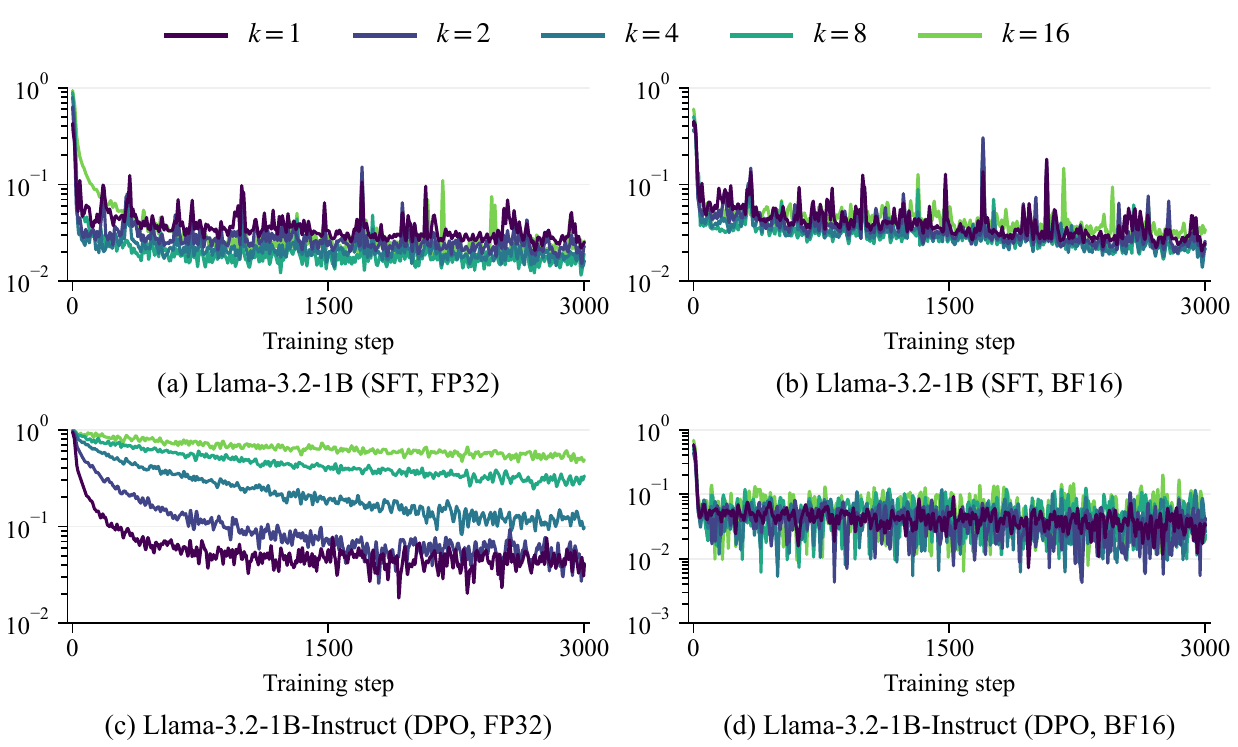}
    \caption{{Relative error of the realized first-order loss approximation across training steps.}
    We plot $|\Delta\gL_k(\eta)+\eta\widetilde D_k(\eta)|/|\eta\widetilde D_k(\eta)|$ at each training step. 
    Under the FP32 idealization, $\widetilde D_k(\eta)=D_k$.
    Each curve corresponds to a last-$k$ run with $k\in\{1,2,4,8,16\}$ at the learning rate specified in~\Cref{tab:app-theory-profiling-settings}. 
    Panels (a, b) show SFT on Llama-3.2-1B with FP32 and BF16 update accumulation, respectively. Panels (c, d) show DPO on Llama-3.2-1B-Instruct with the same precision settings.}
    \label{fig:app:first_order_relative_error_row}
\end{figure}

\paragraph{Accuracy of the first-order approximation.}
\Cref{fig:app:first_order_relative_error_row} reports the per-step relative error of the realized first-order loss approximation.
The ratio $|\Delta\gL_k(\eta)+\eta\widetilde D_k(\eta)|/|\eta\widetilde D_k(\eta)|$ compares the magnitude of the higher-order term with the realized first-order loss reduction.
The small relative errors observed in most settings support the accuracy of the first-order approximation in those settings.
\arxiv{In DPO under FP32 master-weight updates, the relative error is larger and increases with $k$.
Nonetheless, it still stays below one for every $k$ and decreases steadily during training, so the higher-order term remains smaller than the first-order term, as the condition $\epsilon<1$ in \Cref{prop:theory-scaling} requires.}

\begin{figure}[!htbp]
    \centering
    \includegraphics[width=\linewidth]{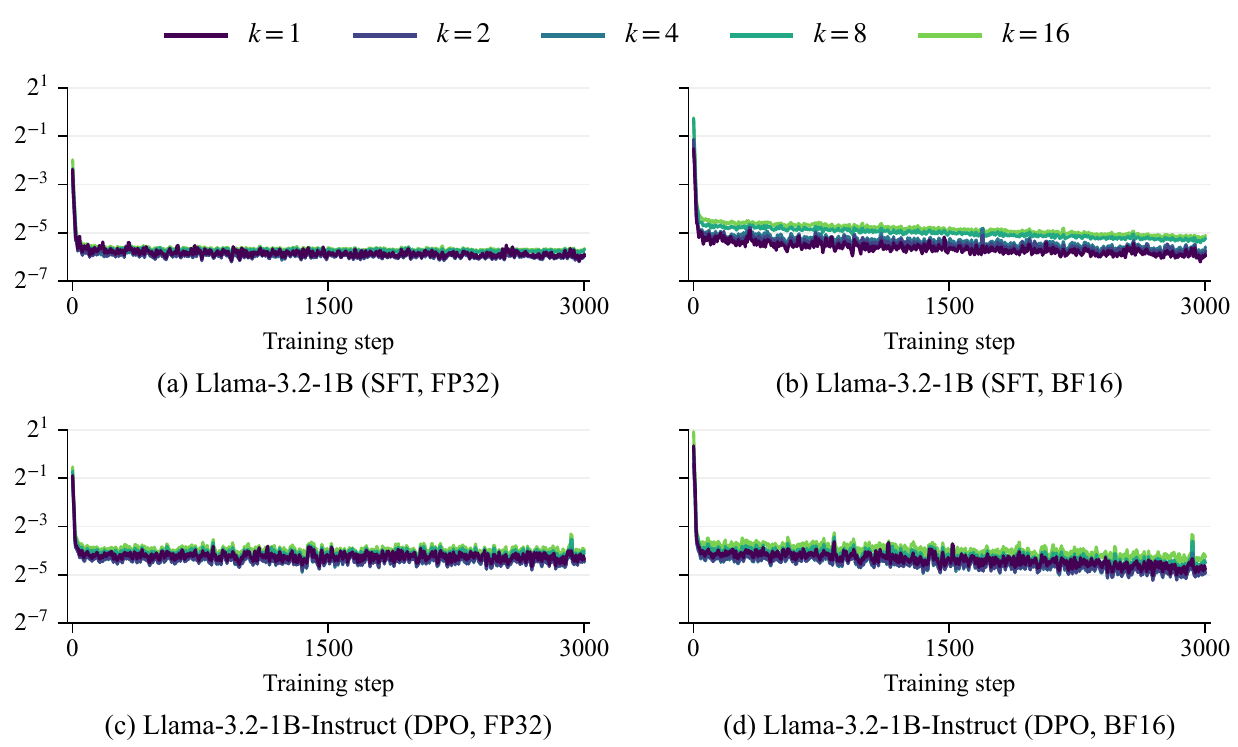}
    \caption{{Realized first-order loss reduction across training steps.} 
    We plot $\eta\widetilde D_k(\eta)$ averaged over each step and the preceding nine steps.
    Under the FP32 idealization, this quantity equals $\eta D_k$.
    Each curve corresponds to a last-$k$ run with $k\in\{1,2,4,8,16\}$
    at the learning rate specified in~\Cref{tab:app-theory-profiling-settings}.
    Panels (a, b) show SFT on Llama-3.2-1B with FP32 and BF16 update
    accumulation, respectively. Panels (c, d) show DPO on
    Llama-3.2-1B-Instruct with the same respective precision settings.}
    \label{fig:app:first_order_loss_decrease_row}
\end{figure}

\paragraph{Stable loss reduction.}
\Cref{fig:app:first_order_loss_decrease_row} shows the realized first-order loss reduction, $\eta\widetilde D_k(\eta)$, across training steps.
We observe that the per-step values of $\eta\widetilde D_k(\eta)$ at the learning rates specified in~\Cref{tab:app-theory-profiling-settings} remain on a comparable scale across $k$, supporting Assumption~\ref{asm:theory-stable}.

\begin{figure}[!htbp]
    \centering
    \includegraphics[width=\linewidth]{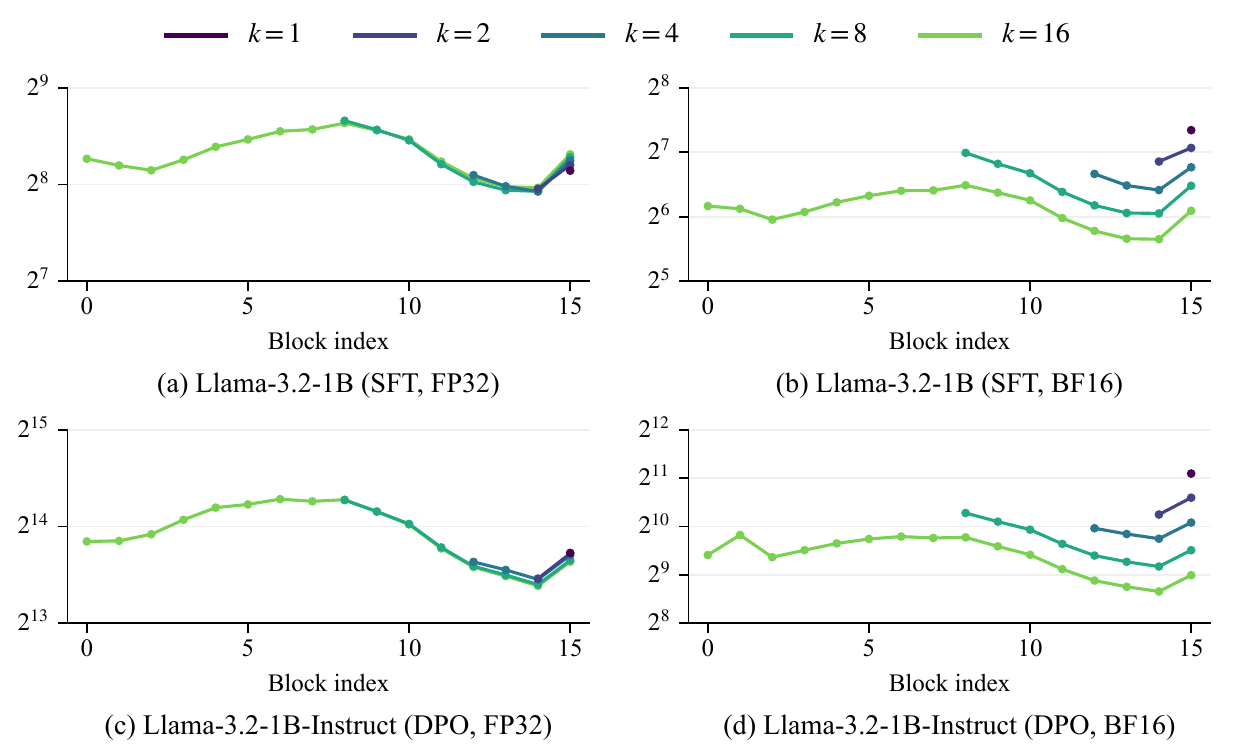}
    \caption{{Realized per-block first-order loss contributions.}
    For each trainable block $l$, we plot the geometric mean over training
    steps of $\widetilde d_l(\eta)=\langle\nabla_{\vtheta_l}\gL,\widetilde{\vu}_l(\eta)\rangle$.
    Under the FP32 idealization, we have $\widetilde d_l(\eta)=d_l$.
    Each curve corresponds to a last-$k$ run with $k\in\{1,2,4,8,16\}$
    at the learning rate specified in~\Cref{tab:app-theory-profiling-settings}.
    The horizontal axis uses block indices $0,\ldots,15$ from bottom to top.
    The vertical axes are logarithmic with panel-specific ranges.
    Panels (a, b) show SFT on Llama-3.2-1B with FP32 and BF16 update
    accumulation, respectively. Panels (c, d) show DPO on
    Llama-3.2-1B-Instruct with the same respective precision settings.}
    \label{fig:app:per_block_loss_decrease}
\end{figure}

\paragraph{Per-block contributions.}
\Cref{fig:app:per_block_loss_decrease} plots the realized per-block contribution, $\widetilde d_l(\eta)$, as a function of block position $l$, taking the geometric mean over training steps. 
Under the FP32 idealization, we have $\widetilde d_l(\eta)=d_l$. 
In the FP32 panels (a, c), these values are approximately invariant to $k$ within each block
and comparable across blocks, which supports Assumption~\ref{asm:theory-contributions}. 
The BF16 panels (b, d) are also included for comparison and reflect the effects of update rounding.

\begin{figure}[!htbp]
    \centering
    \includegraphics[width=\linewidth]{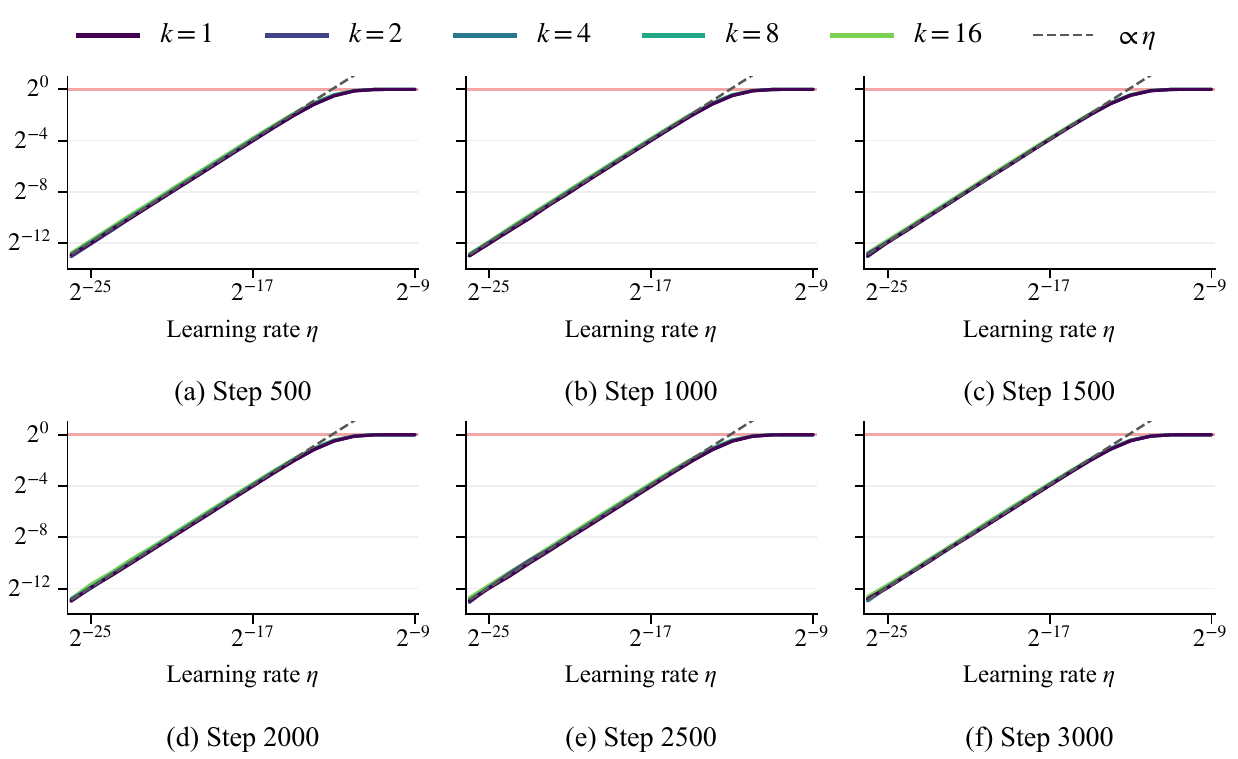}
    \caption{{BF16 update factor for SFT on Llama-3.2-1B.}
    We plot $\phi_k(\eta)=\widetilde D_k(\eta)/D_k$ against the learning rate $\eta$ at six steps during training.
    The colored curves correspond to $k\in\{1,2,4,8,16\}$.
    Panels (a)–(f) correspond to steps $500$, $1000$, $1500$, $2000$, $2500$, and $3000$, respectively.
    Both axes are logarithmic, and the dashed line is a slope-1 reference ($\propto \eta$).}
    \label{fig:app:rounding_llama_row}
\end{figure}

\begin{figure}[!htbp]
    \centering
    \includegraphics[width=\linewidth]{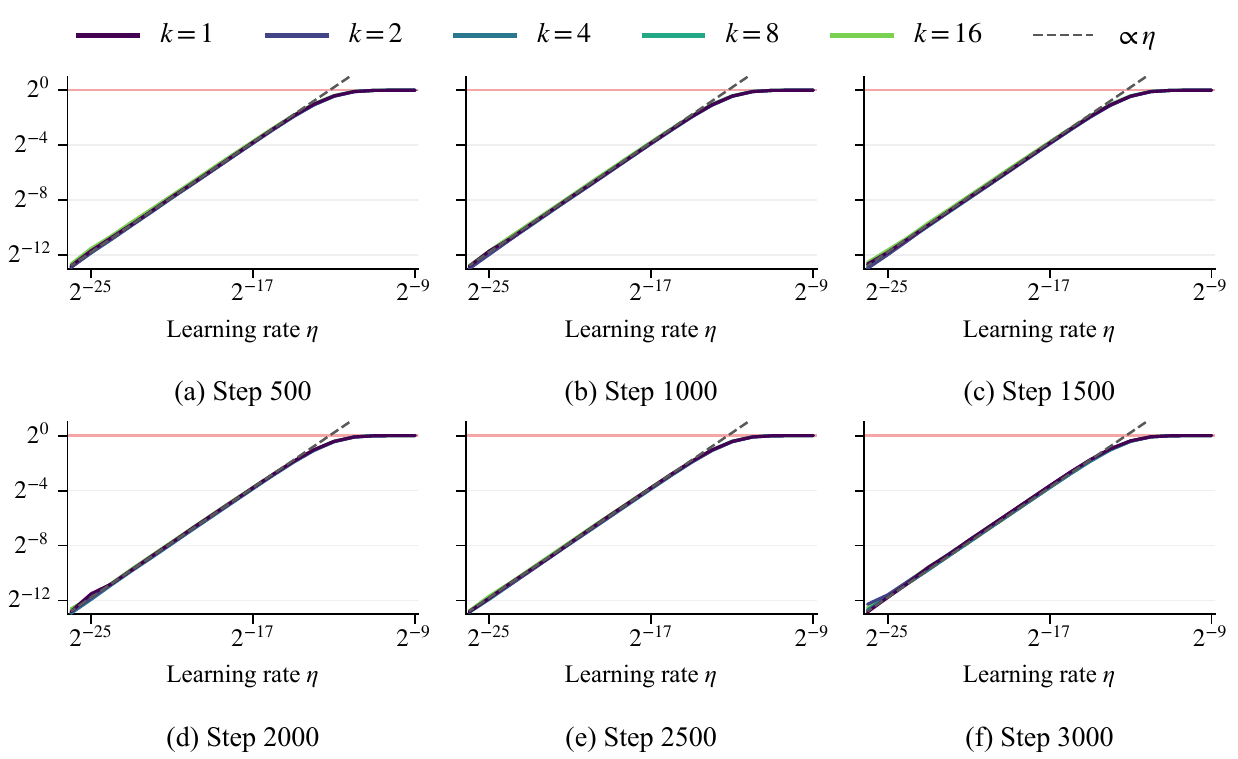}
    \caption{{BF16 update factor for DPO on Llama-3.2-1B-Instruct.}
    We plot $\phi_k(\eta)=\widetilde D_k(\eta)/D_k$ against the learning rate $\eta$ at six steps during training.
    The colored curves correspond to $k\in\{1,2,4,8,16\}$.
    Panels (a)–(f) correspond to steps $500$, $1000$, $1500$, $2000$, $2500$, and $3000$, respectively. 
    Both axes are logarithmic, and the dashed line is a slope-1 reference ($\propto \eta$).}
    \label{fig:app:rounding_llama_instruct_row}
\end{figure}

\paragraph{BF16 update factor.}
\Cref{fig:app:rounding_llama_row,fig:app:rounding_llama_instruct_row} report the BF16 update factor as a function of learning rate for SFT and DPO, respectively. 
Across the displayed training steps, the curves for different $k$ nearly coincide and grow approximately linearly with $\eta$ before saturation. 
These profiles support the case $\alpha=1$ of Assumption~\ref{asm:theory-realization} over the corresponding learning rate regime.

\begin{figure}[!htbp]
    \centering
    \includegraphics[width=\linewidth]{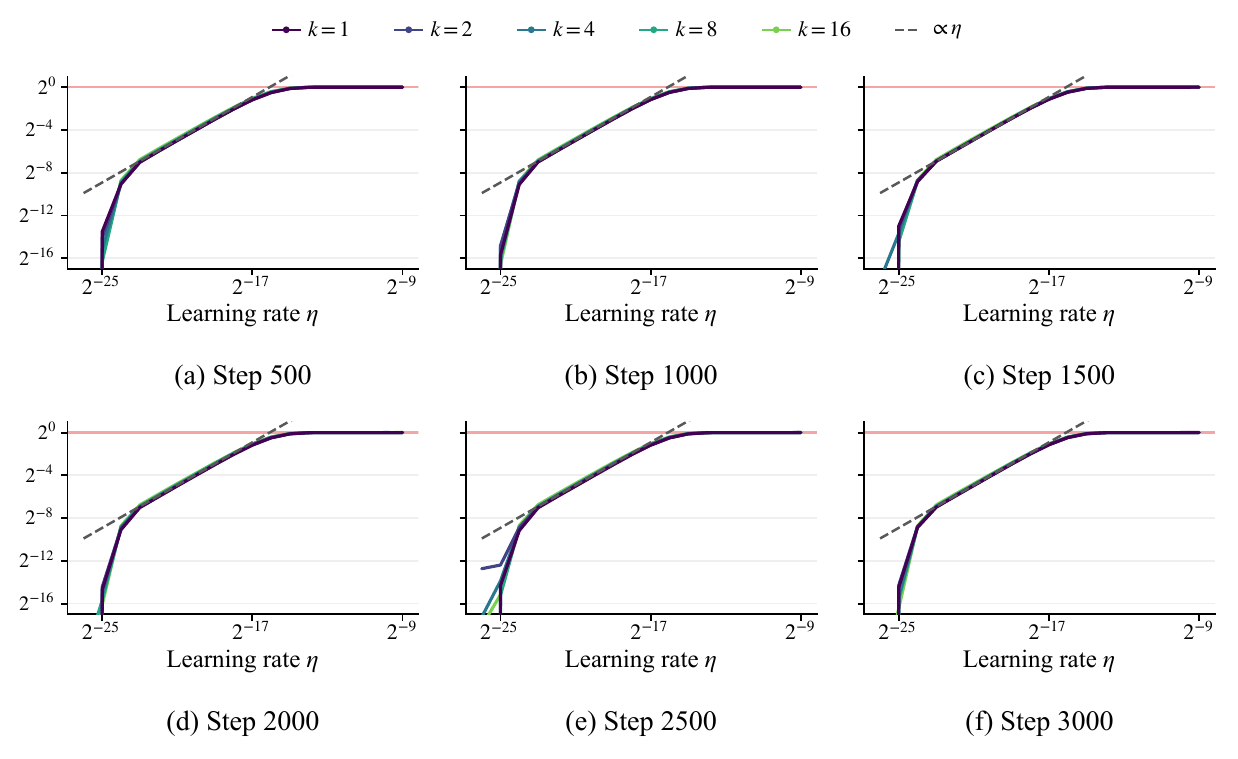}
    \caption{{FP16 update factor for SFT on Llama-3.2-1B.}
    We plot $\phi_k(\eta)=\widetilde D_k(\eta)/D_k$ against the learning rate $\eta$ at six steps during training.
    The colored curves correspond to $k\in\{1,2,4,8,16\}$.
    Panels (a)–(f) correspond to steps $500$, $1000$, $1500$, $2000$, $2500$, and $3000$, respectively.
    Both axes are logarithmic, and the dashed line is a slope-1 reference ($\propto \eta$).}
    \label{fig:app:rounding_llama_row_fp16}
\end{figure}

\paragraph{FP16 update factor.}
\Cref{fig:app:rounding_llama_row_fp16} shows the FP16 update factor versus learning rate for Llama-3.2-1B under SFT. 
Across the shown training steps, curves for different $k$ nearly overlap and increase approximately linearly with $\eta$ before saturating. 
At extremely small learning rates, the update factor drops to zero due to underflow.

\subsection{Update Sparsity and the BF16 Update Factor}
\label{sec:app-theory-bf16-sparsity}

\begin{figure}[!htbp]
    \centering
    \includegraphics[width=\linewidth]{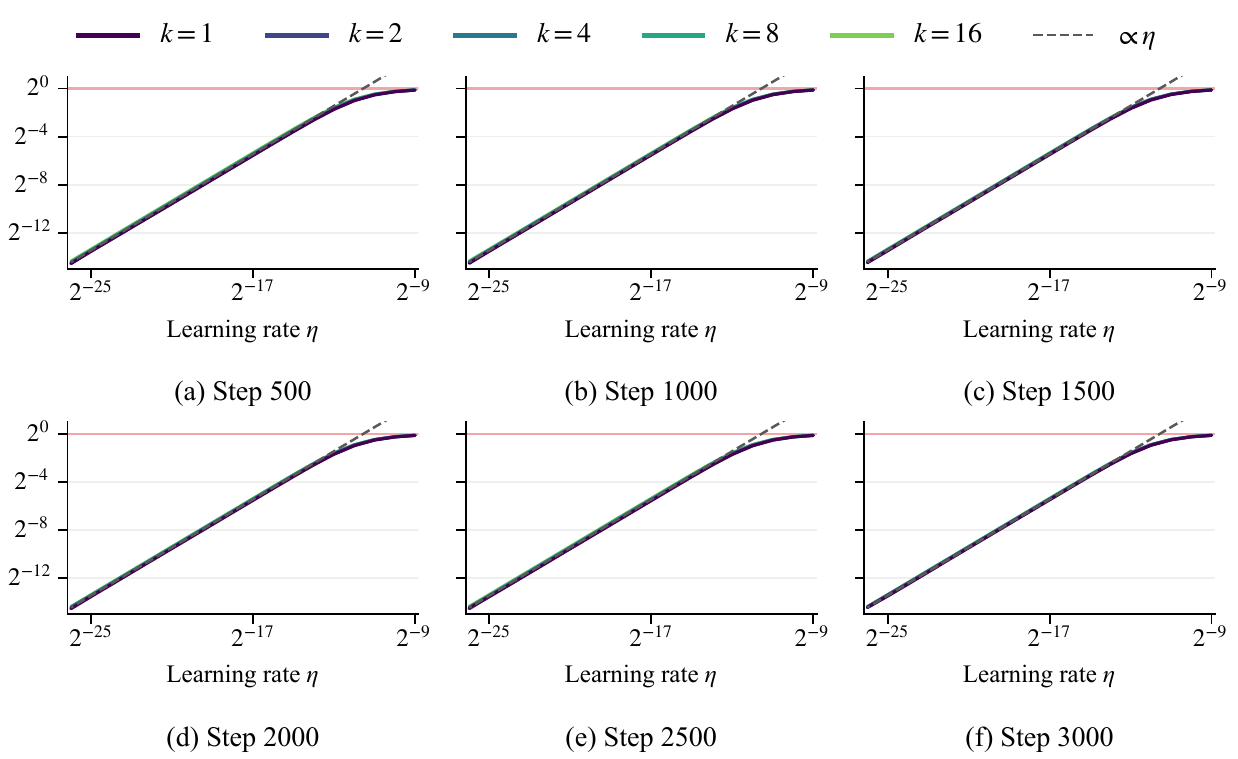}
    \caption{{Fraction of nonzero realized updates for SFT on Llama-3.2-1B.}
    We plot $p_k(\eta)$, the fraction of trainable parameter elements whose realized update $\widetilde u_{l,i}(\eta)$ under direct BF16 updates is nonzero, against the learning rate $\eta$ at six steps during training.
    The colored curves correspond to $k\in\{1,2,4,8,16\}$.
    Panels (a)–(f) correspond to steps $500$, $1000$, $1500$, $2000$, $2500$, and $3000$, respectively.
    Both axes are logarithmic; the dashed line is a slope-1 reference ($\propto \eta$), and the red line marks the upper bound $p_k(\eta)=1$.}
    \label{fig:app:nonzero_updates}
\end{figure}

\begin{figure}[!htbp]
    \centering
    \includegraphics[width=\linewidth]{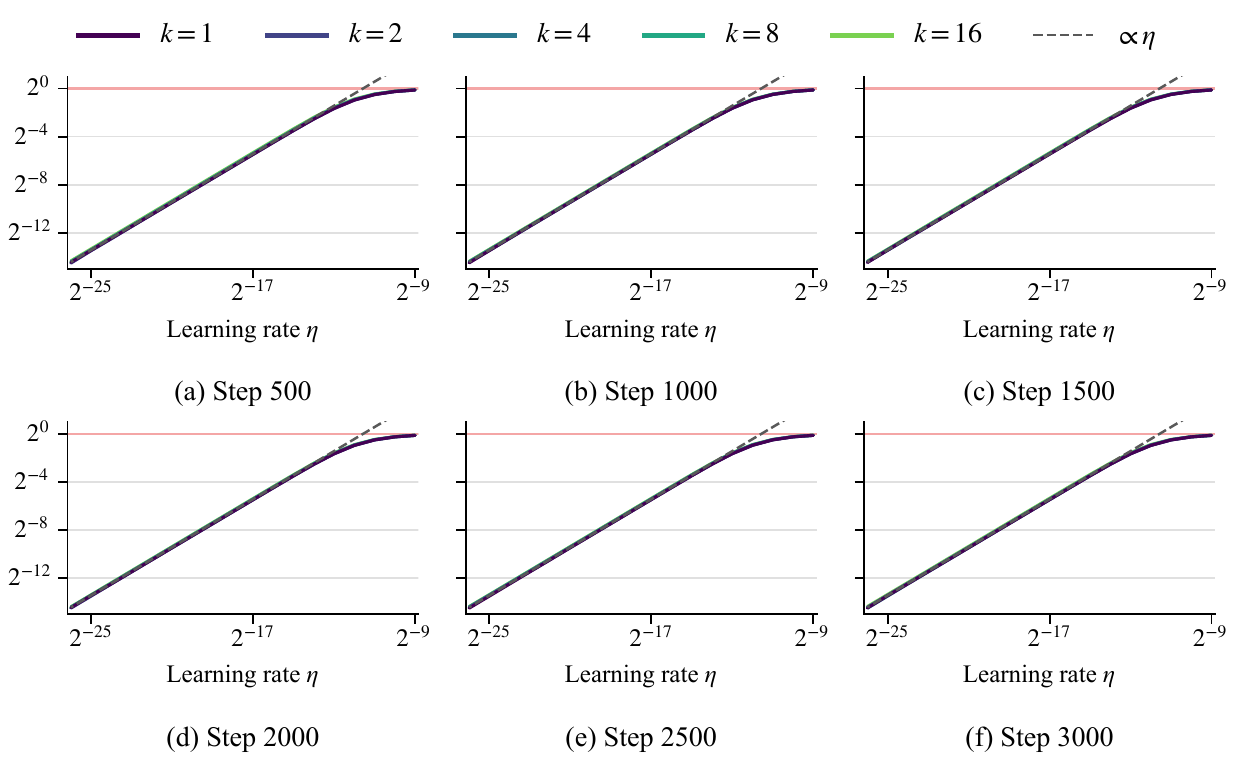}
    \caption{{Fraction of nonzero realized updates for DPO on Llama-3.2-1B-Instruct.}
    We plot $p_k(\eta)$, the fraction of trainable parameter elements whose realized update $\widetilde u_{l,i}(\eta)$ under direct BF16 updates is nonzero, against the learning rate $\eta$ at six steps during training.
    The colored curves correspond to $k\in\{1,2,4,8,16\}$.
    Panels (a)–(f) correspond to steps $500$, $1000$, $1500$, $2000$, $2500$, and $3000$, respectively.
    Both axes are logarithmic; the dashed line is a slope-1 reference ($\propto \eta$), and the red line marks the upper bound $p_k(\eta)=1$.}
    \label{fig:app:nonzero_updates_instruct}
\end{figure}

Write $g_{l,i}$, $u_{l,i}$, and $\widetilde u_{l,i}(\eta)$ for the
coordinates of $\nabla_{\vtheta_l}\gL$, $\vu_l$, and
$\widetilde{\vu}_l(\eta)$, respectively.
Let $n_l$ denote the number of parameters in block $l$ and
$N_k=\sum_{l\in\gA_k}n_l$ the total number of trainable parameter
elements.
Write $\E_k$ for expectation over a coordinate $J=(l,i)$ drawn
uniformly from these trainable elements.
Define
\[
    p_k(\eta)
    =\frac{1}{N_k}
    \sum_{l\in\gA_k}\sum_{i=1}^{n_l}
    \1_{\{\widetilde u_{l,i}(\eta)\ne0\}},
    \qquad
    \mu_k=\E_k[g_Ju_J]=\frac{D_k}{N_k}>0.
\]
Here, $p_k(\eta)$ is the fraction of nonzero realized updates,
and $\mu_k$ is the mean nominal first-order contribution per
trainable parameter element.
For $p_k(\eta)>0$, the update factor factorizes exactly as
\begin{equation}
    \phi_k(\eta)
    =p_k(\eta)\,
    \frac{
        \E_k\!\left[
            g_J\widetilde u_J(\eta)
            \,\middle|\,
            \widetilde u_J(\eta)\ne0
        \right]
    }{\mu_k},
    \label{eq:theory-bf16-factorization}
\end{equation}
because coordinates with zero realized updates contribute nothing.
The second factor compares the mean realized first-order
contribution among coordinates with nonzero updates to the mean
nominal contribution over all trainable coordinates.
\begin{proposition}[Sufficient conditions for a linear BF16 update factor (informal)]
\label{prop:theory-bf16-sparsity}
Under direct BF16 updates, suppose that
\begin{equation}
    p_k(\eta)=\Theta(\eta)
    \qquad\text{and}\qquad
    \frac{\E_k\!\left[g_J\widetilde u_J(\eta)\,\middle|\,\widetilde u_J(\eta)\ne0\right]}{\mu_k}=\Theta(1)
    \label{eq:theory-bf16-sparsity-conditions}
\end{equation}
uniformly over the learning rate regime considered.
Then
\begin{equation}
    \phi_k(\eta)=\Theta(\eta)
    \label{eq:theory-bf16-sparsity-result}
\end{equation}
over this regime; that is, Assumption~\ref{asm:theory-realization} holds with $\alpha=1$.
\end{proposition}
\Cref{fig:app:nonzero_updates,fig:app:nonzero_updates_instruct} show that the fraction of nonzero
realized updates $p_k(\eta)$ grows approximately linearly with $\eta$
before saturation across the displayed values of $k$ and training
steps, supporting the first condition in
\Eqref{eq:theory-bf16-sparsity-conditions}.

\paragraph{Relevance of the linear regime.}
Our derivation of the $k^{-1/2}$ rule for direct BF16 updates assumes that the optimal learning rates lie in the linear regime, before $\phi_k(\eta)$ saturates.
We conjecture that this holds because the realized updates at learning rates near saturation ($\phi_k(\eta)=1$) are too large for effective post-training.
Since post-training starts from useful pretrained representations, adaptation need not involve large changes throughout the model~\citep{hayou2024impact}.
FP32 master-weight updates can change every coordinate, so the optimal FP32 learning rates, which are much smaller than the saturation point, provide empirical evidence of the useful update scale.
Direct BF16 updates require larger learning rates to compensate for increments suppressed by rounding (\Cref{sec:experiments:scaling}).
However, this compensation does not require every coordinate to change.
Under the conditions of Proposition~\ref{prop:theory-bf16-sparsity}, the realized first-order loss reduction is roughly proportional to $\eta$ times the fraction of nonzero updates.
Because this fraction also grows with $\eta$, a larger learning rate allows BF16 to reach the loss reduction obtained at the optimal FP32 learning rate before $\phi_k(\eta)$ saturates.
Raising $\eta$ further toward saturation changes more coordinates by larger increments, which can make the realized update too large for effective adaptation.
Therefore, as long as the optimal FP32 learning rate lies well below the saturation point, the optimal BF16 learning rate lies between the two, within the linear regime.

\subsection{Proofs}
\label{sec:app-theory-proofs}

Throughout this subsection, $\Theta(\cdot)$ denotes positive two-sided bounds with constants independent of $k$ and $L$.
In Assumption~\ref{asm:theory-contributions}, the same bounds on $d_l$ apply to every trainable block in the transformer.
In Assumption~\ref{asm:theory-realization}, the ratio $\phi_k(\eta)/\eta^\alpha$ remains between fixed positive constants throughout the learning rate regime considered.
In the following, we present a proof for Proposition~\ref{prop:theory-scaling} and Corollary~\ref{cor:theory-precision}.

\paragraph{Proposition~\ref{prop:theory-scaling} (restated).}
\emph{Let $1\le k\le L$, $\gA_k=\{L-k+1,\ldots,L\}$, and
$D_k=\sum_{l\in\gA_k}d_l$. Suppose that, for a fixed $\alpha\ge0$,}
\[
 d_l=\Theta(1),\qquad l\in\gA_k,
 \qquad
 \phi_k(\eta)=\Theta(\eta^\alpha)
\]
\emph{throughout the learning rate regime considered. Assume also that the higher-order term satisfies}
\[
 \left|\Delta\gL_k(\eta)+\eta D_k\phi_k(\eta)\right|
 \le\eps\,\eta D_k\phi_k(\eta),
 \qquad 0\le\eps<1,
\]
\emph{where $\Delta\gL_k(\eta)=\gL(\vtheta^+)-\gL(\vtheta)$ and $\eps$ is fixed independently of $k$, $L$, and $\eta$. 
Then every optimal learning rate $\eta_k$ within this regime, satisfying
$-\Delta\gL_k(\eta_k)=\Theta(1)$, obeys}
\[
 \eta_k=\Theta\!\left(k^{-1/(1+\alpha)}\right).
\]

\begin{proof}
Assumption~\ref{asm:theory-contributions} gives
\[
 D_k=\sum_{l\in\gA_k}d_l=\Theta(k).
\]
Combining this with $\phi_k(\eta)=\Theta(\eta^\alpha)$ yields
\[
 \eta D_k\phi_k(\eta)=\Theta\!\left(k\eta^{1+\alpha}\right).
\]
The first-order approximation implies
\[
 (1-\eps)\eta D_k\phi_k(\eta)
 \le-\Delta\gL_k(\eta)
 \le(1+\eps)\eta D_k\phi_k(\eta).
\]
Since $\eps<1$ is fixed, it follows that $-\Delta\gL_k(\eta)=\Theta(k\eta^{1+\alpha})$. 
Therefore, by assumption~\ref{asm:theory-stable},
\[
 k(\eta_k)^{1+\alpha}=\Theta(1).
\]
Dividing by $k$ and taking the positive $(1+\alpha)$-th root gives
\[
 \eta_k=\Theta\!\left(k^{-1/(1+\alpha)}\right).
\]
\end{proof}

\paragraph{Corollary~\ref{cor:theory-precision} (restated).}
\emph{Under the conditions of Proposition~\ref{prop:theory-scaling},
FP32 master-weight updates have $\phi_k(\eta)=1$, corresponding to $\alpha=0$, whereas direct BF16 updates with $\phi_k(\eta)=\Theta(\eta)$ have $\alpha=1$. Consequently,}
\[
 \eta_k=
 \begin{cases}
  \Theta(k^{-1}), & \text{FP32 master-weight updates},\\
  \Theta(k^{-1/2}), & \text{direct BF16 updates}.
 \end{cases}
\]

\begin{proof}
Under the FP32 idealization, $Q$ is the identity, so $\widetilde{\vu}_l(\eta)=\vu_l$ and $\widetilde D_k(\eta)=D_k$. 
Thus $\phi_k(\eta)=1$, and the proposition with $\alpha=0$ gives $\eta_k=\Theta(k^{-1})$.
For direct BF16 updates without an FP32 master copy, the assumed relation $\phi_k(\eta)=\Theta(\eta)$ corresponds to $\alpha=1$.
The proposition then gives $\eta_k=\Theta(k^{-1/2})$.
\end{proof}

\paragraph{Proposition~\ref{prop:theory-bf16-sparsity} (restated).}
\emph{Under direct BF16 updates, suppose that $D_k>0$ and}
\[
    p_k(\eta)=\Theta(\eta),
    \qquad
    \frac{
        \E_k\!\left[
            g_J\widetilde u_J(\eta)
            \,\middle|\,
            \widetilde u_J(\eta)\ne0
        \right]
    }{\mu_k}
    =\Theta(1),
\]
\emph{uniformly over the learning rate regime considered, with
positive constants independent of $k$, $L$, and $\eta$.
Then $\phi_k(\eta)=\Theta(\eta)$ throughout this regime, so
Assumption~\ref{asm:theory-realization} holds with $\alpha=1$.}

\begin{proof}
The two conditions give constants $0<a\le b<\infty$ and
$0<c\le C<\infty$, independent of $k$, $L$, and $\eta$, such that
\[
    a\eta\le p_k(\eta)\le b\eta,
    \qquad
    c\le
    \frac{
        \E_k\!\left[
            g_J\widetilde u_J(\eta)
            \,\middle|\,
            \widetilde u_J(\eta)\ne0
        \right]
    }{\mu_k}
    \le C
\]
throughout the regime considered.
In particular, $p_k(\eta)>0$ for $\eta>0$, so the conditional
expectation is well-defined.
Since $\mu_k=D_k/N_k>0$, the exact factorization in
\Eqref{eq:theory-bf16-factorization} gives
\[
    ac\,\eta
    \le
    p_k(\eta)\,
    \frac{
        \E_k\!\left[
            g_J\widetilde u_J(\eta)
            \,\middle|\,
            \widetilde u_J(\eta)\ne0
        \right]
    }{\mu_k}
    =\phi_k(\eta)
    \le bC\,\eta.
\]
The constants $ac$ and $bC$ are positive and independent of
$k$, $L$, and $\eta$.
Therefore, $\phi_k(\eta)=\Theta(\eta)$ uniformly over this regime,
which is Assumption~\ref{asm:theory-realization} with $\alpha=1$.
\end{proof}

\section{Experimental Setup}
\label{app:experimental_setup}

\subsection{Shared training protocol}
\label{app:exp_setup_shared}

\paragraph{Trainable parameters and learning rate.}
Unless stated otherwise, we finetune the last $k$ blocks of a model with $L$ transformer blocks,
\[
\mathcal{A}_k=\{L-k+1,\ldots,L\}.
\]
All parameters of these blocks are trainable: the self-attention modules, the gated MLPs, and the RMSNorm parameters.
The input embeddings, the LM head, the final normalization layer, and all unselected blocks stay frozen.
Full-depth runs set $k=L$.
Weights are updated in place, and no adapters such as LoRA~\citep{hu2022lora} are used.
Learning rates are swept on a base-2 grid, $\eta=2^{-a}$, with integer exponents $a$ and half-integer refinements near the optimum.

\paragraph{Optimization and batching.}
Table~\ref{tab:setup_shared} lists the settings shared by all experiments.
Batch sizes are effective batch sizes: with $N$ data-parallel processes, a per-device microbatch of $b$, and $g$ accumulated microbatches, one optimizer step uses $Nbg$ examples.
For RLVR, an example is one sampled completion.
For GRPO, an example is one sampled completion.
We smooth the training loss $\ell_t$ at step $t$ with an exponential moving average (EMA), following $\bar{\ell}_t = 0.1\,\ell_t + 0.9\,\bar{\ell}_{t-1}$.

\begin{table}[t]
\centering
\caption{Training settings shared by all experiments, unless stated otherwise.}
\label{tab:setup_shared}
\begin{tabularx}{\linewidth}{@{}l X@{}}
\toprule
Setting & Value \\
\midrule
Training duration & $1$ epoch \\
Optimizer & Fused PyTorch AdamW \\
Adam coefficients & $(\beta_1,\beta_2)=(0.9,0.999)$ \\
Adam epsilon & $10^{-8}$ \\
Weight decay & $0$ \\
Gradient clipping & Maximum gradient norm $1.0$ \\
Mixed-precision compute & BF16 \\
TF32 matrix multiplication & On \\
Padding multiple & $8$ tokens \\
Random seed & $42$ \\
\bottomrule
\end{tabularx}
\end{table}

\paragraph{Parameter-update precision.}
FP32, BF16, and FP16 denote the precision in which parameter updates are accumulated, not necessarily the precision used for forward and backward computation.

For runs without DeepSpeed, we load and store the model weights in the specified dtype. 
PyTorch AdamW stores its first- and second-moment estimates in the same dtype as the parameters it updates. 
Thus, FP32 runs store the weights and optimizer moments in FP32, whereas BF16 runs store all states in BF16 and apply updates directly to the BF16 stored weights, without an FP32 master copy.
FP16 runs likewise store the weights in FP16 and update them without an FP32 master copy, but keep their optimizer moments in FP32 to avoid underflow and maintain stable training.
All configurations utilize PyTorch BF16 autocast, which selects the precision of individual operations without changing the dtype of the stored weights. 
Matrix multiplications in the linear and attention layers use BF16, with inputs and weights cast as needed, while numerically sensitive operations such as cross-entropy loss use FP32. 
Backward operations follow the precision selected for their corresponding forward operations, and each weight’s gradient is stored in that weight’s dtype. 
Autocast therefore controls computation precision but does not change the precision in which parameter updates are accumulated during training.

For larger models, FP32 runs are instead trained on multiple GPUs with the DeepSpeed ZeRO stage~3~\citep{rajbhandari2020zero}, as marked in the ZeRO-3 column of Tables~\ref{tab:setup_sft}, \ref{tab:setup_dpo}, and~\ref{tab:setup_rlvr}.
In our ZeRO-3 configuration, forward and backward computation uses BF16 weights, while parameter updates are accumulated in a separate FP32 master copy. 
Specifically, gradients are accumulated in FP32 with gradient accumulation, and AdamW updates the master weights using FP32 moments. 
After each optimizer step, the updated master weights are cast and copied back to the BF16 weights for subsequent computation. 
These runs thus retain FP32 update accumulation despite using BF16 weights for computation. 
ZeRO-3 also partitions the weights, gradients, master copy, and optimizer moments across GPUs to reduce per-GPU memory requirements.
During forward computation, ZeRO-3 uses all-gather communication to reconstruct each layer’s full weight matrices from the parameter shards distributed across GPUs, which incurs additional communication overhead.
The precision scheme above corresponds to \texttt{bf16.enabled=auto} and \texttt{data\_types.grad\_accum\_dtype=fp32} in our DeepSpeed configuration.
We additionally set \texttt{overlap\_comm=true} to overlap communication with computation and \texttt{contiguous\_gradients=true} to reduce memory fragmentation, and we use no parameter or optimizer offloading, so all shards remain in the GPU memory.

\subsection{Supervised finetuning (SFT)}
\label{app:exp_setup_sft}

Table~\ref{tab:setup_sft} lists the SFT experiments in this work.
The loss is the negative log-likelihood (NLL) averaged over supervised tokens, and sequences are not packed.

\begin{table}[t]
\centering
\caption{SFT experiment details. Train and Valid give the numbers of examples after preprocessing. Length gives the maximum  sequence length in tokens, and Batch the effective batch size. Cosine denotes linear warmup over $5\%$ of the steps followed by cosine decay to $0.1\eta$. Constant denotes zero warmup followed by a constant learning rate. ZeRO-3 indicates whether the FP32 runs use DeepSpeed ZeRO stage~3. BF16 runs do not use DeepSpeed.}
\label{tab:setup_sft}
\footnotesize
\setlength{\tabcolsep}{2.5pt}
\begin{tabularx}{\linewidth}{@{}>{\raggedright\arraybackslash}X >{\raggedright\arraybackslash}X r c c c l c@{}}
\toprule
Model & Dataset & Train & Valid & Length & Batch & Schedule & ZeRO-3 \\
\midrule
Llama-3.2-1B & \tuluthree{} & $320{,}000$ & $32{,}000$ & $1{,}024$ & $32$ & Constant & No \\
DeepSeek-Coder-6.7B & Magicoder-OSS-Instruct & $72{,}198$ & --- & $1{,}024$ & $32$ & Cosine & Yes \\
Mistral-7B-v0.1 & Magicoder-OSS-Instruct & $72{,}360$ & --- & $1{,}024$ & $32$ & Cosine & --- \\
\bottomrule
\end{tabularx}
\end{table}

\begin{table}[t]
\centering
\caption{DPO experiments. Train and Valid give the numbers of preference pairs after preprocessing. Loss gives the DPO variant, and $\beta$ is the corresponding DPO coefficient. Length gives the maximum sequence length in tokens, and Batch the effective batch size. Constant denotes zero warmup followed by a constant learning rate. ZeRO-3 indicates whether the FP32 runs use DeepSpeed ZeRO stage~3. BF16 runs do not use DeepSpeed.}
\label{tab:setup_dpo}
\footnotesize
\setlength{\tabcolsep}{2.5pt}
\begin{tabularx}{\linewidth}{@{}>{\raggedright\arraybackslash}X >{\raggedright\arraybackslash}X r c c c l l c l}
\toprule
Model & Dataset & Train & Valid & Length & Batch & Schedule & Loss & $\beta$ & ZeRO-3 \\
\midrule
Llama-3.2-1B-Instruct & UltraFeedback & $61{,}141$ & $987$ & $1{,}024$ & $8$ & Constant & Sigmoid & $0.1$ & No \\
OLMo-2-1124-7B-SFT & OLMo preference mix & $24{,}000$ & $2{,}400$ & $2{,}048$ & $8$ & Constant & Sigmoid-norm & $5.0$ & Yes \\
\bottomrule
\end{tabularx}
\end{table}

\begin{table}[t]
\centering
\caption{RLVR experiment details. Train and Test give the numbers of problems in the training and test splits. Length gives the maximum completion length in tokens. Batch gives the effective batch size as the number of training problems per optimizer step times the number of completions sampled per problem. Steps gives the number of optimizer steps. Warmup-constant denotes linear warmup over $10$ steps followed by a constant learning rate. ZeRO-3 indicates whether the FP32 runs use DeepSpeed ZeRO stage~3. BF16 runs do not use DeepSpeed.}
\label{tab:setup_rlvr}
\footnotesize
\setlength{\tabcolsep}{2.5pt}
\begin{tabular*}{\linewidth}{@{\extracolsep{\fill}} l l r r c c c l c @{}}
\toprule
Model & Dataset & Train & Test & Length & Batch & Steps & Schedule & ZeRO-3 \\
\midrule
Llama-3.1-8B & GSM8K & $7{,}473$ & $1{,}319$ & $1{,}024$ & $128\times16$ & $100$ & Warmup-constant & Yes \\
\bottomrule
\end{tabular*}
\end{table}

\paragraph{Llama-3.2-1B on \tuluthree.}
In this part, we finetune \texttt{Llama-3.2-1B}~\citep{grattafiori2024llama}\footnote{\url{https://huggingface.co/meta-llama/Llama-3.2-1B}} on \texttt{tulu-3-sft-mixture}~\citep{lambert2025tulu}\footnote{\url{https://huggingface.co/datasets/allenai/tulu-3-sft-mixture}}, drawing disjoint random subsets of the dataset for training and for held-out evaluation.
Conversations are rendered in the Llama~3 chat format with the fixed system message ``You are a helpful assistant'', and the loss covers the tokens of assistant turns only.
We evaluate the held-out NLL every $1{,}000$ optimizer steps.

\paragraph{DeepSeek-Coder-6.7B on Magicoder-OSS-Instruct.}
In this part, we finetune a pretrained code model \texttt{deepseek-coder-6.7b-base}~\citep{guo2024deepseek}\footnote{\url{https://huggingface.co/deepseek-ai/deepseek-coder-6.7b-base}} on a public coding dataset \texttt{Magicoder-OSS-Instruct-75K}~\citep{wei2024magicoder}\footnote{\url{https://huggingface.co/datasets/ise-uiuc/Magicoder-OSS-Instruct-75K}}.
Each example pairs a coding problem with a solution, formatted with the prompt given by the dataset authors\footnote{\url{https://github.com/ise-uiuc/magicoder}}, and the loss covers solution tokens only.
We evaluate the final checkpoint on HumanEval~\citep{chen2021codex} and HumanEval+ with the EvalPlus harness~\citep{liu2023evalplus}, using the training prompt format and greedy decoding with at most $768$ new tokens, and report pass@1.
We leverage the vLLM engine~\citep{kwon2023efficient} for generating the model response.

\paragraph{Mistral-7B-v0.1 on Magicoder-OSS-Instruct.}
To examine how the position of trainable layers affects the results, we adapt \texttt{Mistral-7B-v0.1}~\citep{jiang2023mistral7b}\footnote{\url{https://huggingface.co/mistralai/Mistral-7B-v0.1}} with $32$ layers to \texttt{Magicoder-OSS-Instruct-75K}, using the same prompt format described above.
We adopt direct BF16 updates without an FP32 master copy.
For \emph{random-$k$}, we construct each subset by sampling $k$ blocks uniformly at random without replacement. 
This sampling uses a block-selection seed that is separate from the training seed listed in Table~\ref{tab:setup_shared}. 
We discard duplicate subsets and keep $10$ distinct subsets of layer indices for each $k$.
\Cref{fig:layer_distribution} depicts the layer indices of the $10$ sampled subsets for each $k$, where each row is one subset and dark grey cells denote sampled trainable layers.

\subsection{Direct preference optimization (DPO)}
\label{app:exp_setup_dpo}

Table~\ref{tab:setup_dpo} lists the DPO experiments.
For the standard sigmoid DPO setting, response log probabilities are summed over response tokens and the loss is averaged over the pairs in a batch, without length normalization.
The OLMo setting uses the length-normalized sigmoid variant: each response's policy--reference log-ratio is divided by its response length before applying the log-sigmoid.
Reference log probabilities come from the initial policy: they are computed once for all training and held-out pairs before the first update, so no separate reference model is kept during training.

\paragraph{Llama-3.2-1B-Instruct on UltraFeedback.}
\texttt{Llama-3.2-1B-Instruct}~\citep{grattafiori2024llama}\footnote{\url{https://huggingface.co/meta-llama/Llama-3.2-1B-Instruct}} is finetuned on \texttt{UltraFeedback}~\citep{cui2023ultrafeedback}\footnote{\url{https://huggingface.co/datasets/HuggingFaceH4/ultrafeedback_binarized}}.
The official chat template of this model inserts the current date into the
system message, which is inappropriate for our experiments because the
rendered prompts would then depend on the date of the run; we therefore
replace the system message with the fixed text ``You are a helpful assistant''
and otherwise follow the Llama~3 chat format.
Prompt--response sequences longer than the maximum sequence length are
truncated from the right.
We evaluate the DPO loss on the held-out pairs every $1{,}000$ optimizer
steps.

\paragraph{OLMo-2-1124-7B-SFT on its official mixture.}
We finetune \texttt{OLMo-2-1124-7B-SFT}\footnote{\url{https://huggingface.co/allenai/OLMo-2-1124-7B-SFT}} on \texttt{olmo-2-1124-7b-preference-mix}~\citep{walsh2025}\footnote{\url{https://huggingface.co/datasets/allenai/olmo-2-1124-7b-preference-mix}}, formatting each example with the model's official chat template.
We set the maximum sequence length to $2{,}048$ tokens, retain only pairs whose chosen and rejected sequences are both shorter than this limit, and subsample 24,000 training and 2,400 validation pairs from the retained pairs.
We train with the length-normalized sigmoid DPO loss ($\beta=5.0$) and an effective
batch size of 8 pairs, evaluating on the validation subset every $2{,}048$ optimizer steps.

\begin{table}[t]
\centering
\caption{Software versions (top) and hardware of the training node (bottom).}
\label{tab:setup_env}
\setlength{\tabcolsep}{11pt}
\begin{tabularx}{0.89\linewidth}{@{}l X@{}}
\toprule
Component & Version or specification \\
\midrule
PyTorch~\citep{paszke2019pytorch} & $2.11.0$, CUDA~$13.0$ build \\
Transformers~\citep{wolf-etal-2020-transformers} & $5.15.0$ \\
TRL~\citep{vonwerra2020trl} & $1.10.0$ \\
Datasets~\citep{lhoest-etal-2021-datasets} & $5.0.1$ \\
Accelerate~\citep{accelerate} & $1.14.0$ \\
DeepSpeed & $0.19.5$ \\
Kernels & $0.16.0$ \\
FlashAttention~\citep{dao2023flashattention2} & 2.8.3 \\
vLLM~\citep{kwon2023efficient} & 0.26.0 \\
\midrule
GPU & $4\times$ NVIDIA H200, $140$\,GiB each \\
CPU & $2\times$ Intel Xeon Gold 6448Y, $32$ cores each \\
Host memory & $1{,}007$\,GiB \\
Operating system & Ubuntu 24.04.3 LTS, Linux kernel 6.8.0 \\
NVIDIA driver & $590.44.01$ \\
\bottomrule
\end{tabularx}
\end{table}

\subsection{Reinforcement learning with verifiable rewards (RLVR)}
\label{app:exp_setup_rlvr}

Table~\ref{tab:setup_rlvr} lists the RLVR experiment.
We train the model with the GRPO~\citep{shao2024deepseekmath} objective, using the token-level DAPO loss~\citep{yu2025dapo}, which sums the policy-gradient loss over all completion tokens in a batch and divides by their total number, so every token carries the same weight regardless of completion length.
The advantage of a completion is its reward minus the mean reward of the completions sampled for the same problem, divided by the standard deviation of all rewards in the batch~\citep{liu2026tricks}.
Each batch of completions is used for a single optimizer step, so the policy that samples the completions is also the one being updated, and we use no KL penalty, so no reference model is kept during training.
Completions that reach the maximum completion length without an end-of-sequence token are excluded from the loss.
Completions are sampled with a colocated vLLM engine~\citep{kwon2023efficient}.
We correct the mismatch between the vLLM sampler and the training model with a sequence-level importance sampling ratio and drop completions whose ratio exceeds $3$.
In both precisions, the LM head computes the logits of the training forward pass in FP32, following ScaleRL~\citep{khatri2026art}.

\paragraph{Llama-3.1-8B on GSM8K.}
We train \texttt{Llama-3.1-8B}~\citep{grattafiori2024llama}\footnote{\url{https://huggingface.co/meta-llama/Llama-3.1-8B}} on the training split of \texttt{GSM8K}~\citep{cobbe2021gsm8k}\footnote{\url{https://huggingface.co/datasets/openai/gsm8k}} and evaluate on its test split.
The reward is $1$ for a correct final answer, $1.1$ when the correct answer is also given in the boxed format that the prompt asks for, and $0$ otherwise.
Each optimizer step samples $16$ completions for each of $128$ training problems, with temperature $1.0$ and at most $1{,}024$ new tokens per completion.
Training runs for $100$ optimizer steps, about $1.7$ passes over the training problems, with a linear warmup over the first $10$ steps followed by a constant learning rate.
We evaluate on the test split with $4$ sampled responses per problem.

\subsection{Software and hardware}
\label{app:exp_setup_env}

Table~\ref{tab:setup_env} lists the software and hardware environment of our experiment.

\section{Additional Experimental Results}
\label{app:additional_results}

\subsection{Precision-Dependent Scaling across the Number of Trainable Blocks}
\label{app:additional_results:scaling}

\begin{figure}[t]
\centering
\includegraphics[width=\linewidth]{figures/lr_sweep/exp_scaling/shared_legend.pdf}\par\nointerlineskip
\begin{subfigure}[t]{0.500000\linewidth}\captionsetup{margin={5.12pt,0pt}}
  \includegraphics[width=\linewidth]{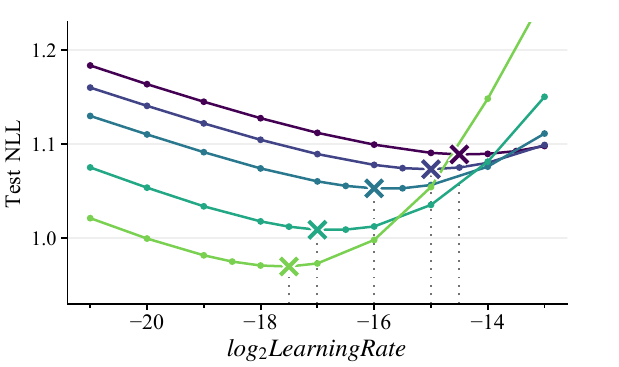}
  \caption{Llama-3.2-1B (SFT)} \label{fig:eval_fp32:fig1}
\end{subfigure}%
\begin{subfigure}[t]{0.500000\linewidth}\captionsetup{margin={5.12pt,0pt}}
  \includegraphics[width=\linewidth]{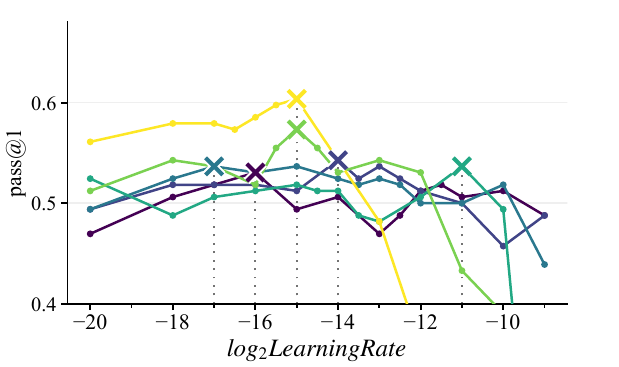}
  \caption{DeepSeek-Coder-6.7B-Base (SFT)}\label{fig:eval_fp32:fig3}
\end{subfigure}\par\medskip
\begin{subfigure}[t]{0.500000\linewidth}\captionsetup{margin={5.12pt,0pt}}
  \includegraphics[width=\linewidth]{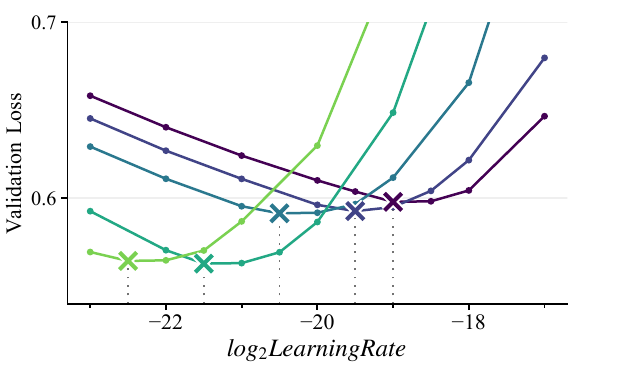}
  \caption{Llama-3.2-1B-Instruct (DPO)}\label{fig:eval_fp32:fig4}
\end{subfigure}%
\begin{subfigure}[t]{0.500000\linewidth}\captionsetup{margin={5.12pt,0pt}}
  \includegraphics[width=\linewidth]{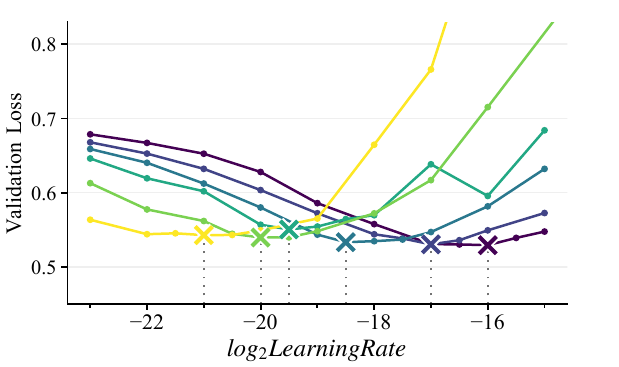}
  \caption{OLMo-2-1124-7B-SFT (DPO)}\label{fig:eval_fp32:fig2}
\end{subfigure}
\caption{
Learning rate sweeps under FP32 master weights, selected by the task-specific validation metric.
(a) SFT of Llama-3.2-1B on \tuluthree{}, validation NLL;
(b) SFT of DeepSeek-Coder-6.7B-Base on Magicoder-OSS-Instruct, HumanEval+ pass@1 (higher is better);
(c) DPO of Llama-3.2-1B-Instruct on UltraFeedback, validation DPO loss;
(d) DPO of OLMo-2-1124-7B-SFT on its official preference mixture, validation DPO loss.
Each curve shows the validation metric of the final checkpoint versus $\log_2$ learning rate with only the last $k$ blocks trainable.
Crosses and dotted lines mark each curve's optimal learning rate: the lowest loss in (a), (c) and (d), and the highest pass@1 in (b).
}
\label{fig:eval_fp32}
\end{figure}

\begin{figure}[t]
\centering
\includegraphics[width=\linewidth]{figures/lr_sweep/exp_scaling/shared_legend.pdf}\par\nointerlineskip
\begin{subfigure}[t]{0.500000\linewidth}\captionsetup{margin={5.12pt,0pt}}
  \includegraphics[width=\linewidth]{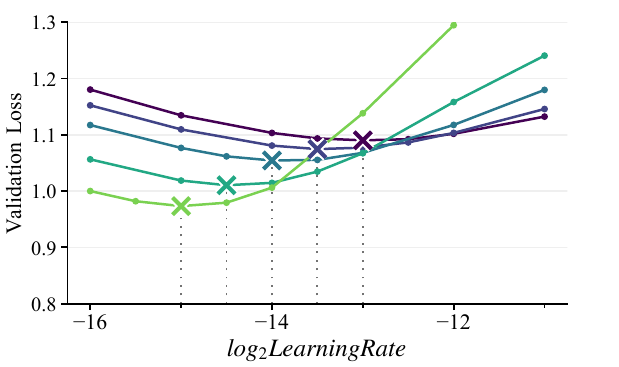}
  \caption{Llama-3.2-1B (SFT)}\label{fig:eval_bf16:fig1}
\end{subfigure}%
\begin{subfigure}[t]{0.500000\linewidth}\captionsetup{margin={5.12pt,0pt}}
  \includegraphics[width=\linewidth]{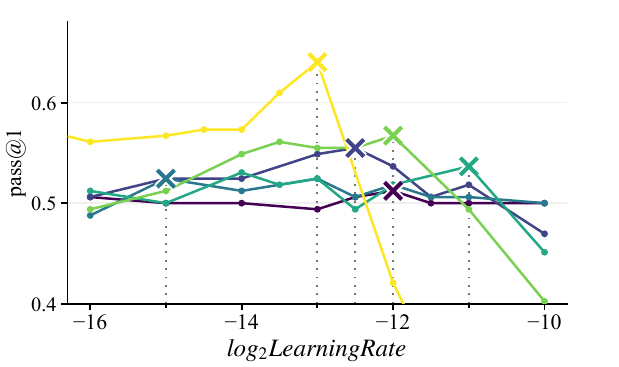}
  \caption{DeepSeek-Coder-6.7B-Base (SFT)}\label{fig:eval_bf16:fig2}
\end{subfigure}\par\medskip
\begin{subfigure}[t]{0.500000\linewidth}\captionsetup{margin={5.12pt,0pt}}
  \includegraphics[width=\linewidth]{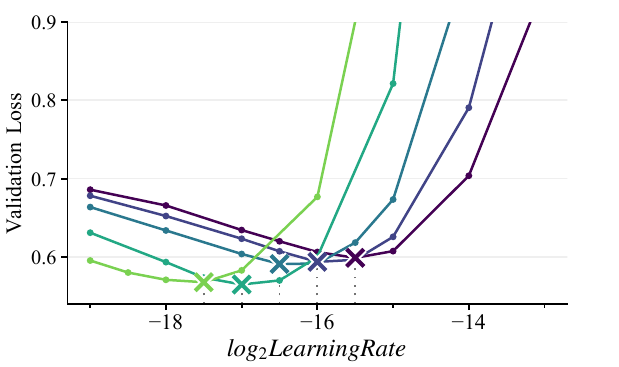}
  \caption{Llama-3.2-1B-Instruct (DPO)}\label{fig:eval_bf16:fig3}
\end{subfigure}%
\begin{subfigure}[t]{0.500000\linewidth}\captionsetup{margin={5.12pt,0pt}}
  \includegraphics[width=\linewidth]{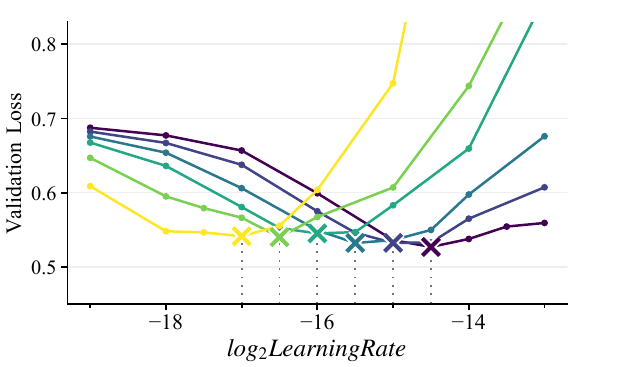}
  \caption{OLMo-2-1124-7B-SFT (DPO)}\label{fig:eval_bf16:fig4}
\end{subfigure}
\caption{
Learning rate sweeps under direct BF16 updates, selected by the task-specific validation metric.
(a) SFT of Llama-3.2-1B on \tuluthree{}, validation NLL;
(b) SFT of DeepSeek-Coder-6.7B-Base on Magicoder-OSS-Instruct, HumanEval+ pass@1 (higher is better);
(c) DPO of Llama-3.2-1B-Instruct on UltraFeedback, validation DPO loss;
(d) DPO of OLMo-2-1124-7B-SFT on its official preference mixture, validation DPO loss.
Each curve shows the validation metric of the final checkpoint versus $\log_2$ learning rate with only the last $k$ blocks trainable.
Crosses and dotted lines mark each curve's optimal learning rate: the lowest loss in (a), (c) and (d), and the highest pass@1 in (b).
}
\label{fig:eval_bf16}
\end{figure}

\paragraph{LR sweep with regard to validation metrics.}
\Cref{fig:eval_fp32,fig:eval_bf16} show the same runs as \Cref{sec:experiments:scaling}, with the learning rate selected by the task-specific validation metric instead of the EMA-smoothed training loss, under FP32 master-weight and direct BF16 weight updates, respectively.
For the three settings evaluated by validation loss, the results agree with the training-loss sweeps: the optimal learning rate decreases with $k$ at roughly $k^{-1}$ under FP32 and $k^{-1/2}$ under BF16.
In addition, the FP32 curves are flatter around their minima, so the selected rate is more sensitive to run-to-run noise.
For DeepSeek-Coder-6.7B-Base, where the validation metric is HumanEval+ pass@1, the picture is different.
For $k < 16$ the curves are nearly flat: learning rates spanning several powers of two give almost the same pass@1, so no single rate stands out.
Only $k = 16$ and $k = 32$ show a clear peak.
By contrast, the training loss has a distinct minimum at every $k$ and changes visibly from one grid point to the next, so it could reveal small differences between learning rates that pass@1 cannot.
We therefore identify the learning rate for transfer from the training loss.

\paragraph{Extension to FP16.}

\begin{figure}[t]
\centering
\includegraphics[width=\linewidth]{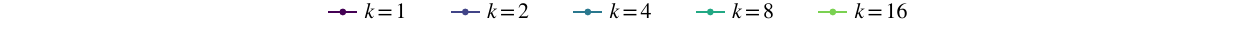}\par\nointerlineskip
\begin{subfigure}[t]{0.500000\linewidth}\captionsetup{margin={5.12pt,0pt}}
  \includegraphics[width=\linewidth]{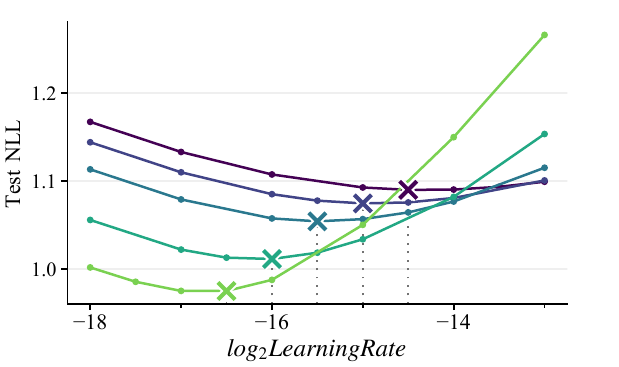}
  \caption{Llama-3.2-1B (SFT)}\label{fig:eval_fp16:fig1}
\end{subfigure}%
\begin{subfigure}[t]{0.500000\linewidth}\captionsetup{margin={5.12pt,0pt}}
  \includegraphics[width=\linewidth]{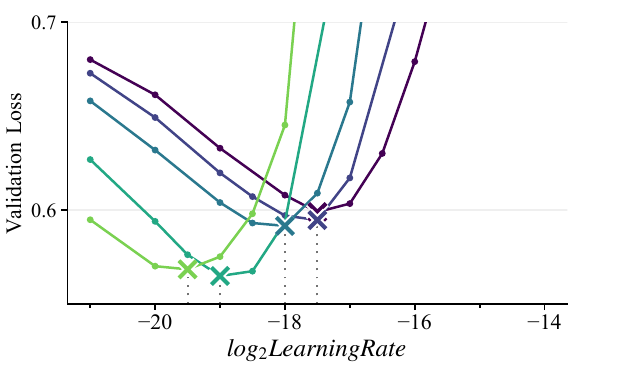}
  \caption{Llama-3.2-1B-Instruct (DPO)}\label{fig:eval_fp16:fig2}
\end{subfigure}
\caption{
Learning rate sweeps for the FP16 experiments, selected by the task-specific validation metric.
(a) SFT of Llama-3.2-1B on \tuluthree{}, validation NLL;
(b) DPO of Llama-3.2-1B-Instruct on UltraFeedback, validation DPO loss.
Each curve shows the validation metric of the final checkpoint versus $\log_2$ learning rate with only the last $k$ blocks trainable.
Crosses and dotted lines mark each curve's optimal learning rate, corresponding to the lowest loss in both panels.
}
\label{fig:eval_fp16}
\end{figure}

\Cref{fig:eval_fp16} shows the learning rate sweeps under direct FP16 updates, with the learning rate selected by the final validation loss.
The optimal learning rate scales approximately as $k^{-1/2}$ for both SFT and DPO under direct FP16 updates, consistent with the findings in \Cref{sec:experiments:scaling}.

\paragraph{Extension to Reinforcement Learning with Verifiable Rewards (RLVR)}

\begin{figure}[t]
\centering
\includegraphics[width=\linewidth]{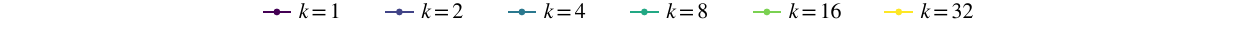}\par\nointerlineskip
\begin{subfigure}[t]{0.500000\linewidth}\captionsetup{margin={5.12pt,0pt}}
  \includegraphics[width=\linewidth]{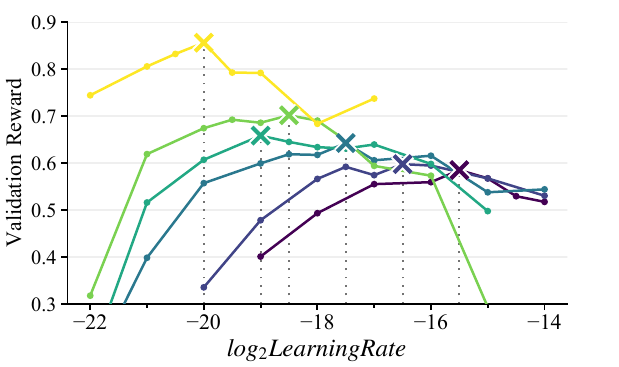}
  \caption{Float32 weights}\label{fig:eval_grpo:fp32}
\end{subfigure}%
\begin{subfigure}[t]{0.500000\linewidth}\captionsetup{margin={5.12pt,0pt}}
  \includegraphics[width=\linewidth]{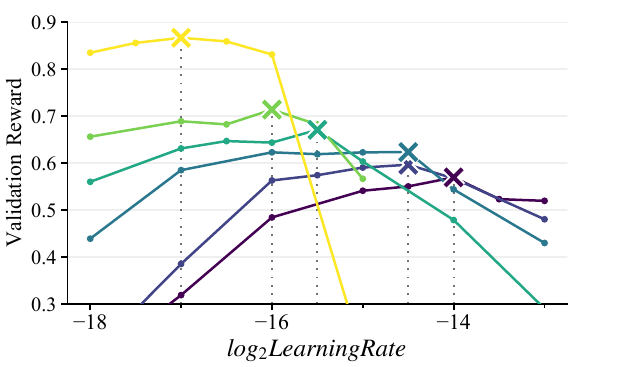}
  \caption{BFloat16 weights}\label{fig:eval_grpo:bf16}
\end{subfigure}
\caption{
Learning rate sweeps for RLVR of Llama-3.1-8B on GSM8K under (a) FP32 master-weight updates and (b) direct BF16 updates.
Each curve presents the validation reward versus $\log_2$ learning rate with only the last $k$ blocks trainable.
Crosses and dotted lines mark each curve's optimal learning rate, selected by the validation reward.
}
\label{fig:eval_grpo}
\end{figure}

\Cref{fig:eval_grpo} shows the learning rate sweeps with regard to validation reward for on-policy RLVR of Llama-3.1-8B on GSM8K dataset.
We observe a similar pattern to our findings in \Cref{sec:experiments:scaling}.

\begin{figure}[t]
\centering
\includegraphics[width=\linewidth]{figures/lr_sweep/exp_scaling/shared_legend.pdf}\par\nointerlineskip
\begin{subfigure}[t]{0.500000\linewidth}\captionsetup{margin={5.12pt,0pt}}
  \includegraphics[width=\linewidth]{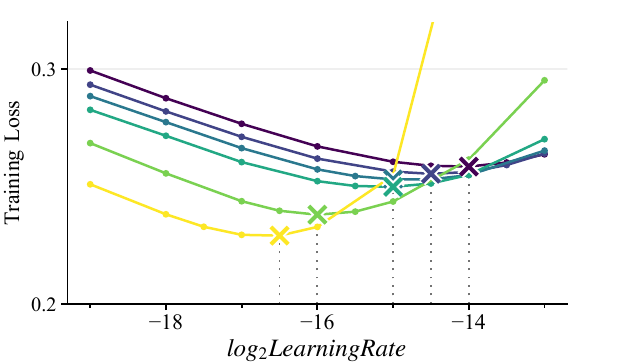}
  \caption{Self-attention modules only}\label{fig:exp_lr_sweep_ablation:attn}
\end{subfigure}%
\begin{subfigure}[t]{0.500000\linewidth}\captionsetup{margin={5.12pt,0pt}}
  \includegraphics[width=\linewidth]{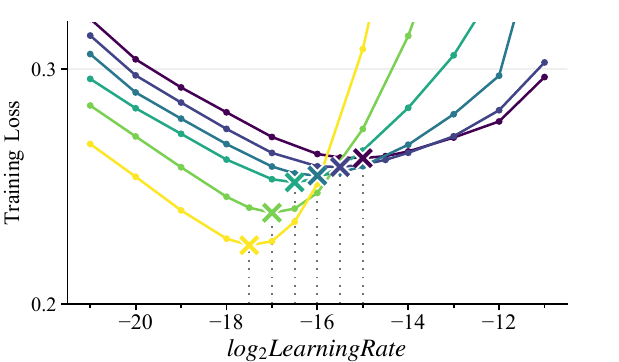}
  \caption{MLP only}\label{fig:exp_lr_sweep_ablation:mlp}
\end{subfigure}%
\caption{
Learning rate sweeps when only the (a) self-attention modules or (b) MLP modules of the last $k$ blocks are trainable.
Both panels show SFT of Mistral-7B-v0.1 on Magicoder-OSS-Instruct under direct BF16 updates, for $k \in \{1, 2, 4, 8, 16, 32\}$.
Each curve shows final training loss (EMA-smoothed) versus $\log_2$ learning rate.
Crosses and dotted lines mark each curve's optimal rate.
}
\label{fig:exp_lr_sweep_ablation}
\end{figure}

\paragraph{Ablation study on finetuning only self-attention or MLP modules.}
We test whether the scaling rule changes when only the self-attention or only the MLP modules of the last $k$ blocks are trainable.
We finetune Mistral-7B-v0.1 on Magicoder-OSS-Instruct under direct BF16 updates.
Results in \Cref{fig:exp_lr_sweep_ablation} show that the optimal learning rate scales as $k^{-1/2}$ in both cases.
At every $k$, the optimal learning rate for the MLP modules is typically lower than that for the self-attention modules.

\begin{figure}[t]
\centering
\includegraphics[width=\linewidth]{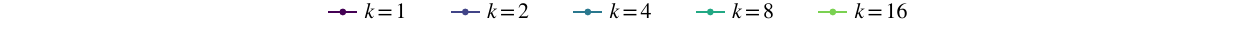}\par\nointerlineskip
\begin{subfigure}[t]{0.500000\linewidth}\captionsetup{margin={5.12pt,0pt}}
  \includegraphics[width=\linewidth]{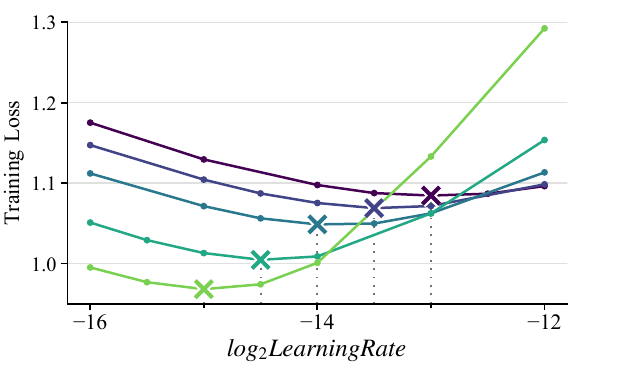}
  \caption{Final training loss}\label{fig:tulu3_bf16_wd:fig1}
\end{subfigure}%
\begin{subfigure}[t]{0.500000\linewidth}\captionsetup{margin={5.12pt,0pt}}
  \includegraphics[width=\linewidth]{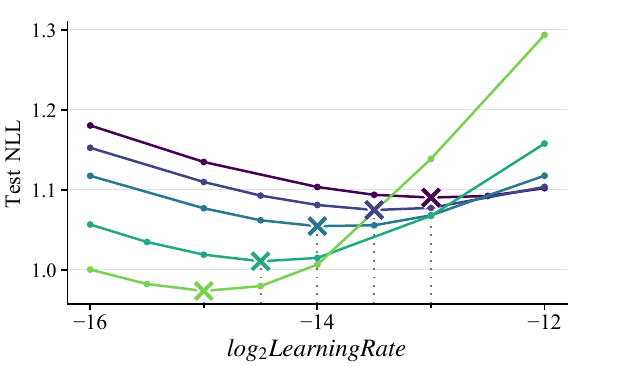}
  \caption{Validation NLL}\label{fig:tulu3_bf16_wd:fig2}
\end{subfigure}
\caption{\arxiv{Learning rate sweeps with a weight decay of 0.01. Both panels show SFT of Llama-3.2-1B on T\"ulu~3 under direct BF16 updates with only the last $k$ blocks trainable, for $k \in \{1, 2, 4, 8, 16\}$. Each curve shows (a) the final training loss (EMA-smoothed) or (b) the validation NLL of the final checkpoint versus $\log_2 \eta$. Crosses and dotted lines mark each curve's optimal learning rate.}}
\label{fig:tulu3_bf16_wd}
\end{figure}

\paragraph{Ablation study on weight decay.}
\arxiv{We test whether the scaling rule changes with a nonzero weight decay. 
We repeat the SFT sweeps of Llama-3.2-1B on T\"ulu~3 under direct BF16 updates, with a weight decay of 0.01 and all other settings unchanged. 
Results in Figure~\ref{fig:tulu3_bf16_wd} show that the optimal learning rate still scales as $k^{-1/2}$, whether selected by the final EMA-smoothed training loss or by the validation NLL.
At every $k$, the optimal learning rate is the same with and without weight decay.
}

\subsection{\arxiv{Effect of gradient and optimizer-state precision}}
\label{app:optimizer_precision}
\begin{figure}[t]
\centering
\includegraphics[width=\linewidth]{figures/lr_sweep/exp_scaling/shared_legend.pdf}\par\nointerlineskip
\begin{subfigure}[t]{0.333333\linewidth}\captionsetup{margin={18.72pt,0pt}}
  \includegraphics[width=\linewidth]{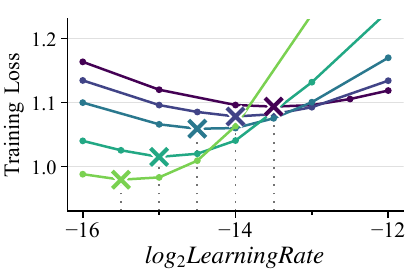}
  \caption{\arxiv{Llama-3.2-1B (SFT)}}
\end{subfigure}%
\begin{subfigure}[t]{0.333333\linewidth}\captionsetup{margin={18.68pt,0pt}}
  \includegraphics[width=\linewidth]{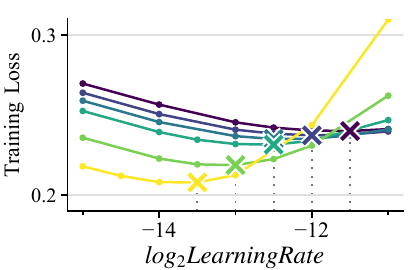}
  \caption{\arxiv{DSCoder-6.7B (SFT)}}
\end{subfigure}%
\begin{subfigure}[t]{0.333333\linewidth}\captionsetup{margin={18.72pt,0pt}}
  \includegraphics[width=\linewidth]{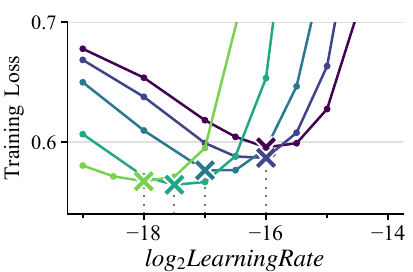}
  \caption{\arxiv{Llama-3.2-1B (DPO)}}
\end{subfigure}%
\caption{\arxiv{Learning rate sweeps under direct BF16 updates with FP32 gradients and optimizer states. (a) SFT of Llama-3.2-1B on T\"ulu~3; (b) SFT of DeepSeek-Coder-6.7B-Base on Magicoder-OSS-Instruct; (c) DPO of Llama-3.2-1B-Instruct on UltraFeedback. Each curve shows final training loss (EMA-smoothed) versus $\log_2$ learning rate with only the last $k$ blocks trainable. 
}}
\label{fig:lr_sweeps_bf16_fp32optim}
\end{figure}

\arxiv{
Our main experiments with direct BF16 updates store the weights, gradients, and AdamW first- and second-moment estimates in BF16 (Appendix~\ref{app:exp_setup_shared}).
In this ablation, we test whether the precision of gradients and optimizer states affects the observed scaling rules for direct BF16 updates.
For SFT tasks, we train Llama-3.2-1B on T\"ulu~3 and DeepSeek-Coder-6.7B-Base on Magicoder-OSS-Instruct.
For DPO tasks, we train Llama-3.2-1B-Instruct on UltraFeedback using the standard DPO loss with $\beta = 0.1$.
In each setting, we apply direct BF16 updates but keep the gradients, gradient accumulation, and optimizer states in FP32.
For DeepSeek-Coder-6.7B-Base, we adopt DeepSpeed ZeRO-3 with two GPUs as in its FP32 runs, but without the FP32 master copy.
Therefore, in all three settings, these runs differ from their FP32 counterparts in~\Cref{sec:experiments:scaling} only in update precision.
Results in \Cref{fig:lr_sweeps_bf16_fp32optim} show that the optimal learning rate still scales approximately as $k^{-1/2}$ in all three settings.
The change from $k^{-1}$ to $k^{-1/2}$ scaling therefore comes from update precision, not from the precision of gradients or optimizer states.
}

\subsection{\arxiv{Effect of stochastic rounding}}
\label{app:stochastic_rounding}

\begin{figure}[t]
\centering
\includegraphics[width=\linewidth]{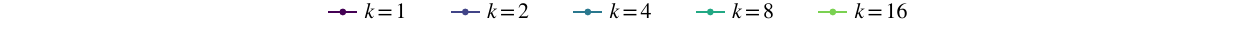}\par\nointerlineskip
\begin{subfigure}[t]{0.500000\linewidth}\captionsetup{margin={5.12pt,0pt}}
  \includegraphics[width=\linewidth]{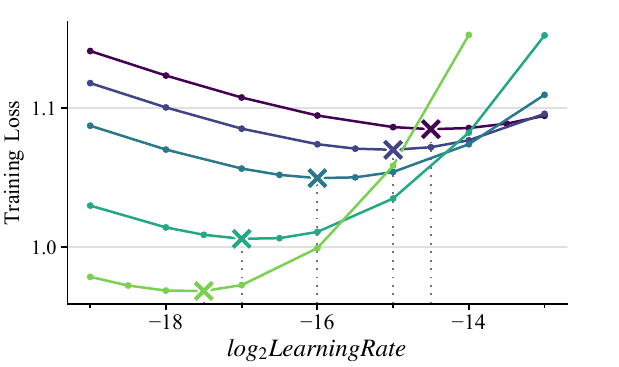}
  \caption{\arxiv{Final training loss}}\label{fig:tulu3_bf16_sr:fig1}
\end{subfigure}%
\begin{subfigure}[t]{0.500000\linewidth}\captionsetup{margin={5.12pt,0pt}}
  \includegraphics[width=\linewidth]{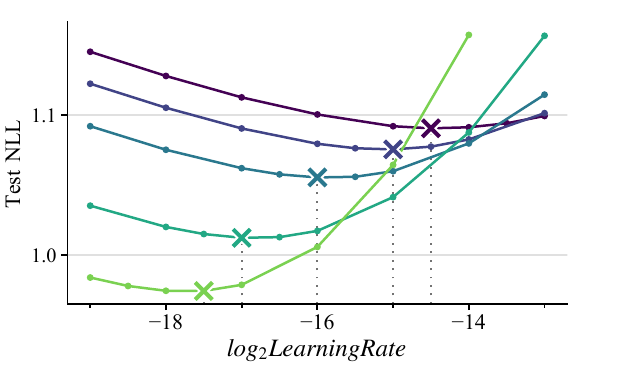}
  \caption{\arxiv{Validation NLL}}\label{fig:tulu3_bf16_sr:fig2}
\end{subfigure}
\caption{\arxiv{Learning rate sweeps with stochastic rounding of the weight updates. Both panels show SFT of Llama-3.2-1B on T\"ulu~3 under direct BF16 updates with only the last $k$ blocks trainable, for $k \in \{1, 2, 4, 8, 16\}$. Each curve shows (a) the final training loss (EMA-smoothed) or (b) the validation NLL of the final checkpoint versus $\log_2 \eta$. 
Crosses and dotted lines mark each curve’s optimal learning rate.
}}
\label{fig:tulu3_bf16_sr}
\end{figure}

\arxiv{Our analysis in~\Cref{sec:theory} attributes the $k^{-1/2}$ scaling under direct BF16 updates to round-to-nearest, which suppresses small updates.
In this ablation, we test this explanation with stochastic rounding, which rounds each coordinate up or down at random so that the realized update is unbiased.
For fixed weights, gradients, and update directions, stochastic rounding gives $\mathbb{E}[\phi_k(\eta)] = 1$, so the expected first-order loss reduction is $\eta D_k$.
Keeping it comparable across $k$, as in Assumption~\ref{asm:theory-stable}, then gives $\eta_k = \Theta(k^{-1})$ when $D_k = \Theta(k)$.
We repeat the SFT runs of Llama-3.2-1B on T\"ulu~3 from Appendix~\ref{app:optimizer_precision}, but round the updated weights to BF16 stochastically instead of to the nearest value, so the two sets of runs differ only in the rounding mode.
Results in \Cref{fig:tulu3_bf16_sr} show that the optimal learning rate scales approximately as $k^{-1}$, whether selected by the final EMA-smoothed training loss or by the validation NLL.
Without the suppression of small updates, the optimal learning rate drops to the FP32 level.
Thus, the change of scaling might come from the rounding operation used.
}

\clearpage

\subsection{Effect of Trainable-Block Placement}
\begin{figure}[t]
\centering
\begin{subfigure}[t]{0.48\linewidth}
  \includegraphics[width=\linewidth]{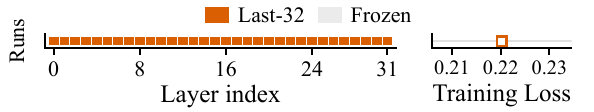}
  \caption{$k = 32$}
\end{subfigure}\hfill
\begin{subfigure}[t]{0.48\linewidth}
  \includegraphics[width=\linewidth]{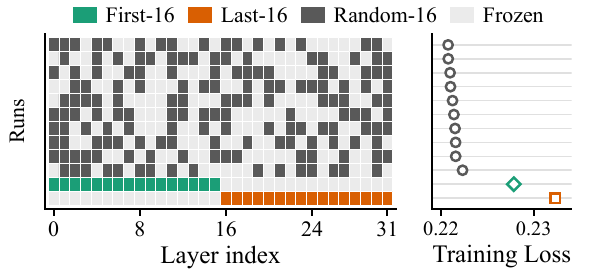}
  \caption{$k = 16$}
\end{subfigure}
\par\medskip
\begin{subfigure}[t]{0.48\linewidth}
  \includegraphics[width=\linewidth]{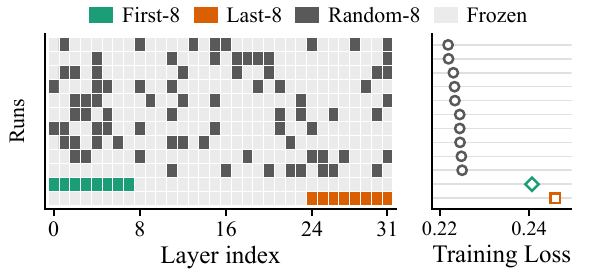}
  \caption{$k = 8$}
\end{subfigure}\hfill
\begin{subfigure}[t]{0.48\linewidth}
  \includegraphics[width=\linewidth]{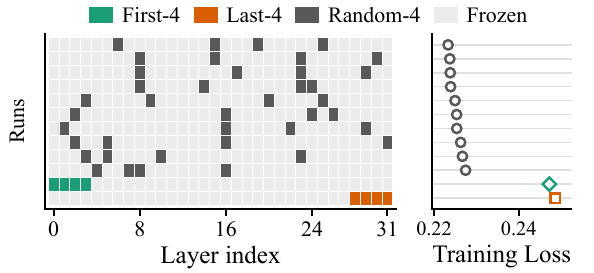}
  \caption{$k = 4$}
\end{subfigure}
\par\medskip
\begin{subfigure}[t]{0.48\linewidth}
  \includegraphics[width=\linewidth]{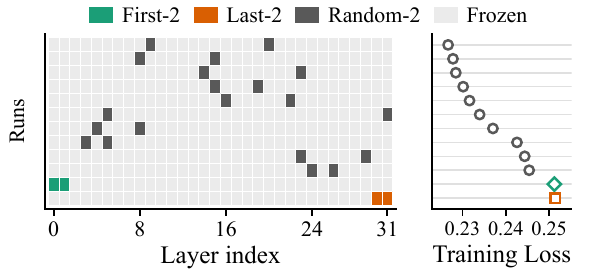}
  \caption{$k = 2$}
\end{subfigure}\hfill
\begin{subfigure}[t]{0.48\linewidth}
  \includegraphics[width=\linewidth]{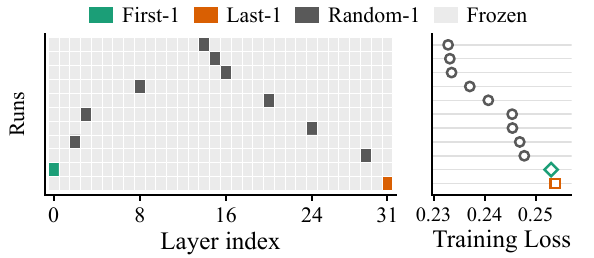}
  \caption{$k = 1$}
\end{subfigure}
\caption{Trainable-layer placements for the runs in~\Cref{fig:layer_placement}. Each subfigure corresponds to a fixed number of trainable blocks $k$. Within each subfigure, each row shows one placement: the first $k$ layers
(green), the last $k$ layers (orange), or $k$ randomly selected layers (grey; one row per subset). Filled and pale cells indicate trainable and frozen layers, respectively.
The right axis reports each placement's final training loss at its best learning rate. Within each subfigure, rows are ordered from lowest to highest loss.}
\label{fig:layer_distribution}
\end{figure}

\Cref{fig:layer_distribution} shows the positions of the trainable blocks within each sampled subset, with subsets ordered by final training loss.

\clearpage

\section{FLOP Cost Analysis}
\label{app:flops_analysis}

\subsection{Setup and Counting Conventions}

\paragraph{The learning rate search procedure.}
Given a single pretrained checkpoint, we run a sweep of $R$ independent learning rate trials. 
In each trial the top $k_0$ transformer blocks are trainable, including their normalization parameters, and every other parameter stays frozen.
The searched learning rate is then scaled to the value used for a finetuning run in which all transformer blocks are trainable. 
The baseline is a vanilla sweep of $R$ full-depth trials over the same learning rate grid. 
The proxy pipeline must pay for the final full-depth run as well, while the vanilla sweep pays for nothing beyond the sweep.

The finetuning setting here is offline LLM post-training on a fixed tokenized dataset, such as SFT and DPO. 
We assume that each example has fixed token sequences, attention masks, and position IDs, and that the frozen prefix is evaluated deterministically. 
Its outputs can therefore be cached once and reused across trials. 
Cached examples may still be shuffled, provided the sequence context that determines each cached feature remains unchanged.
This caching characteristic does not extend to on-policy reinforcement learning, whose training tokens are sampled from the current policy rather than from a fixed dataset.

\paragraph{Notation.}

\begin{table}[t]
\caption{Notation used in the cost model.}
\label{tab:flops_notation}
\begin{center}
\begin{tabular}{@{}ll@{}}
\toprule
{\bf Symbol} & {\bf Meaning} \\
\midrule
$B$ & batch size, in sequences \\
$T$ & sequence length, in tokens \\
$L$ & number of transformer blocks \\
$d$ & hidden size \\
$d_{\mathrm{ffn}}$ & feed-forward hidden size \\
$n_h$, $d_h$ & number of attention heads and dimension per head \\
$V$ & vocabulary size \\
$k_0$ & number of trainable top blocks, $k_0 \in \{1,\dots,L\}$ \\
$R$ & number of learning rate trials in the sweep \\
$N = 12Ld^2$ & non-embedding parameter count \\
$\gamma$ & head cost relative to one block \\
\bottomrule
\end{tabular}
\end{center}
\end{table}

\Cref{tab:flops_notation} summarizes the symbols used in the analysis.
The nominal architecture is a dense decoder-only transformer with full multi-head self-attention and a two-layer feed-forward network (FFN), where $n_h d_h = d$ and $d_{\mathrm{ffn}} = 4d$. 
Its language-model head is untied from the input embedding and frozen in every run, including the full-depth runs.

\paragraph{Counting convention and omitted terms.}
We report \emph{analytical FLOPs}, counting the floating-point operations in the dense matrix products of the forward and backward passes, with each multiply--add contributing two FLOPs~\citep{kaplan2020scaling,narayanan2021efficient,hoffmann2022training}.
The count is hardware-independent, so it does not change with kernel efficiency or device utilization. 
It is also not the arithmetic a run actually performs, which includes every term excluded below. 
Counts are per token unless stated otherwise.
For DPO, the analysis counts only policy-model computation and assumes that reference log probabilities are already available.

Considered matrix products are the attention projections, the two attention matrix products, the feed-forward network, and the LM head. 
Everything else is omitted, including the embedding lookup; bias and residual additions; LayerNorm or RMSNorm normalization; elementwise nonlinearities; rotary and other positional encodings; attention scaling, masking and dropout; the attention softmax; and the vocabulary softmax with its cross-entropy. 
Each omitted term is lower order in the model dimension. 
The largest is the attention softmax, at $O(BT^2 n_h)$ per layer against the attention products at $4BT^2 d = 4BT^2 n_h d_h$. 
It is therefore smaller by a factor of order $d_h$ wherever those products appear. 
The vocabulary softmax costs $O(V)$ per token against the counted head at $2dV$.

Two further terms, activation recomputation (or gradient checkpointing) and optimizer updates, are omitted. 
Under standard blockwise gradient checkpointing, recomputation adds approximately one additional forward pass of each checkpointed block per training step. 
Checkpointing all trainable blocks therefore adds approximately $L\,\Cblkfwd$ to a full-depth step, whereas a proxy step incurs only $k_0\,\Cblkfwd$ and, for small $k_0$, may not need checkpointing at all.
Optimizer-update cost is approximately proportional to the number of trainable parameters, so training $k_0$ of the $L$ blocks reduces this cost by a factor of about $L/k_0$.
These omitted costs can increase the compute savings relative to the analytical counts reported here, depending on the checkpointing policy.

\subsection{Proxy-Trial Costs and Cache Amortization}

\paragraph{Forward propagation.}
Consider a batch of $B$ sequences of length $T$. 
Within a single transformer block, the query, key and value projections cost $6BTd^2$, the attention scores $2BT^2d$, the attention--value product $2BT^2d$, and the output projection $2BTd^2$. 
The feed-forward network applies two linear layers at $2BTd\,d_{\mathrm{ffn}}$ each, or $16BTd^2$ together when $d_{\mathrm{ffn}} = 4d$. 
Summing,
\[
\begin{aligned}
\Cblkfwd &= 6BTd^2 + 2BT^2d + 2BT^2d + 2BTd^2 + 16BTd^2 \\
      &= 24BTd^2 + 4BT^2d .
\end{aligned}
\]
A forward pass comprises $L$ such blocks and the language-model head, which projects the final hidden states onto the vocabulary at
\[
\Cheadfwd = 2BTdV .
\]
Summing over the blocks and the head,
\begin{equation}\label{eq:cfwd}
\begin{aligned}
\Cfullfwd &= L\,\Cblkfwd + \Cheadfwd \\
&=L(24BTd^2 + 4BT^2d) + 2BTdV.
\end{aligned}
\end{equation}

\paragraph{Backward propagation.}
For a matrix product $Y = XW$, the backward pass computes the input gradient $\nabla_X = \nabla_Y W^{\top}$ and the weight gradient $\nabla_W = X^{\top}\nabla_Y$, each at the same cost as the forward product. 
A trainable block therefore costs $2\,\Cblkfwd$. 
The head is frozen, so its backward pass computes only the gradient with respect to the final hidden states, at $\Cheadfwd$. 
In a full-depth run every block is trainable, giving
\[
\Cfullbwd = 2L\,\Cblkfwd + \Cheadfwd .
\]
With \Eqref{eq:cfwd}, one full-depth run costs
\begin{equation}\label{eq:cfull}
\begin{aligned}
\Cfull  &= \Cfullfwd + \Cfullbwd \\
        &= 3L\,\Cblkfwd + 2\,\Cheadfwd \\
        &= 3L\left(24BTd^2 + 4BT^2d\right) + 4BTdV .
\end{aligned}
\end{equation}

\paragraph{Attention convention.}
The attention products above are counted as dense $T \times T$ matrix
products, at $4BT^2d$ per block. 
Counting only the causal pairs $j \le i$ gives $2BdT(T+1)$, close to half. The factor applies to the attention term alone, which reaches the scale of the $24BTd^2$ from the projections and the feed-forward network only when $T \gtrsim 6d$.

\paragraph{Consistency with the $6N$ rule.}
Dropping the attention products and the LM head from \Eqref{eq:cfull} leaves $72BTLd^2$, or $6N$ per token with $N = 12Ld^2$, the standard estimate of training cost in \citet{kaplan2020scaling,hoffmann2022training}. 
\Eqref{eq:cfull} is that rule with two explicit corrections, one for the attention products and one for the frozen head,
\[
\Cfull = BT\left(6N + 12LTd + 4dV\right) .
\]

\subsection{Cost of one proxy run}

\paragraph{Trainability assumptions.}
In a proxy trial, the top $k_0$ blocks, numbered $L-k_0+1$ through $L$, are trainable, while blocks $1$ through $L-k_0$, the input embedding, and the LM head are frozen. 
Each trainable block costs $2\,\Cblkfwd$ in the backward pass, including block $L-k_0+1$, whose trainable input normalization requires the input gradient of its first matrix product.

\paragraph{Without feature caching.}
Without caching, every trial recomputes the outputs of blocks $1$ through $L-k_0$, so its forward pass comprises all $L$ blocks and the LM head. 
The backward pass reaches only the LM head and the $k_0$ trainable blocks, since nothing below block $L-k_0+1$ receives a gradient. 
A trial therefore costs
\begin{equation}\label{eq:cnc}
\begin{aligned}
\Cuncached
&= \underbrace{L\,\Cblkfwd + \Cheadfwd}_{\text{forward}}
 + \underbrace{2k_0\,\Cblkfwd + \Cheadfwd}_{\text{backward}} \\
&= (L + 2k_0)\,\Cblkfwd + 2\,\Cheadfwd .
\end{aligned}
\end{equation}
Let
\[
\gamma = \frac{\Cheadfwd}{\Cblkfwd}
\]
denote the forward cost of the LM head relative to one transformer block.
For the nominal architecture above, $\gamma = V/(12d + 2T)$. 
Dividing \Eqref{eq:cfull} by \Eqref{eq:cnc} gives the compute-reduction factor of a trial against one full-depth run,
\begin{equation}\label{eq:rhonc}
\rhonc = \frac{\Cfull}{\Cuncached}
       = \frac{3L + 2\gamma}{L + 2k_0 + 2\gamma} .
\end{equation}

The bound $\rhonc < 3$ is strict for every $L$, every $k_0 \ge 1$ and every $\gamma \ge 0$, and it is tight. 
In situations where $k_0 = 1$ with $V \ll 12d + 2T$,
\[
\rhonc \approx \frac{3L}{L + 2} ,
\]
which approaches $3$ as $L$ grows. 
Freezing the lower blocks removes their backward pass and nothing else. 
Their forward pass is repeated in every trial, while a full run costs at most three times a single forward pass, so without caching no choice of $k_0$ can save more than a factor of three.

\paragraph{With feature caching.}

Under the fixed-input and deterministic-prefix assumptions above, the outputs of the frozen blocks can be computed once and stored for reuse, at the cost of disk storage that is cheap relative to accelerator memory. 
The cache point is the output of block $L-k_0$, which is the input of block $L-k_0+1$. 
The cache is built with one forward pass of the dataset through blocks $1$ through $L-k_0$,
\begin{equation}\label{eq:cache}
\Cbuild = (L - k_0)\,\Cblkfwd ,
\end{equation}
paid once for the whole sweep. 
A cached trial makes one pass over the dataset, reading the stored features and passing them forward and backward through the $k_0$ trainable blocks and the LM head,
\[
\Ccached = 3k_0\,\Cblkfwd + 2\,\Cheadfwd = (3k_0 + 2\gamma)\,\Cblkfwd .
\]
The head contributes little to the full-depth and uncached costs when $\gamma \ll L$, but it can contribute substantially to a cached proxy when $\gamma$ is comparable to $k_0$.
In particular, its fraction of the cached-trial cost is $2\gamma/(3k_0 + 2\gamma)$.
A sweep of $R$ cached trials costs $\Cbuild + R\,\Ccached$, against $R\,\Cfull$ for $R$ full-depth trials, so
\begin{equation}\label{eq:rhoc}
\rhocached(R,k_0) = \frac{R\,(3L + 2\gamma)}{(L - k_0) + R\,(3k_0 + 2\gamma)} ,
\end{equation}
with limit
\begin{equation}\label{eq:rhocinf}
\rhocached(\infty,k_0) = \lim_{R \to \infty} \rhocached(R,k_0)
= \frac{3L + 2\gamma}{3k_0 + 2\gamma} .
\end{equation}
For $k_0 < L$ the cache term $(L-k_0)/R$ falls as $R$ grows, so $\rhocached$ increases with $R$ and approaches \eqref{eq:rhocinf} from below.
At $R = 1$ the denominator of~\Eqref{eq:rhoc} is $(L-k_0) + 3k_0 + 2\gamma = L + 2k_0 + 2\gamma$, so $\rhocached(1,k_0) = \rhonc$. 
Building the cache and reading it once is a full forward pass, so a single cached pass saves nothing. 
With $E$ epochs per trial, every $R$ in~\Eqref{eq:rhoc} becomes $RE$ and~\Eqref{eq:rhocinf} is unchanged. 
The $R$-dependence enters only through amortization, which is why~\Eqref{eq:rhonc} carries no $R$ at all. 
Any saving beyond $\rhonc$ comes from spreading \Eqref{eq:cache} over more proxy trials.

\subsection{End-to-End Cost of Learning Rate Transfer}

\paragraph{The two sweep procedures.}
The proxy pipeline builds the cache and runs $R$ cached trials. 
The learning rate selected by the sweep is then scaled for full-depth training, and the target is finetuned once at the scaled value. 
The full-depth sweep, in contrast, runs $R$ trials and then keeps the
best one. 
The final run of the proxy pipeline costs the same as one full-depth trial.

\begin{proposition}\label{prop:e2e}
For $1 \le k_0 \le L$ and $R \ge 1$, the ratio of the analytical FLOPs required by the full-depth sweep to those required by the learning rate transfer with cache is
\begin{equation}\label{eq:rhoe2e}
\rhoe(R,k_0)
= \frac{R\,(3L + 2\gamma)}
{\underbrace{(L - k_0)}_{\text{cache build}}
 + \underbrace{R\,(3k_0 + 2\gamma)}_{\text{proxy trials}}
 + \underbrace{(3L + 2\gamma)}_{\text{full-depth run}}} .
\end{equation}
\end{proposition}
\begin{proof}
The full-depth sweep costs $R\,\Cfull$ and the pipeline
$\Cbuild + R\,\Ccached + \Cfull$. Taking the ratio and dividing through
by $\Cblkfwd$ gives~\Eqref{eq:rhoe2e}.
\end{proof}

The ratio in~\Eqref{eq:rhoe2e} increases in $R$ and decreases in $k_0$, so a larger sweep or a smaller $k_0$ raises it. 
However, neither raises it without limit.

\begin{corollary}\label{cor:bounds}
For $1 \le k_0 \le L$ and $R \ge 1$,
\begin{equation}\label{eq:bounds}
\rhoe(R,k_0) < \min\!\left\{ R,\; \frac{3L + 2\gamma}{3k_0 + 2\gamma} \right\}
\;\le\; \min\!\left\{ R,\; \frac{L}{k_0} \right\} .
\end{equation}
\end{corollary}
\begin{proof}
In~\Eqref{eq:rhoe2e}, the cache term is nonnegative, while the proxy-trial and final-run terms are strictly positive.
Consequently, discarding any of these terms raises the ratio $\rhoe(R,k_0)$. 
Discarding the cache and proxy-trial terms leaves $R$, and discarding the cache and final-run terms leaves $(3L + 2\gamma)/(3k_0 + 2\gamma)$. 
For the second inequality, $k_0(3L + 2\gamma) \le L(3k_0 + 2\gamma)$ reduces to $\gamma k_0 \le \gamma L$, which holds because $\gamma \ge 0$ and $k_0 \le L$.
\end{proof}

Two separate limits are at work, the sweep size $R$ and the network depth $L$. 
The final run is common to both procedures and costs as much as one full-depth trial, so the proxy pipeline always pays at least a $1/R$ fraction of what the full-depth sweep pays, whatever $k_0$ is. 
A proxy trial trains $k_0$ blocks against $L$ for a full-depth trial,
so no proxy trial saves more than a factor of $L/k_0$, which is at most $L$,
attained at $k_0 = 1$.

\subsection{FLOP Estimates for Llama-2 Configurations}
\label{app:numbers}
\begin{table}[t]
\caption{Llama-2 configurations used.}
\label{tab:configs}
\begin{center}
\begin{tabular}{@{}lrrrrr@{}}
\toprule
{\bf Model} & $L$ & $d$ & $n_h$ & $d_{\mathrm{ffn}}$ & $V$ \\
\midrule
Llama-2 7B  & 32 & 4096 & 32 & 11008 & 32000   \\
Llama-2 13B & 40 & 5120 & 40 & 13824 & 32000  \\
Llama-2 70B & 80 & 8192 & 64 & 28672 & 32000  \\
\bottomrule
\end{tabular}
\end{center}
\end{table}

\begin{figure}[!ht]
\begin{center}
\includegraphics[width=\textwidth]{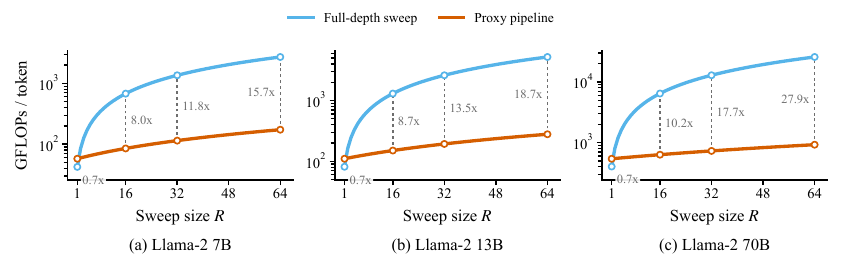}
\end{center}
\caption{Analytical FLOPs per token against sweep size $R$, at $k_0 = 1$ and $T = 2048$. 
The proxy pipeline includes the cache build, $R$ cached trials and one final full-depth run, so it is more expensive than a single
full-depth trial at $R = 1$. 
Grey vertical dashed lines mark the gap between the two curves at $R \in \{1, 16, 32, 48, 64\}$, labeled with the ratio $\rhoe(R,1)$
of~\Eqref{eq:rhoe2e}.}
\label{fig:flops_analysis_llama2}
\end{figure}

We instantiate \Cref{prop:e2e} using the nominal architecture with the depths, hidden sizes, and vocabulary sizes of the three Llama-2 models~\citep{touvron2023llama}, listed in \Cref{tab:configs}.
We fix $k_0 = 1$ and $T = 2048$, and vary the number of trials $R$.
Let
\[
\cblkfwd = \Cblkfwd/(BT) = 24d^2 + 4Td
\]
denote the forward cost of one block per token. 
A full-depth sweep of $R$ trials then costs
\[
R\,(3L + 2\gamma)\,\cblkfwd
\]
per target token, and the proxy pipeline costs
\[
\big[ (L - 1) + R\,(3 + 2\gamma) + (3L + 2\gamma) \big]\,\cblkfwd .
\]

\Cref{fig:flops_analysis_llama2} presents analytical FLOPs per token against sweep size for the three Llama-2 models. 
At $R = 1$ the ratio is below one in every model. 
In a single-trial sweep, the full-depth baseline requires exactly one full-depth run. 
By contrast, the proxy pipeline must perform that same run in addition to building the feature cache and executing one proxy trial, so it cannot be cheaper.
The ratio increases with $R$ as the cache build and final full-depth run are amortized over more proxy trials, but the gains diminish as repeated proxy trials account for a larger share of the total cost. 
For small sweeps, the cost of building the cache and performing the final full-depth run limits the compute reduction, since both costs are incurred once regardless of the number of trials.
For sufficiently large sweeps, repeated proxy trials dominate the pipeline cost, and the reduction approaches the cached per-trial factor $\rhocached(\infty,k_0)=(3L + 2\gamma)/(3k_0 + 2\gamma)$.
For $k_0 = 1$, this limit remains below $L$ because every cached trial still incurs the forward and backward costs of the frozen LM head.

\paragraph{Departures from the nominal architecture.}
Llama-2 utilizes a gated FFN with three matrices, and the 70B variant additionally uses grouped-query attention with eight key--value heads. 
These differences change the forward cost of a block but affect the compute-reduction factors only through $\gamma$. 
At $T = 2048$, accounting for the actual architectures increases the full-depth training FLOPs by approximately $0.5\%$, $0.8\%$, and $6.0\%$ relative to the nominal estimates for 7B, 13B, and 70B, respectively. 
Nonetheless, with $k_0 = 1$, the relative change in $\rhoe(R,1)$ remains below $0.5\%$ for all models over the plotted range $1 \le R \le 64$.
Thus, the nominal architecture closely approximates the compute-reduction factors reported here, despite a modest discrepancy in absolute FLOPs for 70B.

\section{Cached Learning Rate Search: Prototype Evaluation}

\label{app:case_study}

\begin{figure}[!t]
\begin{center}
\includegraphics[width=\textwidth]{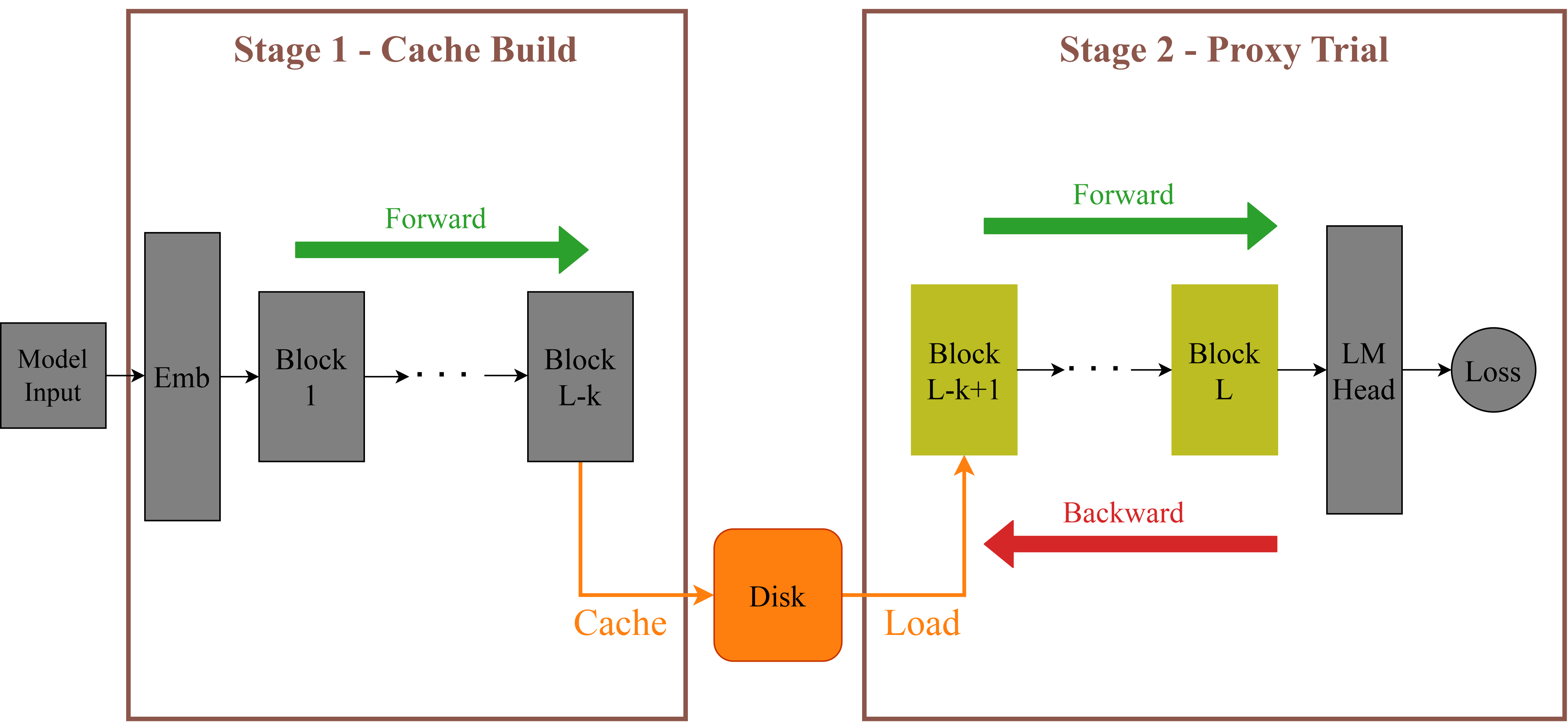}
\end{center}
\caption{Two-stage prototype for learning rate search. Stage 1 (cache build) stores the output of block $L-k_0$ on a local disk after a forward pass through the frozen prefix. Stage 2 (cached proxy trials) reuses these features to train blocks $L-k_0+1$ through $L$ for each candidate learning rate. The LM head remains frozen, but gradients propagate through it to the trainable blocks.}
\label{fig:case_study:prototype}
\end{figure}

To evaluate the practical efficiency of our methods, we build a prototype consisting of two stages: cache build (Stage 1) and cached proxy trials (Stage 2).
The prototype architecture is shown in~\Cref{fig:case_study:prototype}.
In Stage 1, we pass the input tokens through the frozen prefix up to block $L-k_0$ and cache the resulting features on a local disk.
This stage runs in inference mode and requires only forward propagation.
In Stage 2, we load the cached features and train blocks $L-k_0+1$ through $L$ for each candidate learning rate.
Each trial performs forward and backward propagation through these $k_0$ blocks and the frozen LM head.
The $k_0$ trainable blocks serve as a proxy for the full-depth model.

We evaluate the prototype using the 32-block \texttt{deepseek-coder-6.7b-base}~\citep{guo2024deepseek} and the Magicoder-OSS-Instruct dataset~\citep{wei2024magicoder}, which contains 75k coding question and solution pairs.
We use an effective batch size of $32$ and a maximum sequence length of $1024$ tokens.
We restrict this case study to $k_0=1$, where only the last block is trainable during cached proxy trials.
After measuring memory and runtime, we verify in~\Cref{app:case_study:transfer_quality} that the learning rate transferred from this proxy yields a competitive full-depth model.

\subsection{Peak Allocated GPU Memory}
\label{app:two_stage_prototype:peak_GPU_mem}


We measure the peak allocated GPU memory of each stage as the per-GPU batch size increases from $1$ to $1024$ sequences.
To evaluate worst-case memory use, we construct training batches in which every sequence contains exactly $1024$ tokens.
For comparison, we also measure peak memory for full-depth training with and without gradient checkpointing.
We run all experiments on a single H200 GPU with 140GiB of memory.
The results are shown in~\Cref{fig:main:case_study:memory_fp32,fig:main:case_study:memory_bf16}.

At small batch sizes, both stages use substantially less memory than full-depth training.
This difference reflects the memory required by full-depth training to store model weights, optimizer states, and gradients, even at a batch size of $1$.
The resulting initial memory requirement is particularly high with \texttt{float32} master weights, as shown in~\Cref{fig:main:case_study:memory_fp32}.
Gradient checkpointing reduces the rate at which memory usage grows with batch size in full-depth training, at the cost of additional recomputation time.
Stage 2 does not use gradient checkpointing, and its memory curve is steeper than that of a checkpointed full-depth training.
Applying gradient checkpointing to Stage 2 could similarly reduce this growth rate and allow larger batch sizes during training.

\subsection{Learning rate search on a consumer GPU}

\begin{table}[t]
  \centering
  \small
  \caption{Hardware configuration for the case study on a consumer GPU.}
  \label{tab:case_study_setup_hardware}
  \setlength{\tabcolsep}{11pt}
  \begin{tabularx}{0.8\linewidth}{@{}l X@{}}
  \toprule
  Component & Specification \\
  \midrule
  GPU & $1\times$ NVIDIA GeForce RTX 5060 Ti, $15.9$\,GiB \\
  CPU & $1\times$ AMD Ryzen 9 9900X, $12$ cores, $24$ threads \\
  Host memory & $30.5$\,GiB \\
  Operating system & Ubuntu 24.04.4 LTS, Linux kernel 7.0.0-31-generic \\
  NVIDIA driver & $595.84$ \\
  \bottomrule
  \end{tabularx}
\end{table}

The preceding memory analysis suggests that our prototype can run on a consumer GPU, but its runtime on such hardware remains to be assessed.
We therefore benchmark the prototype on a single NVIDIA GeForce RTX 5060 Ti with 15.45 GiB of GPU memory available to PyTorch.
The specific hardware configurations are in~\Cref{tab:case_study_setup_hardware}.
With this system, we evaluate on the same 6.7B model with a maximum sequence length of $1024$ tokens.
Stage 2 utilizes gradient accumulation to maintain an effective batch size of $32$.

\begin{table}[!t]
\centering
\caption{Runtime (in hours) and feature-cache storage (GiB) for learning rate search with the $k_0=1$ prototype on a single RTX 5060 Ti with 15.45 GiB of GPU memory available to PyTorch.
In this case, only the last transformer block is trainable during proxy trials.
We report the fastest benchmarked configuration for each stage.
For Stage 2, BF16 denotes \texttt{bfloat16} weights and optimizer states, and FP32 denotes \texttt{float32} master weights with \texttt{bfloat16} autocast.
Each proxy trial makes one pass over the dataset.
Sweep runtimes include one cache build and the specified number of proxy trials, but exclude the final full-depth training run.}
\label{tab:case_study:runtime}
\begin{tabular}{lcc}
\toprule
 & BF16 & FP32 \\
\midrule
Stage 1 batch size & $1$ & $1$ \\
Stage 2 micro-batch $\times$ accumulation & $1 \times 32$ & $2 \times 16$ \\
\midrule
Stage 1, run once & 3.78\,h & 3.78\,h \\
Stage 2, per learning rate & 0.58\,h & 0.71\,h \\
Cache build + one proxy trial & 4.36\,h & 4.49\,h \\
Cache build + five proxy trials & 6.68\,h & 7.34\,h \\
Cache build + ten proxy trials & 9.58\,h & 10.89\,h \\
\midrule
Feature-cache size & 336.8\,GiB & 336.8\,GiB \\
\bottomrule
\end{tabular}
\end{table}

\Cref{tab:case_study:runtime} summarizes the fastest benchmarked configurations for each precision, the per-stage runtime, and the sweep runtime for different learning rate grid sizes.
Cache build accounts for most of the runtime when evaluating a single learning rate, but this cost is paid only once across a sweep.
Each additional learning rate requires only a Stage 2 trial, making it substantially less expensive to extend the search after the cache has been built.
The precision choice has a modest effect on the per-trial runtime of Stage 2, whereas cache-build runtime is comparable across the two configurations.
In Stage 1, both configurations use \texttt{bfloat16} for the main forward computations, with autocast applied when weights are stored in \texttt{float32}.
Since this stage performs only forward propagation without weight updates, the two precisions have comparable cache-build runtime.

\begin{table}[!t]
\centering
\caption{Effect of batch size on one-pass runtime.
$B$ is the Stage 1 batch size and the Stage 2 micro-batch size.
Stage 2 uses $32/B$ gradient accumulation steps.
All runtimes are in hours.
The final column reports the total number of tokens processed over one dataset pass, including padding.}
\label{tab:case_study:runtime_batch_size}
\begin{tabular}{rrrrc}
\toprule
$B$ &
\shortstack{Stage 1\\(BF16)} &
\shortstack{Stage 2\\(BF16)} &
\shortstack{Stage 2\\(FP32)} &
\shortstack{Tokens after padding\\(millions)} \\
\midrule
1  & \textbf{3.78} & \textbf{0.58} & 0.82 & 44.35 \\
2  & 4.44 & 0.68 & \textbf{0.71} & 51.02 \\
4  & 5.07 & 0.74 & 0.81 & 56.87 \\
8  & 5.02 & 0.71 & 0.74 & 61.79 \\
16 & 5.87 & 0.85 & 0.89  & 65.79 \\
\bottomrule
\end{tabular}
\end{table}

On the consumer GPU tested here, the largest feasible batch does not minimize runtime.
As shown in~\Cref{tab:case_study:runtime_batch_size}, the fastest benchmarked configurations use small batches in both stages. 
Because stage 1 is already compute-bound on this GPU at a batch size of 1, larger batches yield no throughput gain. 
Moreover, since each batch is padded to its longest sequence, larger batches introduce additional padding tokens, which incur extra computation and further increase the runtime.

The prototype's low GPU memory consumption comes with a disk-storage requirement, reported in~\Cref{tab:case_study:runtime}. Both precision configurations use the same cache of \texttt{bfloat16} features, so reducing the precision of weights and optimizer states in Stage 2 does not reduce the cache size.
Nonetheless, disk storage is generally less expensive per unit of capacity than GPU memory, and existing cache compression techniques could further reduce the storage requirement.

Under the tested configurations, the final full-depth run still requires more GPU memory than the RTX 5060 Ti provides.
Nonetheless, the prototype allows learning rate sweeps to run on more accessible consumer hardware, so that high-end GPUs can be reserved solely for the final full-depth run.
Decoupling the two phases in this way may reduce the cost of hyperparameter search and make larger search grids practical within a fixed compute budget.

\paragraph{Full-depth training on H200 GPUs.}
\Cref{tab:case_study:fulldepth} reports full-depth runtime for the same model and dataset on the H200 computing node used for our main experiments.
For each precision and GPU setting, we benchmark the runtime with different configurations and adopt the fastest benchmarked configuration for reporting.
With FP32 update precision, ten trials on the RTX 5060 Ti take 10.89\,h, including cache construction but excluding the final full-depth run. 
This is comparable to the approximately 9.1\,h required for ten full-depth trials on two H200s with DeepSpeed ZeRO-3.

\begin{table}[!t]
\centering
\small
\renewcommand{\arraystretch}{1.15}
\caption{Full-depth training on the H200 computing node, using the fastest benchmarked configuration for each precision and GPU setting. Each run trains all $32$ blocks for one epoch at an effective batch
size of $32$. BF16 uses \texttt{bfloat16} weights and optimizer states, and FP32 uses \texttt{float32} master weights with \texttt{bfloat16} autocast. Micro-batch sizes are per GPU. Single-GPU runs do not use DeepSpeed. Peak memory is the maximum allocated memory per GPU during training.}
\label{tab:case_study:fulldepth}
\begin{tabular}{@{}lccccrr@{}}
\toprule
Precision &
\shortstack{H200\\GPUs} &
ZeRO-3 &
\shortstack{Micro-batch\\$\times$ accumulation} &
\shortstack[c]{Gradient\\ Checkpointing} &
\shortstack[r]{Peak memory\\(GiB/GPU)} &
\shortstack[r]{Runtime\\(h)} \\
\midrule
BF16 & 1 & No  & $4 \times 8$  & No & 73.0  & 1.48 \\
FP32 & 1 & No  & $2 \times 16$ & No & 123.2 & 1.93 \\
FP32 & 2 & Yes & $8 \times 2$  & No & 104.9 & 0.91 \\
\bottomrule
\end{tabular}
\end{table}

\begin{table}[!t]
\centering
\small
\setlength{\tabcolsep}{10pt}
\renewcommand{\arraystretch}{1.15}
\caption{GPU hours by hardware tier for a search over $R$ candidate
learning rates.
The prototype includes one cache build and $R$ proxy trials on the
RTX 5060 Ti with 15.45 of available memory, followed by one final full-depth run on the H200 with 140 GiB of memory.
A full-depth sweep uses the H200 node for all $R$ trials.
BF16 employs one H200 and FP32 uses two with ZeRO-3
(\Cref{tab:case_study:fulldepth}), so FP32 H200 hours are twice the
wall-clock time.}
\label{tab:case_study:gpu_hours}
\begin{tabular}{@{}lcrrr@{}}
\toprule
 & & \multicolumn{2}{c}{Prototype} &
\multicolumn{1}{c}{Full-depth sweep} \\
\cmidrule(lr){3-4}\cmidrule(l){5-5}
Precision & $R$ & RTX 5060 Ti & H200 & H200 \\
\midrule
BF16 & 1  & 4.36  & 1.48 & 1.48  \\
     & 5  & 6.68  & 1.48 & 7.38  \\
     & 10 & 9.58  & 1.48 & 14.76 \\
\midrule
FP32 & 1  & 4.49  & 1.83 & 1.83  \\
     & 5  & 7.34  & 1.83 & 9.14  \\
     & 10 & 10.89 & 1.83 & 18.29 \\
\bottomrule
\end{tabular}
\end{table}

\Cref{tab:case_study:gpu_hours} reports GPU hours separately by hardware tier, including the final full-depth run for the prototype. 
For the tested sweep sizes, the prototype uses 4.36--10.89\,h on the consumer GPU and requires one full-depth run on the H200 rather than $R$.
Although full-depth training exceeds the RTX 5060 Ti's memory capacity under the tested configurations, the search uses no high-end GPUs.

\subsection{Quality of the Transferred Learning Rate}
\label{app:case_study:transfer_quality}

\begin{table}[!t]
\centering
\small
\setlength{\tabcolsep}{5pt}
\setlength{\belowcaptionskip}{6pt}
\renewcommand{\arraystretch}{1.15}

\caption{Pass@1 (\%) for \texttt{deepseek-coder-6.7b-base} on
HumanEval and HumanEval+ after being finetuned on Magicoder-OSS-Instruct
at the transferred learning rate. $\hat{\eta}_1$ is the rate selected by the $k_0=1$ proxy sweep, and $\hat{\eta}_{32}$ is the rate obtained by scaling it for $L=32$ with the scaling rule of the corresponding update precision. Full-depth trains all $32$ transformer blocks and keeps the input embeddings, final normalization layer, and LM head frozen. Full-parameter trains every parameter. Our runs train for one epoch at an effective batch size of $32$ with a maximum sequence length of $1024$ tokens, and the final checkpoint is evaluated with EvalPlus using greedy decoding. The first two rows are reference values reported by \citet{wei2024magicoder}: the base model without finetuning, and Magicoder-DS, which the dataset authors obtained by finetuning all parameters for two epochs with Adafactor at a learning rate of $5\times10^{-5}$, a batch size of $512$, and a sequence length of $1216$, keeping the weights and optimizer states in \texttt{float32} under \texttt{bfloat16} autocast, that is, FP32 master-weight updates in our terminology. HumanEval and HumanEval+ contain $164$ problems each, so one problem corresponds to around $0.6$ percentage points.}
\label{tab:case_study:transfer_quality}

\begin{tabular*}{\linewidth}{@{\extracolsep{\fill}}llccrr@{}}
\toprule
\multirow{2}{*}{Model / training scope}
  & \multirow{2}{*}{Precision}
  & \multicolumn{2}{c}{Learning rate}
  & \multicolumn{2}{c}{Pass@1 (\%)} \\
\cmidrule(lr){3-4}\cmidrule(l){5-6}
  & & Proxy $\hat{\eta}_1$ & Training & HumanEval & HumanEval+ \\
\midrule
\begin{tabular}[c]{@{}l@{}}
DeepSeek-Coder-6.7B-Base \\
(no finetuning)
\end{tabular}
  & --- & --- & --- & 47.6 & 39.6 \\
\addlinespace[2pt]
Magicoder-DS-6.7B
  & FP32 & --- & --- & 66.5 & 60.4 \\
\midrule
Full-depth (ours)
  & \multirow{2}{*}{BF16}
  & \multirow{2}{*}{$2^{-11.5}$}
  & \multirow{2}{*}{$2^{-14}$}
  & 64.0 & 57.3 \\ 
Full-parameter (ours)
  & & & & 64.6 & 59.1 \\ 
\addlinespace[4pt]
Full-depth (ours)
  & \multirow{2}{*}{FP32}
  & \multirow{2}{*}{$2^{-12}$}
  & \multirow{2}{*}{$2^{-17}$}
  & 64.0 & 57.9 \\ 
Full-parameter (ours)
  & & & & 65.2 & 59.8 \\ 
\bottomrule
\end{tabular*}
\end{table}

We next evaluate learning rate transfer from the $k_0=1$ proxy to full-depth training on the same model and dataset.
The proxy sweeps select $\hat{\eta}_1 = 2^{-11.5}$ under direct BF16 updates and $\hat{\eta}_1 = 2^{-12}$ under FP32 master-weight updates.
For $L=32$, the precision-dependent scaling rules give $\hat{\eta}_{32} = 2^{-11.5}/\sqrt{32} = 2^{-14}$ under BF16 and $\hat{\eta}_{32} = 2^{-12}/32 = 2^{-17}$ under FP32.
For each precision, we train two configurations at the transferred rate.
The \emph{full-depth} configuration trains all $L=32$ transformer blocks while keeping the input embeddings, final normalization layer, and LM head frozen, matching the target of our transfer rule.
The \emph{full-parameter} configuration trains every parameter, as in Magicoder-DS~\citep{wei2024magicoder}.
All runs follow the training protocol of the main experiments.
We evaluate the final checkpoints on HumanEval and HumanEval+ with the EvalPlus harness and the vLLM engine, using greedy decoding, as described in Appendix~\ref{app:experimental_setup}.

\Cref{tab:case_study:transfer_quality} reports the pass@1 metric alongside two reference values from \citet{wei2024magicoder}: the base model without finetuning, and Magicoder-DS, the released model finetuned on the same dataset.
At the transferred rates, all four runs raise pass@1 from $47.6\%$ to $64.0$--$65.2\%$ on HumanEval and from $39.6\%$ to $57.3$--$59.8\%$ on HumanEval+.
Their scores are within $2.5$ percentage points of Magicoder-DS on HumanEval and $3.1$ on HumanEval+, corresponding to at most four and five of the $164$ coding problems, respectively.
However, Magicoder-DS uses two training epochs with the Adafactor optimizer at a batch size of $512$, whereas our runs use one epoch with the AdamW optimizer at a batch size of $32$.
The comparison therefore does not isolate the learning rate choice, but shows that the transferred rates yield competitive checkpoints under our own training protocol.

Two further observations follow.
First, BF16 and FP32 differ by at most one solved problem for each configuration and benchmark, despite an eightfold difference in their transferred learning rates.
This shows that directly using BF16 for accumulating updates can save the GPU memory with little performance degradation.
Second, full-parameter training outperforms full-depth training at the same rate by one or two problems on HumanEval and three on HumanEval+.
Although our analysis covers only transformer blocks, the transferred rates remain effective in this setting when all parameters are trained, without separate learning rate tuning conducted.




\subsection{\arxiv{Multi-seed evaluation of the transferred learning rate}}
\label{app:multi_seed}

\arxiv{Our runs in~\Cref{tab:case_study:main_quality} adopt a single seed of $42$.
To analyze the robustness of the transferred learning rate across multiple seeds, we repeat the learning rate transfer for the full-depth configuration on two more seeds under each precision.
On each seed, we run the $k_0 = 1$ proxy search, scale the selected rate to $k = 32$ via the scaling rule, and do a full-depth run at the transferred rate.
Only the seed changes, and the model, dataset, training protocol, and evaluation are the same as in Appendix~\ref{app:case_study:transfer_quality}.}

\paragraph{\arxiv{Transferred rates.}}
\arxiv{Under BF16, the proxy sweep selects $2^{-11.5}$ on all three seeds, so the transferred rate is always $2^{-14}$.
Under FP32, it selects $2^{-12}$ on seeds 13 and 42 and $2^{-11.5}$ on seed 11, which gives transferred rates of $2^{-17}$ and $2^{-16.5}$.
These are two neighboring grid points, consistent with the flatter FP32 curves at small $k$ noted in~\Cref{sec:experiments:scaling}.}

\paragraph{\arxiv{Sweep rates from full-depth sweep.}}
\arxiv{For comparison, we also do full-depth runs at the two best rates of the full-depth sweep in~\Cref{sec:experiments:scaling}, which uses seed 42.
The rate with the highest HumanEval+ pass@1 is $2^{-13}$ under BF16 and $2^{-15}$ under FP32.
The rate with the lowest final training loss is $2^{-16}$ under FP32, and under BF16 it equals the transferred rate.
These two rates are known only after the full-depth sweep, which the transfer avoids.
We do a full-depth run at each of these rates on seeds 11 and 13.
We do not repeat the full-depth sweep on the new seeds, so the best rates on these seeds are unknown.}

\paragraph{\arxiv{Results.}}
\arxiv{\Cref{tab:multi_seed} reports pass@1 of every full-depth run, with the mean and standard deviation over the three seeds.
Overall, the transferred rate is robust across seeds and performs on par with the best rates of the full-depth sweep.
Three observations follow.
First, the proxy search is stable.
It selects the same rate on all three seeds under BF16 and two neighboring grid points under FP32.
Second, the sweep rate with the highest HumanEval+ under seed $42$ is less robust.
It scores higher than the transferred rate on the seed it was selected on, but lower on both benchmarks on other seeds 11 and 13, which is reflected in its large standard deviation.
Third, under FP32, the sweep rate with the lowest final training loss scores slightly higher than the transferred rate on HumanEval+ on every seed, while neither rate is consistently better on HumanEval.}

\begin{table}[!t]
\centering
\caption{\arxiv{Per-seed results of the multi-seed evaluation for full-depth training of DeepSeek-Coder-6.7B-Base. In the full-depth sweep under seed $42$, the rate with the highest HumanEval+ pass@1 is $2^{-13}$ under BF16 and $2^{-15}$ under FP32, and the rate with the lowest final training loss under FP32 is $2^{-16}$. Under BF16, the lowest-loss rate equals the transferred rate, so it has no separate row. Pass@1 is reported in \%, where higher is better. The last column is the mean $\pm$ standard deviation.}}
\label{tab:multi_seed}
{\small\setlength{\tabcolsep}{4pt}
\begin{tabular*}{0.99\linewidth}{@{\extracolsep{\fill}}llcccc@{}}
\toprule
Precision & LR & Seed 11 & Seed 13 & Seed 42 & Mean $\pm$ std \\
\midrule
\multicolumn{6}{@{}l}{\textit{HumanEval+ pass@1}} \\
BF16 & Transferred & 60.4 & 57.9 & 57.3 & 58.5 $\pm$ 1.6 \\
BF16 & $2^{-13}$ & 57.3 & 56.7 & 64.0 & 59.3 $\pm$ 4.1 \\
FP32 & Transferred & 59.8 & 60.4 & 57.9 & 59.3 $\pm$ 1.3 \\
FP32 & $2^{-15}$ & 57.3 & 54.3 & 60.4 & 57.3 $\pm$ 3.0 \\
FP32 & $2^{-16}$ & 60.4 & 61.0 & 58.5 & 60.0 $\pm$ 1.3 \\
\midrule
\multicolumn{6}{@{}l}{\textit{HumanEval pass@1}} \\
BF16 & Transferred & 64.6 & 64.0 & 64.0 & 64.2 $\pm$ 0.4 \\
BF16 & $2^{-13}$ & 59.8 & 62.2 & 69.5 & 63.8 $\pm$ 5.1 \\
FP32 & Transferred & 64.6 & 65.9 & 64.0 & 64.8 $\pm$ 0.9 \\
FP32 & $2^{-15}$ & 62.2 & 59.8 & 64.6 & 62.2 $\pm$ 2.4 \\
FP32 & $2^{-16}$ & 63.4 & 66.5 & 64.0 & 64.6 $\pm$ 1.6 \\
\bottomrule
\end{tabular*}}
\end{table}

\section{\arxiv{Cache-Free Learning Rate Search: Prototype Evaluation}}
\label{app:cache_free}

\begin{figure}[!t]
\begin{center}
\includegraphics[width=\textwidth]{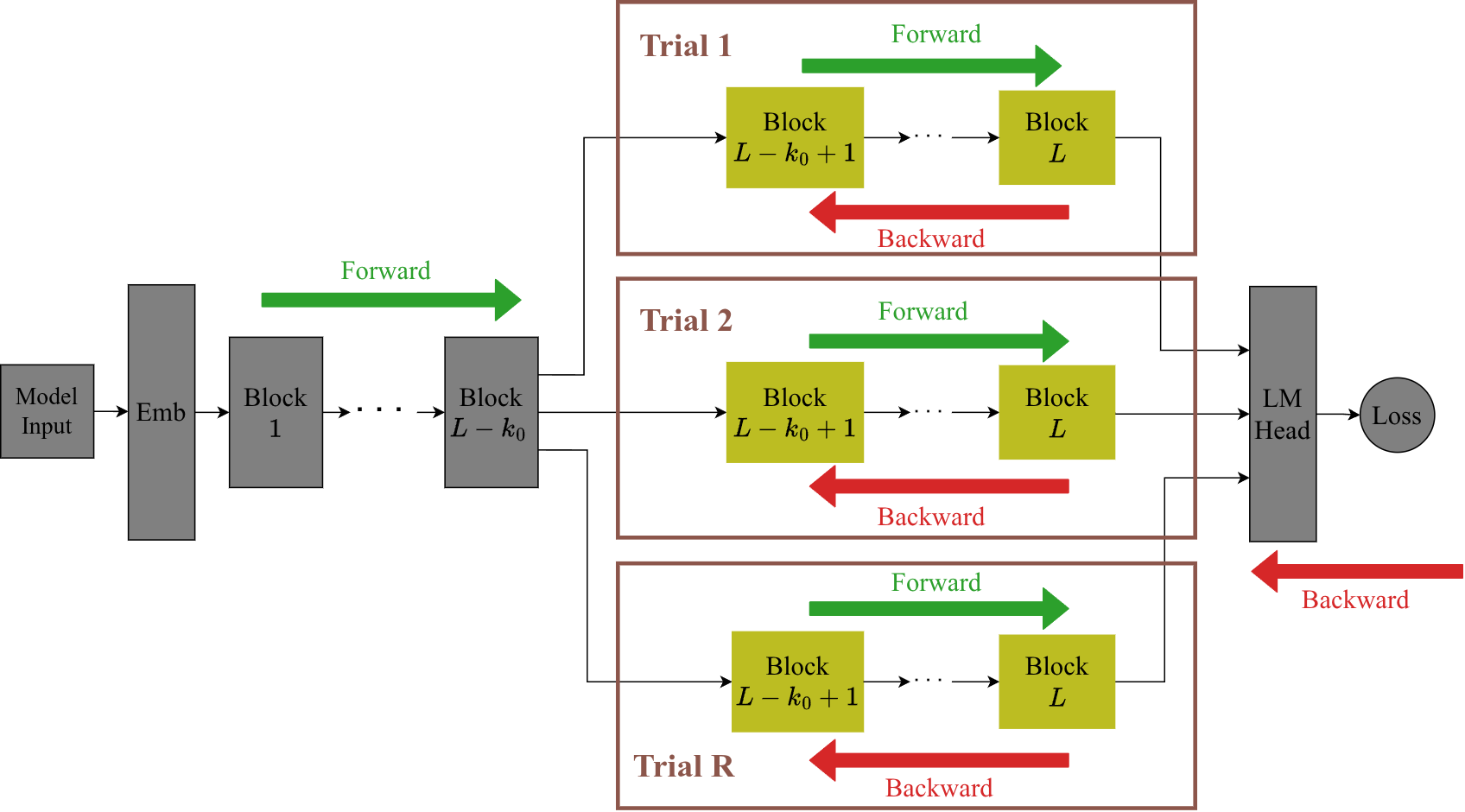}
\end{center}
\caption{\arxiv{Cache-free prototype for learning rate search. The $R$ proxy trials run simultaneously, each of which trains its own copy of blocks $L-k_0+1$ through $L$ at a different candidate learning rate. The frozen prefix processes each batch once and passes the output of block $L-k_0$ directly to all copies. The LM head remains frozen and is shared by all trials, but gradients propagate through it to each copy. The whole sweep thus takes one pass over the dataset. The cost is GPU memory that grows with $R$, so this prototype is suited to high-end GPUs with plenty of memory.}}
\label{fig:case_study:prototype_2}
\end{figure}

\arxiv{
This appendix describes the details of the cache-free prototype evaluated in \Cref{sec:experiments:case_study}.
Its architecture is shown in \Cref{fig:case_study:prototype_2}.
The prototype keeps one copy of the frozen part of the model and $R$ copies of blocks $L-k_0+1$ through $L$, one per candidate learning rate, each with its own gradients and AdamW states.
For each batch, the frozen prefix runs once with forward propagation only, and the output of block $L-k_0$ stays in the GPU memory instead of being written to the local disk.
The $R$ copies then consume this output one after another, each performing forward and backward propagation through its own blocks and the shared LM head, as in a cached proxy trial.
A single pass over the dataset thus completes every proxy trial.
We conduct all experiments on a single H200 GPU with 140~GiB of memory.
}

\subsection{\arxiv{Peak Allocated GPU Memory}}
\label{app:cache_free_prototype:peak_GPU_mem}

\arxiv{
As in \Cref{app:two_stage_prototype:peak_GPU_mem}, we use DeepSeek-Coder-6.7B-Base with $k_0=1$, so each copy contains only the last transformer block, and we construct batches in which every sequence contains exactly 1024 tokens to evaluate the worst-case memory use.
For \Cref{fig:main:case_study_cache_free:memory_fp32,fig:main:case_study_cache_free:memory_bf16}, we measure the peak allocated GPU memory with $R = 1$, $2$, $5$, $10$, $15$, \dots, $40$ trials at per-GPU batch sizes of $B = 1$, $16$, $32$, and $64$.
Every configuration that fits on the GPU follows
\begin{equation}
M_{\mathrm{peak}}(R, B) \;=\; M_0(B) \;+\; 4\,P\,b\,R,
\label{eq:cache_free:memory}
\end{equation}
where $P$ million is the number of parameters in one copy, $b$ is the number of bytes per parameter, and the factor of four counts the weights, gradients, and AdamW first- and second-moment estimates of a copy.
This gives the increments of 1.51~GiB and 3.02~GiB per trial reported in \Cref{sec:experiments:case_study}.
The offset $M_0(B)$ covers the frozen weights and the activations of the prefix and of a single copy.
It grows from 12.3~GiB at $B=1$ to 28.3~GiB under BF16 and 29.8~GiB under FP32 at $B=64$.
Under FP32, 40 trials fit at $B=1$ and $B=16$ but run out of memory at $B=32$ and $B=64$.
}

\subsection{\arxiv{Learning Rate Search on a High-End GPU}}
\label{app:cache_free_prototype:runtime}

\arxiv{
We measure the runtime of the search for sweep sizes $R \in \{5, 10, 15, 20\}$.
For each $R$, we benchmark several configurations and report the fastest one.
The configurations vary the per-GPU batch size $B$ and use $32/B$ gradient accumulation steps to keep the effective batch size at 32.
Both the search and the final full-depth run use a single H200 GPU, and the full-depth sweep reuses the runtime results in \Cref{tab:case_study:fulldepth}.
\Cref{tab:cache_free:gpu_hours} lists the GPU hours behind \Cref{fig:main:case_study_cache_free:running_time}, together with the intermediate sweep size $R=15$.
The per-candidate cost of about 3 minutes is the forward and backward propagation through one copy of the last block and the frozen LM head.
Larger sweeps should widen the gap to the full-depth sweep further, as long as the $R$ copies fit in the GPU memory.
}

\begin{table}[!t]
\centering
\small
\setlength{\tabcolsep}{10pt}
\renewcommand{\arraystretch}{1.15}
\caption{\arxiv{H200 GPU hours to search over $R$ candidate learning rates with the cache-free prototype versus the full-depth sweep.
The cache-free prototype trains $R$ copies of the last block, one per candidate learning rate, in a single pass over the dataset on one H200. It then performs one final full-depth run with the transferred learning rate on one H200 (\Cref{tab:case_study:fulldepth}). \emph{Total} is the sum of the two.
The full-depth sweep trains all $R$ candidates at full depth, using one H200 for BF16 and two H200s with ZeRO-3 for FP32.
All runs use the fastest benchmarked micro-batch $\times$ gradient-accumulation configuration at an effective batch size of 32.
\emph{Speed-up} is the full-depth sweep GPU hours divided by the prototype's \emph{Total}.}}
\label{tab:cache_free:gpu_hours}
\begin{tabular}{@{}lcrrrcr@{}}
\toprule
 & & \multicolumn{3}{c}{Cache-free prototype} & \multicolumn{1}{c}{Full-depth sweep} & \\
\cmidrule(lr){3-5}\cmidrule(lr){6-6}
Precision & $R$ & Search & Final run & Total & H200 & Speed-up \\
\midrule
BF16 & 5  & 0.73 & 1.48 & 2.21 & 7.38  & 3.34$\times$ \\
     & 10 & 0.99 & 1.48 & 2.47 & 14.76 & 5.98$\times$ \\
     & 15 & 1.24 & 1.48 & 2.72 & 22.14 & 8.14$\times$ \\
     & 20 & 1.48 & 1.48 & 2.96 & 29.52 & 9.97$\times$ \\
\midrule
FP32 & 5  & 0.76 & 1.93 & 2.69 & 9.14  & 3.40$\times$ \\
     & 10 & 1.03 & 1.93 & 2.96 & 18.29 & 6.18$\times$ \\
     & 15 & 1.30 & 1.93 & 3.23 & 27.43 & 8.49$\times$ \\
     & 20 & 1.57 & 1.93 & 3.50 & 36.58 & 10.45$\times$ \\
\bottomrule
\end{tabular}
\end{table}

\end{document}